\documentclass{article}

    \usepackage[preprint]{neurips_2021}

\usepackage[utf8]{inputenc} 
\usepackage[T1]{fontenc}    
\usepackage{hyperref}       
\usepackage{url}            
\usepackage{booktabs}       
\usepackage{amsfonts}       
\usepackage{nicefrac}       
\usepackage{microtype}      
\usepackage{xcolor}         

\setcitestyle{numbers,square}
\usepackage{paralist}
\usepackage{graphicx}
\usepackage{amsmath}
\usepackage{amsthm}
\usepackage{cleveref}

\newtheorem{assumption}{Assumption}[section]
\newtheorem{theorem}{Theorem}[section]

\usepackage[ruled,vlined,norelsize,\languagename]{algorithm2e}
\SetKwInput{KwInput}{Input}
\SetKwInput{KwOutput}{Output} 

\newcommand{\indep}{\perp \!\!\! \perp}
\usepackage{scalerel}

\title{COCO Denoiser: Using Co-Coercivity for Variance Reduction in Stochastic Convex Optimization}

\usepackage{soul}
\usepackage{marvosym}
\newcommand\markISR{\Leo}
\newcommand\markCMU{\Virgo}

\DeclareMathOperator*{\argmin}{argmin}

\author{%
  Manuel Madeira\textsuperscript{\markISR}
  \And
  Renato Negrinho\textsuperscript{\markCMU}
  \And
  João Xavier\textsuperscript{\markISR}
  \And
  Pedro M. Q. Aguiar\textsuperscript{\markISR}
\AND \\[-4ex]
  {\tt manuel.madeira@tecnico.ulisboa.pt}, 
  {\tt negrinho@cs.cmu.edu}\\
  \{{\tt jxavier}, 
  {\tt aguiar}\}{\tt @isr.tecnico.ulisboa.pt}
\\[2ex]
  \textsuperscript{\markISR{}}Instituto de Sistemas e Rob\'otica, Instituto Superior Técnico, Lisboa, Portugal \\
  \textsuperscript{\markCMU{}}Carnegie Mellon University, Pittsbugh PA, USA
  }

\begin{document}

\maketitle

\begin{abstract}\label{sec:abstract}
First-order methods for stochastic optimization have undeniable relevance, in part due to their pivotal role in machine learning. Variance reduction for these algorithms has become an important research topic. In contrast to common approaches, which rarely leverage global models of the objective function,
we exploit convexity and $L$-smoothness to improve the noisy estimates outputted by the stochastic gradient oracle.
Our method, named COCO denoiser, is the joint maximum likelihood estimator of multiple function gradients from their noisy observations, subject to co-coercivity constraints between them.
The resulting estimate is the solution of a convex Quadratically Constrained Quadratic Problem. Although this problem is expensive to solve by interior point methods, we exploit its structure to apply an accelerated first-order algorithm, the Fast Dual Proximal Gradient method. Besides analytically characterizing the proposed estimator, we show empirically that increasing the number and proximity of the queried points leads to better gradient estimates. We also apply COCO in stochastic settings by plugging it in existing algorithms, such as SGD, Adam or STRSAGA, outperforming their vanilla versions, even in scenarios where our modelling assumptions are mismatched. \footnote{Code for the experiments and plots is available at \textit{https://github.com/ManuelMLMadeira/COCO-Denoiser}.}
\end{abstract}

\section{Introduction}
\label{sec:introduction}

\begin{figure}[tbp]
\centering
\includegraphics[width=\textwidth]{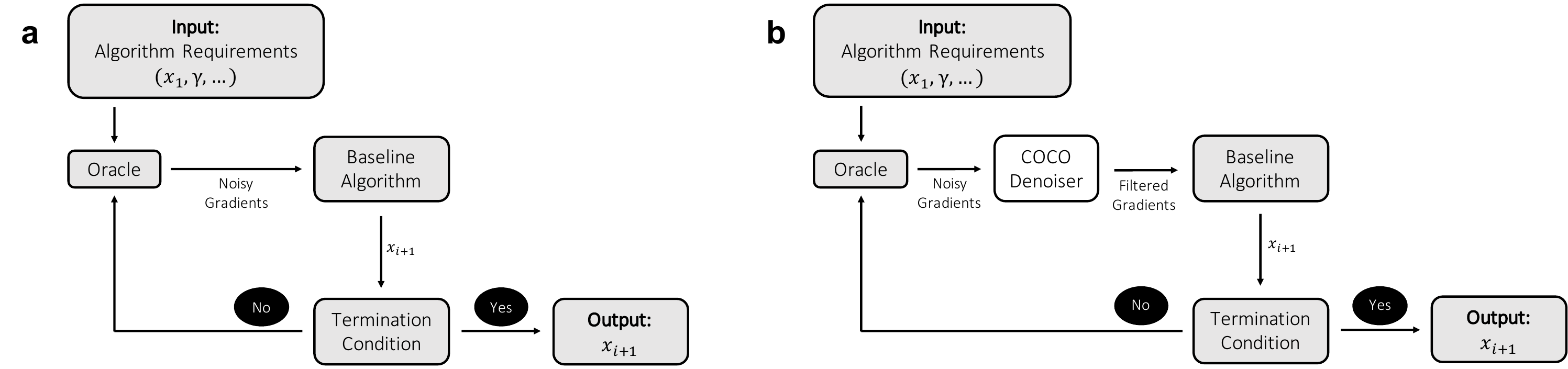}
\caption{(\textbf{a}) Typical workflow for stochastic optimization; (\textbf{b}) using the proposed COCO denoiser as a plug-in.}
\label{fig:COCO_schematic}
\end{figure}

First-order methods for stochastic optimization are commonly used in cases where the exact gradient can not be easily obtained, either due to the computational cost (\textit{e.g.}, in large-scale machine learning problems, where the computation of an exact gradient requires a full pass over the entire dataset), or due to the intrinsic nature of problem (\textit{e.g.}, in streaming applications, where only noisy versions of the gradient are available~\citep{jothimurugesan_variance-reduced_2018}). Their widespread use motivated the optimization community to develop variance reduction approaches, which rarely exploit global models for the objective function, $f: \mathbb{R}^d \to \mathbb{R}$. In contrast, we leverage on the convexity and $L$-smoothness of $f$, which are assumptions that, although commonly used for the analysis of convex optimization algorithms, have been left out of algorithm design for denoising stochastic gradients.
These properties can be merged into the so-called \textit{gradient co-coercivity}, which we exploit to denoise a set of gradients $g_1, \ldots, g_k$, obtained from a first-order stochastic oracle~\citep{bubeck_convex_2015} consulted at iterates $x_1, \ldots, x_k$, respectively. We refer to our method as the co-coercivity (COCO) denoiser and plug it in in existing stochastic first-order algorithms (see \Cref{fig:COCO_schematic}), showing that it leads to reduced variance across all cases (see Figure~\ref{fig:synthetic_dataset_SGD&Adam_Performance}).
We are inspired by applications where the oracle queries are very expensive, therefore measuring the progress of different algorithms as a function of the number of gradient evaluations.

We formulate the denoising problem as the joint maximum likelihood estimation of a set of gradients from their noisy observations, constrained by the pairwise co-coercivity constraints (see Section~\ref{subsec:max_likelihood}). The noise is assumed to be zero mean Gaussian and independent across oracle queries. This estimator can be expressed as convex quadratically constrained quadratic problem (QCQP)~\citep{boyd_convex_2004}, which can be solved by available optimization packages, \textit{e.g}, CVX~\citep{grant_cvx_2014}. Unfortunately, CVX quickly becomes unable to solve problems with large dimension $d$ and number of gradients $k$ due to the computational complexity scaling of the general algorithm used by the solver. For this reason, we exploit the structure of the estimation problem to apply the Fast Dual Proximal Gradient (FDPG) method~\citep{beck_fast_2014}, which yields an algorithm that is able to compute approximate solutions in reasonable time (see Section~\ref{section:Solution_Methods_QCQP} and Appendix~\ref{sec:detailed derivation FDPG}). 
As the number of co-coercivity constraints increases quadratically with the number of gradients simultaneously estimated, 
we also consider denoising a fixed number $K$ of last visited points, $x_{k-K+1}, \ldots, x_k$. We refer to this estimator as COCO$_K$.
Our experiments show that variance monotonically decreases with the number of gradients used, providing a natural way to trade off variance reduction and computation (see Section~\ref{sec:experimental_analysis}). 


By theoretically analysing COCO, we provide insight about the estimator results based on its inputs. Moreover, the standard analysis based on denoising by projecting into the feasible set gives us a statement about error reduction jointly for all the estimated gradients involved (see Section~\ref{sec:estimator_properties_theoretical}). While this statement does not provide insight about the error reduction for each individual gradient (\textit{e.g.}, the error of some gradients might worsen when compared to the noisy oracle estimate), we explore this question empirically and observe it not to be the case, with the amount of error reduction for each gradient being related with the tightness of the COCO constraints (see Section~\ref{sec:estimator_properties} and Appendix~\ref{sec:app_estimator_properties}). This COCO error reduction is shown to explicitly imply a variance reduction of the gradient estimates in comparison to the ones directly provided by the oracle. We also theoretically (for a simple scenario) and empirically verify that this tightness is increased when considering closer iterates, smaller noise magnitude, or better estimated Lipschitz constants of the objective gradient. Namely, for sufficiently close iterates, the COCO denoiser remarkably recovers the typical error reduction rate associated with the averaging of random variables following Gaussian distributions (see Figure~\ref{fig:MSE_totalVsVariance}).

To evaluate the impact of the proposed denoiser in stochastic optimization, we consider two scenarios (see Section~\ref{section:stochastic_optimization}), using: \textit{i)} synthetic data that follow the assumptions underlying the design of COCO and \textit{ii)} readily available datasets \citep{chang_libsvm_2011} for an online learning task (logistic regression~\citep{boyd_convex_2004}), in which the gradient noise does not satisfy those assumptions, therefore testing the robustness of our approach.
Our experiments show that COCO leads to improved performance in the variance regime, when plugged in existing algorithms such as SGD~\citep{robbins_stochastic_1951}, Adam~\citep{kingma_adam_2015}, and STRSAGA~\citep{jothimurugesan_variance-reduced_2018}.

\section{Related Work}
\label{sec:related_work}
 
The machine learning explosion of the last few years has strongly contributed to the fast development of stochastic optimization. In the core of this field, we find SGD~\citep{robbins_stochastic_1951}, which, despite its simplicity, remains a fundamental algorithm as its widespread uses. The objective function $f$ might be unknown as long as a stochastic first-order oracle (which provides noisy gradient estimates, $g_k$) can be queried. Based on those queries, the iterates are $x_{k+1} = x_k - \gamma_k \; g_k$, where $\gamma_k$ denotes the step size. The analysis of convergence of this method\footnote{In this paper, our chief interest lies in $E[||x_k - x^*||]$, where $x^*$ denotes the minimizer of the objective function, even though it is also often found in literature to be $E[||f(x_k) - f(x^*)||]$.}, considers two terms: \textit{i)} the bias term, which represents the dependence of the convergence on the initial distance to the optimum (\textit{e.g.}, $\|f(x_0)-f(x^*)\|$), and \textit{ii)} the variance term, which represents the dependence of the convergence on the noise of the oracle itself. Although the bias term vanishes under a convenient selection of a fixed step size, that does not happen to the variance term. For this reason, when the former becomes negligible, the algorithm gets stuck, originating random iterates within a "ball of uncertainty". 

\paragraph{Decreasing step sizes} The natural approach to progressively reduce that variance ball and, therefore, enable convergence of SGD is by using a diminishing step size, \textit{i.e,}, by selecting at iteration $k$ a step size $\gamma_k = {C}/{k}$, where $C$ is a problem dependent constant. Using this strategy, we attain the asymptotically optimal convergence rates when $\mathbb{E}[\scalerel*{\widehat{\nabla f}(x)}{j^2}] = \nabla f(x)$ (\textit{e.g.}, whenever the noise affecting the gradients is additive and zero mean)~\citep{nemirovsky_problem_1983, agarwal_information-theoretic_2012} of $O(1/\smash{\sqrt{k}})$ for non-strongly convex objectives. Nonetheless, besides not being robust to the choice of $C$, this strategy also implies an excessively conservative step size decay rate of $O(1/k)$. 

We approach variance reduction by addressing the stochasticity of the oracle itself in a short horizon of queries, while the decreasing step size implies an infinite horizon scheduling of the step size. For this reason, the step size tuning is an orthogonal direction to ours and we consider the step size to be fixed. Therefore, we are interested in showing that for any given step size, the SGD performance is enhanced when coupled to the COCO denoiser. We also show that when we tune the step size of SGD in order to have the same variance regime as the SGD coupled to COCO, the latter presents a better bias regime (see \Cref{fig:COCOvstunedSStovariance} in \Cref{sec:appendix_comparison_other_VR}). 


\paragraph{Averaging algorithms} 
This class of algorithms was introduced with the so-called Polyak-Ruppert (PR) averaging~\citep{polyak_acceleration_1992}, which instead of simply taking into account the convergence of the iterates $x_k$, considers the sequence $\Bar{x}_k = 1/(k+1) \;\smash{\sum_{j=0}^k} x_j$. This scheme reduces variance  and improves on the diminishing step size strategy by allowing us to use step sizes with slower decays (i.e., $O(1/k^{\alpha}), 1/2 \leq \alpha \leq 1$~\citep{moulines_non-asymptotic_2011}) or, under some additional conditions, even fixed step sizes ($\alpha = 0$). This is, for example, the case of the least-squares regression problem, where PR averaging with fixed step size was shown to accelerate the convergence rate to $O(1/k)$~\citep{bach_non-strongly-convex_2013}. Furthermore, the utility of averaging algorithms is reinforced in the strongly convex case, since by plugging it in along with momentum in a regularized least-squares problem~\citep{dieuleveut_harder_2017}, it led to the first algorithm achieving $O(1/k^2)$ on the bias term without compromising the optimal $O(1/k)$ of the variance term~\citep{tsybakov_optimal_2003}. More recently, more refined averaging schemes have been proposed, e.g., tail-averaging approach in~\citep{jain_accelerating_2018}.

Although the averaging algorithms can be updated online, the PR averaging technique ends up requiring an infinite horizon averaging of all the visited iterates. Nevertheless, we experimentally show that COCO coupling to SGD outperforms the PR averaging in a quadratic function for a fixed step size as long as the amount of points used by both strategies is the same (i.e. when COCO uses all the points visited) - see \Cref{fig:COCOvstunedSStovariance} in \Cref{sec:appendix_comparison_other_VR}.

\paragraph{Adaptive algorithms} 
The adaptive (step size) algorithms (e.g., AdaGrad~\citep{duchi_adaptive_2011}, Adadelta~\citep{zeiler_adadelta_2012}, RMSprop~\citep{hinton_lecture_2012}, Adam \citep{kingma_adam_2015}, AdaMax \citep{kingma_adam_2015} or  Nadam~\citep{dozat_incorporating_2016}), despite not contributing to variance reduction, address the difficulties of tuning $C$. These algorithms generically receive as input an initial step size and then adjust it successively in each dimension according to the magnitude of the progress in that same dimension. Therefore, this approach drastically improves the performance over SGD, namely in problems with high condition number. Among these algorithms, typically the most used is Adam, being particularly popular in training deep neural networks. Therefore, we consider Adam to be the representative of this class of algorithms.

\paragraph{Variance-reduced methods and STRSAGA}

In the last decade, with the rise of objective functions that can be decomposed as a finite sum of functions (a typical setting in machine learning problems), the research community turned their attention to the so-called variance-reduced (VR) methods. These approaches made it possible to close the gap between the asymptotic convergence rate gap of the deterministic and stochastic setting by using the oracle call to update a full-gradient estimate instead of greedily following the noisy gradient received. Some of these approaches are SAG~\citep{schmidt_minimizing_2017}, SAGA~\citep{defazio_saga_2014} or SVRG~\citep{johnson_accelerating_2013}. Although all these algorithms have \textit{a priori} access to the whole dataset, some streaming VR inspired alternatives have emerged, such as SSVRG~\citep{frostig_competing_2015} and STRSAGA~\citep{jothimurugesan_variance-reduced_2018}. Considering the superior empirical results of the latter, we consider STRSAGA to be the representative of the variance-reduced based online algorithms.

\section{COCO Denoiser}
\label{sec:COCO}

This section describes our approach. First, we formulate COCO as the Maximum Likelihood estimator constrained by the co-coercivity conditions; then, we propose efficient methods to compute its solution.

\subsection{Maximum Likelihood Estimation}\label{subsec:max_likelihood}

Let the objective function $f: \mathbb{R}^d \to \mathbb{R}$ be convex and $L$-smooth.
A standard result in convex analysis states that the gradient of $f$ is co-coercive~\citep{boyd_convex_2004}, which is expressed as
\begin{equation}
    \begin{aligned}\label{eq:co-coercivity}
        &\forall x,y \in \mathbb{R}^n: \quad \frac{1}{L} \|\nabla f(y) -  \nabla f(x)\|^2 \leq \langle \nabla f(y) -  \nabla f(x),\;  y-x \rangle . 
    \end{aligned}
\end{equation}

Our approach hinges on the following assumptions.

\begin{assumption}
\label{ass:L_constant}
    A Lipschitz constant $L$ for the gradient of $f$ is known.
\end{assumption}

This assumption prevents the gradients from changing in arbitrarily fast from one point to another. It is commonly adopted in stochastic optimization for algorithm analysis as it applies to a large class of functions. Along with strong convexity, these are the most common structural properties (quadratic upper and lower bound, respectively) in convex analysis. In fact, typical high-dimensional machine learning problems have correlated variables, yielding non-strongly convex objective functions~\citep{bach_non-strongly-convex_2013}. 
By additionally considering that estimating the parameter from $L$-smoothness ($L$) is often easier than estimating the one from strong convexity, we emphasize the pertinence of this assumption.

\begin{assumption}
\label{ass:gaussian_noise}
    There is access to an oracle which, given an input $x \in \mathbb{R}^d$, outputs a noisy version of the gradient of $f$ at $x$: $g(x; w) = \nabla f(x) + w$, where $w \in \mathbb{R}^d$ is a sample of a zero mean Gaussian distribution, $w \sim \mathcal{N}(0,\Sigma)$,
    with known covariance $\Sigma = \sigma^2  I$. The noise samples are independent across the oracle queries.
\end{assumption}

The main motivation for this simple noise model comes is its mathematical convenience. Moreover, as stated by the Central Limit Theorem, the sum (or mean) of independent and identically distributed random variables with finite variances converges to a normal distribution as the number of variables increases. In machine learning, the mini-batch scheme is a common procedure to obtain gradient estimates at a point, which can be defined as a mean of independent random variables.

The oracle is consulted at points $x_1, \ldots, x_K$, returning the data vector $g = [g_1,\; \ldots,\; g_K]^T \in \mathbb{R}^{Kd}$, from which our goal is to estimate the true gradients $\nabla f(x_1),\; \ldots,$ $\nabla f(x_K)$, arranged in the parameter vector $\theta = [\theta_1,\; \ldots,\; \theta_K]^T \in \mathbb{R}^{Kd}$, where $\theta_k = \nabla f(x_k) \in \mathbb{R}^d$. 

From \Cref{ass:gaussian_noise}, the observation model is immediate:
\begin{equation*}\label{eq:g=theta+w}
    g = \theta + w, 
\end{equation*}
where $w = [w_1,\; \ldots,\; w_K]^T \sim \mathcal{N}(0,\Sigma_w)$, with $\Sigma_w$ block-diagonal, each block being $\Sigma$. 

The Maximum likelihood estimate \cite{vershynin_estimation_2015} of $\theta$ is then
\begin{equation*}
    \hat{\theta} = \underset{\theta \in \Theta}{\operatorname{argmax}} \;\; p(g|\theta)  ,\qquad \text{with} \qquad p(g|\theta) = \frac{1}{\sqrt{(2\pi)^{n}|\Sigma|}}e^{-\frac{1}{2}(g-\theta)^T\Sigma_w^{-1}(g-\theta)} ,
\end{equation*}
where the parameter vector $\theta$ is constrained by the co-coercivity condition (\ref{eq:co-coercivity}), \textit{i.e.},
\begin{equation*}
\begin{aligned}
     \theta \in \Theta = \left\{ (\theta_1, \ldots, \theta_K): \; \frac{1}{L} \| \theta_m - \theta_l \|^2 \leq \langle \theta_m - \theta_l, \; x_m - x_l \rangle, \; 1 \leq m < l \leq K\right\} . 
\end{aligned}
\end{equation*}

Consequently, the maximum likelihood estimate $\smash{\hat{\theta}}$ comes from solving the following optimization problem:
\begin{equation}\label{eq:COCO_formalization}
    \begin{aligned}
    & \underset{\theta_1, \ldots, \theta_K}{\text{minimize}}
    & & \sum_{i=1}^{K} \left\| g_i - \theta_i \right\|^2  \\
    & \text{subject to}
    & & \frac{1}{L} \| \theta_m - \theta_l \|^2 \leq \langle 
    \theta_m - \theta_l, x_m - x_l \rangle, \; 1 \leq m < l \leq K.
    \end{aligned}
\end{equation}
The solution does not depend on $1/\sigma^2$ and therefore, it is omitted from \Cref{eq:COCO_formalization}. From this observation, it is clear that knowing $\sigma^2$ is not a requirement.
Since both the objective and the constraints in (\ref{eq:COCO_formalization}) are convex quadratics, the resulting problem is a convex Quadratically Constrained Quadratic Problem (QCQP)~\citep{boyd_convex_2004}.
%
%
Since there is one constraint for each pair of query points, the total number of constraints in (\ref{eq:COCO_formalization}) is ${K(K-1)}/{2}$. This quadratic growth motivates an approach where we keep only the $K$ last query points ($1 < K \leq k$). For example, for $K=2$, the denoiser works only with $x_{k-1},\; x_{k}, \;g_{k-1}\; \text{and}\; g_{k}$. We define COCO$_K$ to be the denoiser that uses a window of length $K$.


\subsection{Efficient Algorithms for COCO\texorpdfstring{$\boldsymbol{_K}$}{K} }\label{section:Solution_Methods_QCQP}

We first present the closed-form solution for the optimization problem (\ref{eq:COCO_formalization}) for $K=2$; then, we propose an iterative method to efficiently compute its approximate solution for arbitrary $K$.

\begin{algorithm}[tbp]\label{algorithm:FDPG}
\caption{FDPG for the COCO denoiser}
    \DontPrintSemicolon
    \KwInput{Initial Point: $s_0$; Number of steps: $T$; $L$-Smoothness Constant: $L_{p^*}$; Momentum Auxiliary Iterate: $y_0 = s_0$; Initial Momentum Constant: $t_1 = 1$}
   \For{$k = 1, \ldots, T$}{
    $s_k = \operatorname{prox}_{\frac{1}{L_{p^*}} q^*}\left(y_{k-1} - \frac{1}{L_{p^*}} \nabla p^*(-A^T y_{k-1})\right)$\;
    $t_k = \frac{1+\sqrt{1+4t_{k-1}^2}}{2}$\; 
    $y_k = s_k + \frac{t_{k-1}-1}{t_{k}} (s_k - s_{k-1})$\;
    } 
    \KwOutput{Final Point: $s_T$}
\end{algorithm}


\paragraph{Closed-form solution for COCO$_2$} The result, obtained by instantiating the Karush-Kuhn-Tucker conditions for the QCQP \eqref{eq:COCO_formalization}, is given by the following theorem, proved in \Cref{section:proof_theorem_closedformK=2}.
\begin{theorem}\label{theorem:closed_form_COCO_2}
    For $K=2$, the solution to the optimization problem~(\ref{eq:COCO_formalization}) is given by:
    
    If $\; \left\lVert g_{1} - g_{2} \right\rVert^2 \leq L \;\langle g_1 - g_2, \;x_{1}-x_{2} \rangle$,
        \begin{equation*}\label{eq:COCO_closed_form_K=2}
            \begin{cases} 
                \hat{\theta}_1 = g_1 \\
                \hat{\theta}_2 = g_2 ;
             \end{cases}
        \end{equation*}
        If $\; \left\lVert g_{1} - g_{2} \right\rVert^2 > L \;\langle g_1 - g_2, \;x_{1}-x_{2} \rangle$,
        \begin{equation*}
            \begin{cases} 
                \hat{\theta}_1 = \frac{g_1 + g_2 + \frac{L}{2}(x_1-x_2)}{2}  + \| \frac{L}{4}(x_1-x_2) \| \frac{g_1-g_2 - \frac{L}{2}(x_1-x_2)}{ \left\lVert g_1-g_2 - \frac{L}{2}(x_1-x_2) \right\rVert } \\
                \hat{\theta}_2 = \frac{g_1 + g_2 - \frac{L}{2}(x_1-x_2)}{2}  - \| \frac{L}{4}(x_1-x_2) \| \frac{g_1-g_2 - \frac{L}{2}(x_1-x_2)}{ \left\lVert g_1-g_2 - \frac{L}{2}(x_1-x_2) \right\rVert } . 
             \end{cases}
        \end{equation*}
\end{theorem}

The solution above has an intuitive interpretation: when the observed gradients are co-coercive ($\| g_1 - g_2 \|^2 \leq L \langle g_1 - g_2,\;x_1 - x_2\rangle$), they are on the feasible set of the problem, so they coincide with the estimated ones; when they are not co-coercive ($\| g_1 - g_2 \|^2 > L \langle g_1 - g_2,\;x_1 - x_2\rangle$), their difference is orthogonally projected onto the feasible set, which is a ball.
Despite its simplicity, this closed-form solution is of the utmost relevance in practice, since our experiments show that COCO leads to significant improvements in stochastic optimization, even for this simple case of  $K=2$.

\paragraph{Efficient solution for COCO$_K$, $\boldsymbol{K\geq3}$}
Packages like \textit{CVX}~\citep{grant_cvx_2014} can solve a wide range of convex problems, including the QCQP in (\ref{eq:COCO_formalization}), but their generality prevents the exploration of the specific structure of the  problem at hand.
%
%
Therefore, we present a first-order algorithm which explores the COCO structure\footnote{A more detailed derivation of the method is provided in \Cref{sec:detailed derivation FDPG}.}. The dual problem of the QCQP in (\ref{eq:COCO_formalization}) can be shown to be
\begin{equation}\label{eq:COCO_dual_reformulation}
    \begin{aligned}
    & \underset{s}{\text{minimize}}
    & & \underbrace{\frac{1}{2} \| - A^T s \|^2}_{p^*(-A^Ts)} 
     + \underbrace{ \frac{ }{ }\sum_{1\leq m < l \leq K} r_{ml} \|s_{ml}\| - s_{ml}^Tc_{ml} }_{q^*(s)} ,
    \end{aligned}
\end{equation}
where $s = [s_{12},\; s_{1 3},\; \ldots, \;s_{1 K},\;s_{2 3},\; s_{2 K},\; \ldots,\;s_{K-1 K}]^T$ is the dual variable,
$A$ is a structured matrix, $c_{ml} = \left(g_m - ({L}/{2})\; x_m\right) - \left(g_l - ({L}/{2})\; x_l\right)$ and $r_{ml} = {L}\|x_m - x_l\|/{2}$.

The first term in (\ref{eq:COCO_dual_reformulation}), $p^*(-A^Ts) = 1/2 s^T A A^T s$, is differentiable, with $\nabla_s \; p^*(-A^Ts) =  AA^Ts$. Note that $p^*(-A^Ts)$ is necessarily $L$-smooth, with a Lipschitz constant of $L_{p^*} = \sigma_{\max}^2(A)$. On the other hand, despite its non-differentiability, a proximity operator can be efficiently computed for the second term, $q^*(s)$: $\operatorname{prox}_{\mu q^*}(s) = s - \mu(v_{\text{proj}}-c)$, where $v_{\text{proj}} = \argmin_{v\in\mathcal{B}}\| v-(c+{s}/{\mu}) \|^2$ collects the projections of $c_{ml} + {s_{ml}}/{\mu}$ onto the ball $\mathcal{B}(0,\; r_{ml}) = \{ x\in\mathbb{R}^n: \|x \| \leq r_{ml} \}$.


Hence, we can use the Fast Dual Proximal Gradient (FDPG) method~\citep{beck_fast_2014}. This approach consists of applying FISTA to the dual problem of the original one. Since FDPG is a first-order method with a low cost per iteration, we obtain a computationally efficient solution for COCO. After computing an approximate solution for the dual problem, $s^*$, we can easily recover the primal solution for the QCQP:  $\smash{\hat{\theta}} = -A^T s^* + g$. The FDPG method for the COCO denoiser is summarized in Algorithm \ref{algorithm:FDPG}.

\paragraph{Convergence rate and memory requirements}

Although the rate of convergence of the dual objective function sequence is $O\left({1}/{k^2}\right)$ (the convergence rate of FISTA~\citep{beck_fast_2009}), it can be shown that the rate of convergence of the primal problem is $O\left({1}/{k}\right)$~\citep{beck_fast_2014}. From this, it is immediate that the iteration complexity\footnote{We refer to iteration complexity as the smallest $k$ verifying the inequality $E\left[ f(x_k)-f(x^*) \right] < \epsilon$.} of our method is $O(1/\epsilon)$; since the iteration runtime is $O(K^2 d)$, the cost of computing the proximity operator, the total algorithmic runtime
is $O(K^2 d/\epsilon)$.
%
Naturally, for $K=2$, the closed-form solution for COCO$_2$ in \Cref{theorem:closed_form_COCO_2} is preferable to the FDPG method, since it has a total runtime of $O(d)$ and lower memory requirements (the closed-form solution only requires two points and respective gradients, both $d$-dimensional vectors, to be kept in memory, an $O(Kd)$ memory overhead, while for the FDPG just the storage of $A$ lead to a $O(K^2d^2)$ overhead). 

\begin{figure}[tbp]
\centering
\includegraphics[width=0.32\textwidth]{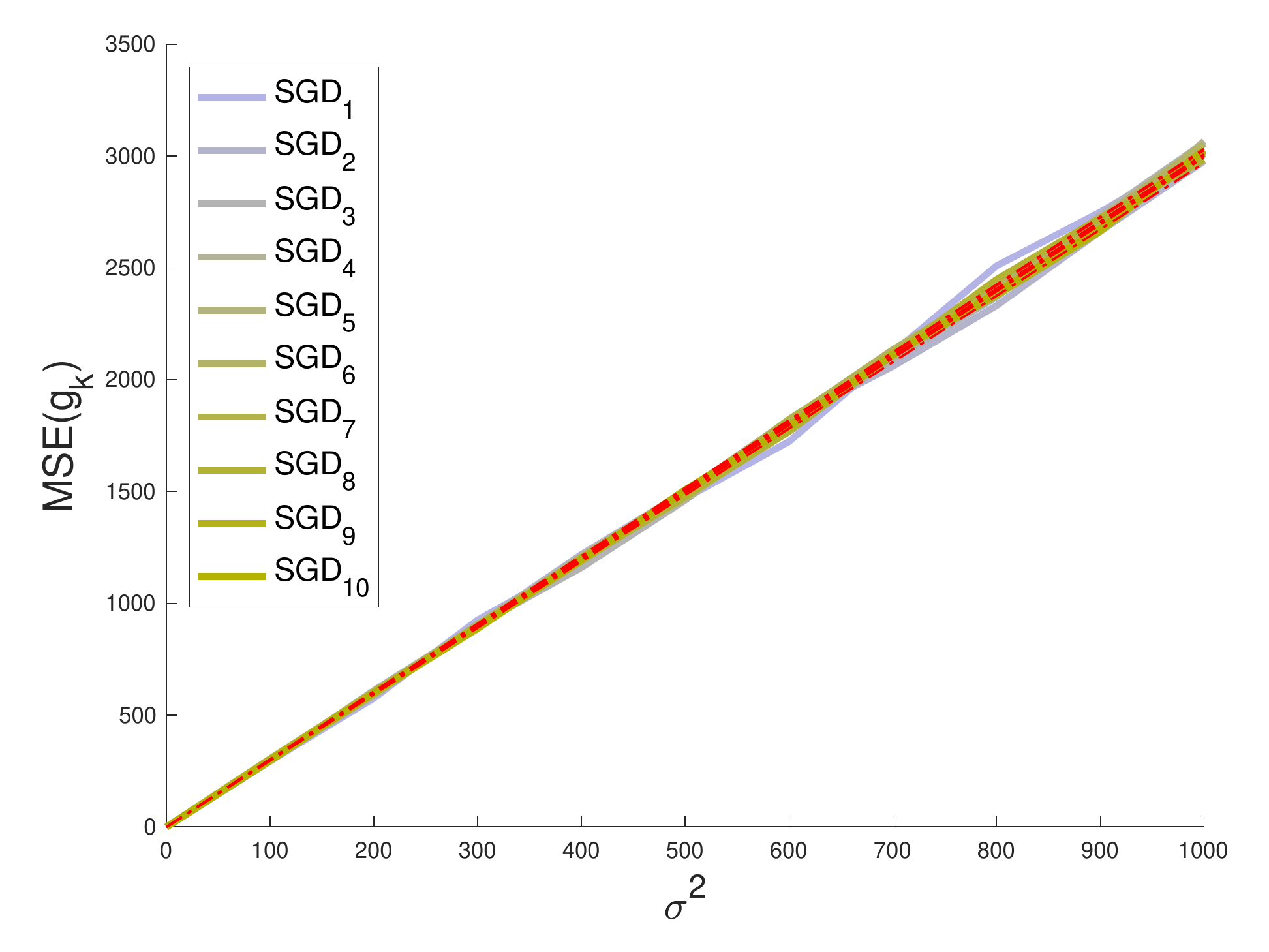}
\includegraphics[width=0.32\textwidth]{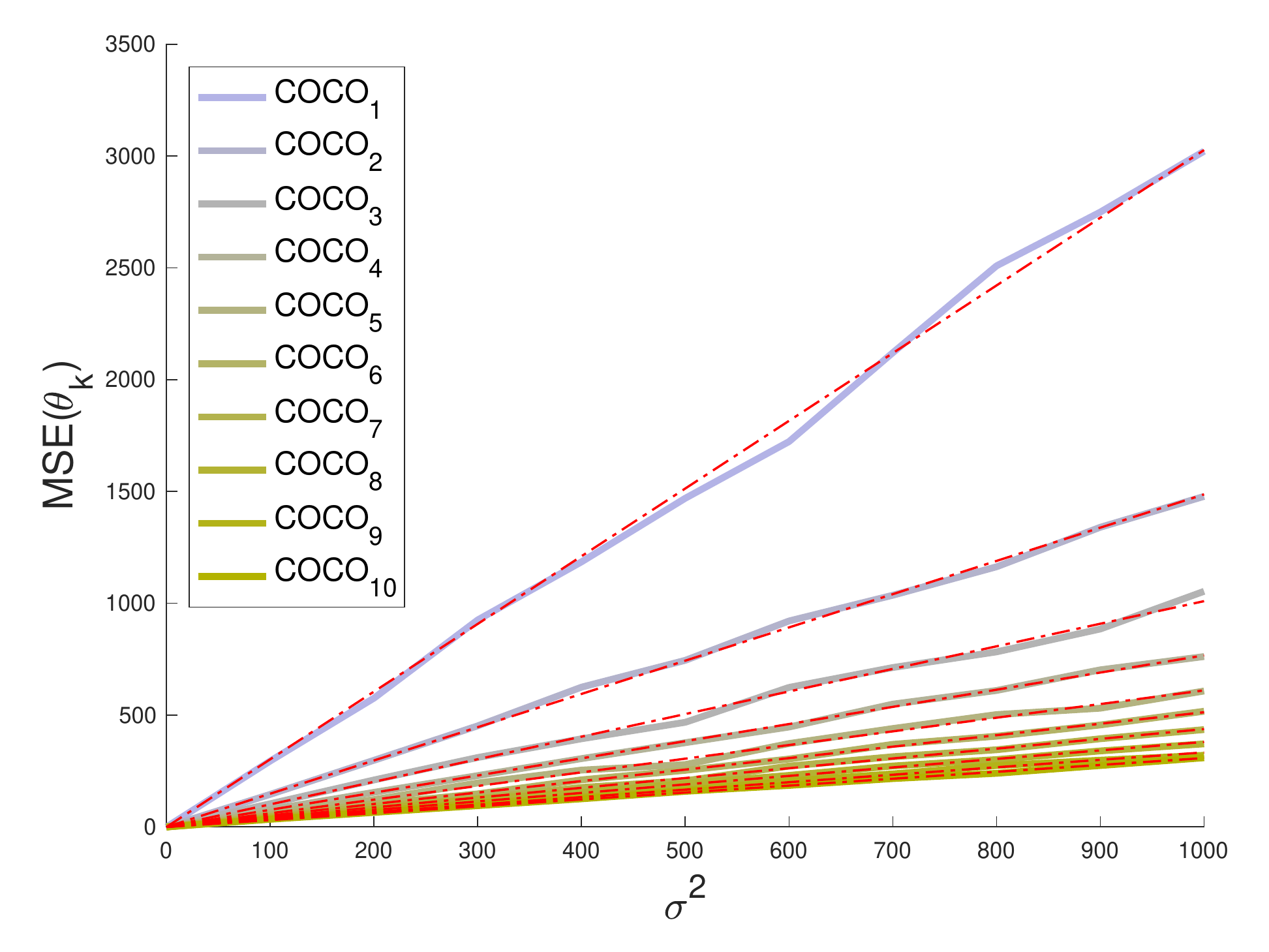}
\includegraphics[width=0.32\textwidth]{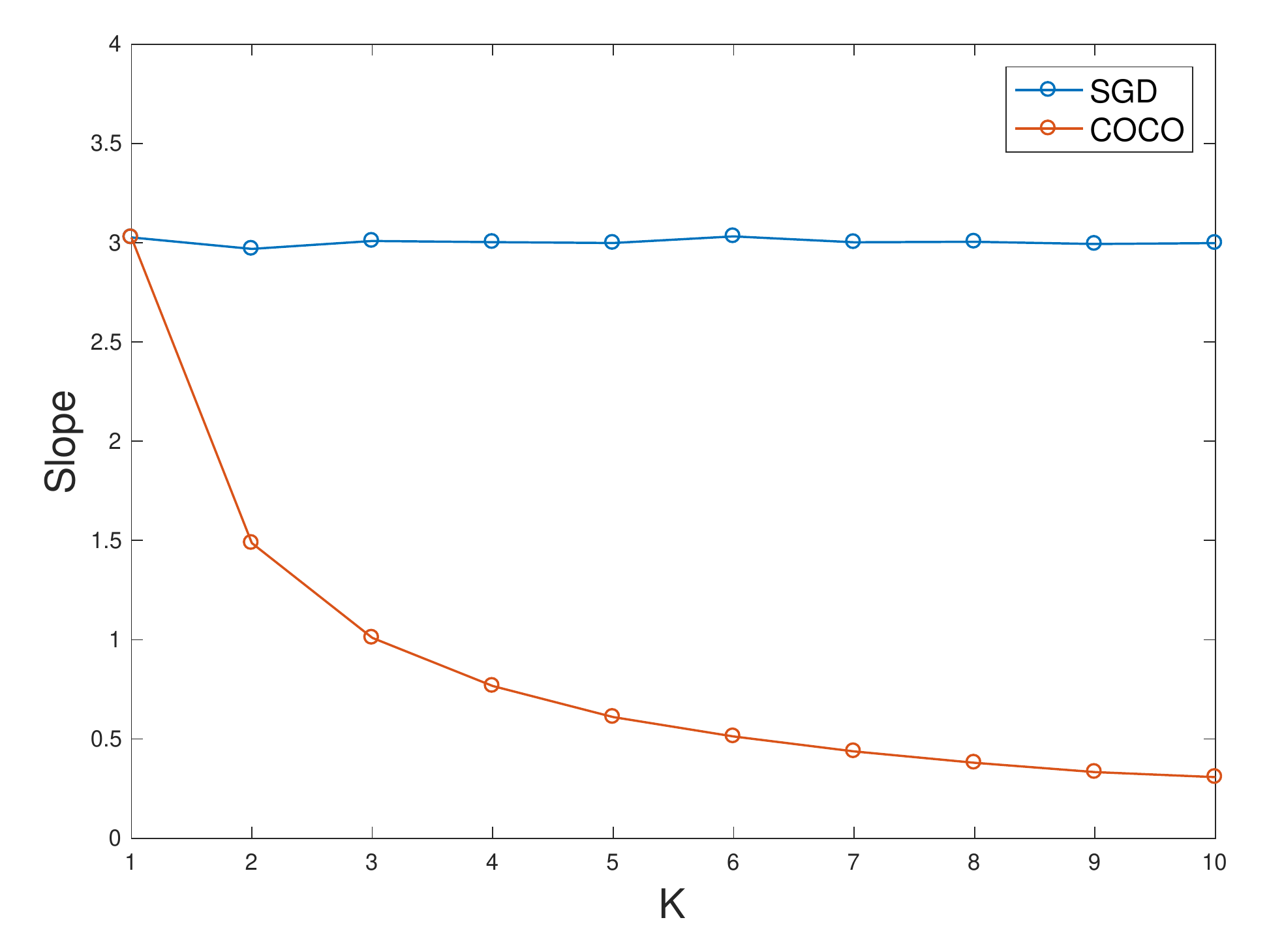}
\caption{Measuring the amount of noise reduction as a function of the noise level. $\operatorname{MSE}(g_k)$ (\textit{Left}) and $\operatorname{MSE}(\smash{\hat{\theta}_k})$ (\textit{Center}), estimated via Monte-Carlo method, as functions of the noise variance $\sigma^2$, for several numbers of points considered ($1 \leq K \leq 10$). For each simulation, a different set of $K$ points is sampled from an uniform distribution over the cube $x_k \in  [-5,\; 5]^3$. We consider an Hessian with linearly-spaced eigenvalues between $1$ and ${1}/{3}$ and $N=1000$. The dashed-dotted red lines result from linear regressions with intercept fixed at $0$. Number of Monte-Carlo simulations: $N=1000$. On the \textit{right}, each of the regressed slopes is depicted as a function of the number of points.}
\label{fig:MSE_totalVsVariance}
\end{figure}

\section{COCO Estimator Properties}\label{sec:estimator_properties}

In this section, we prove a relationship between the COCO output and its input, prove that the COCO gradient estimates jointly outperform the oracle, and show that the co-coercivity constraints become tighter for closer points. We then present empirical evidence on COCO outperforming the oracle in terms of element-wise gradient estimation and on the $O(1/K)$ noise reduction rate achieved by COCO for sufficiently close points.

\subsection{Theoretical Analysis}\label{sec:estimator_properties_theoretical}

\paragraph{Relation between the oracle and COCO estimates} We start by relating the centroid of the noisy gradients (COCO input) with the centroid of the denoised ones (COCO output), via the following theorem, which, as proven in the \Cref{sec:proof_sum_thetas_equal_sum_gs}, holds for generic $\Sigma$. 
\begin{theorem}\label{theo:sum_thetas_equal_sum_gs}
    The gradients estimated by COCO and the raw observations have the same centroid:
    \begin{equation*}
    \frac{1}{K} \sum_{i=1}^{K} \hat{\theta}_i = \frac{1}{K} \sum_{i=1}^{K} g_i.
\end{equation*}
\end{theorem}

\paragraph{Mean squared error (MSE) of COCO estimates} The following theorem, also proved in the \Cref{sec:proof_Total_MSE_COCO<Total_MSE_Raw}, states that the COCO estimator outperforms the oracle in terms of MSE ($\operatorname{MSE}(\smash{\hat{\theta}})\!=\!E\left[||\smash{\hat{\theta}} - \nabla f||^2\right]$, $\operatorname{MSE}(g)\!=\!E\left[||g - \nabla f||^2\right]$, with
 $\nabla f$ collecting the gradients $\nabla f(x_k)$).
\begin{theorem}\label{theorem:Total_MSE_COCO<Total_MSE_Raw}
    The following inequality holds:
        \begin{equation}
            \operatorname{MSE}(\hat{\theta}) \leq \operatorname{MSE}(g).
        \end{equation}
\end{theorem}

\paragraph{COCO constraints tightness} Each constraint in COCO involves a pair of  gradients, $g_i$ and $g_j$. If they are not co-coercive, COCO outputs co-coercive estimates $\smash{\hat{\theta}}_{i}$ and $\smash{\hat{\theta}}_{j}$.
It is thus interesting to know how often $g_i$ and $g_j$ do not respect the co-coercivity constraint. 
Analysing a one-dimensional setup (detailed in \Cref{sec:app_coco_tightness}), we conclude that the co-coercivity constraint becomes ``looser" (\textit{i.e.}, the probability of $g_i$ and $g_j$ being co-coercive increases) as the distance between $x_i$ and $x_j$ increases. We also observe that the more the Lipschitz constant $L$ is overestimated, the looser the co-coercivity constraint becomes. In \Cref{sec:MSEvsDistance}, we experimentally extend this result by observing that the constraint looseness implies a worse denoising capability from COCO.

\subsection{Experimental Analysis}\label{sec:experimental_analysis}



\begin{figure}[tbp]
\centering
\includegraphics[width=0.32\textwidth]{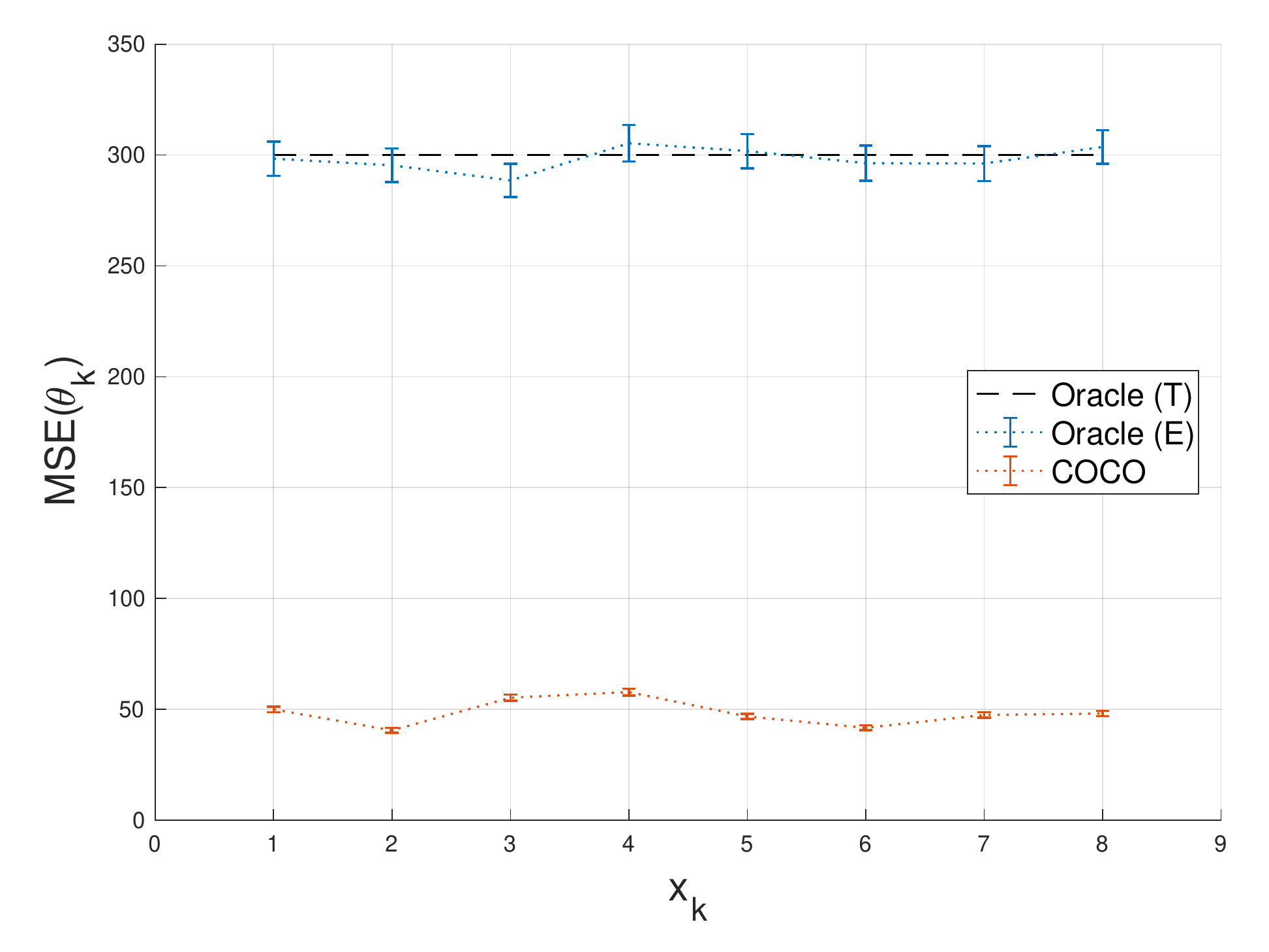}
\includegraphics[width=0.32\textwidth]{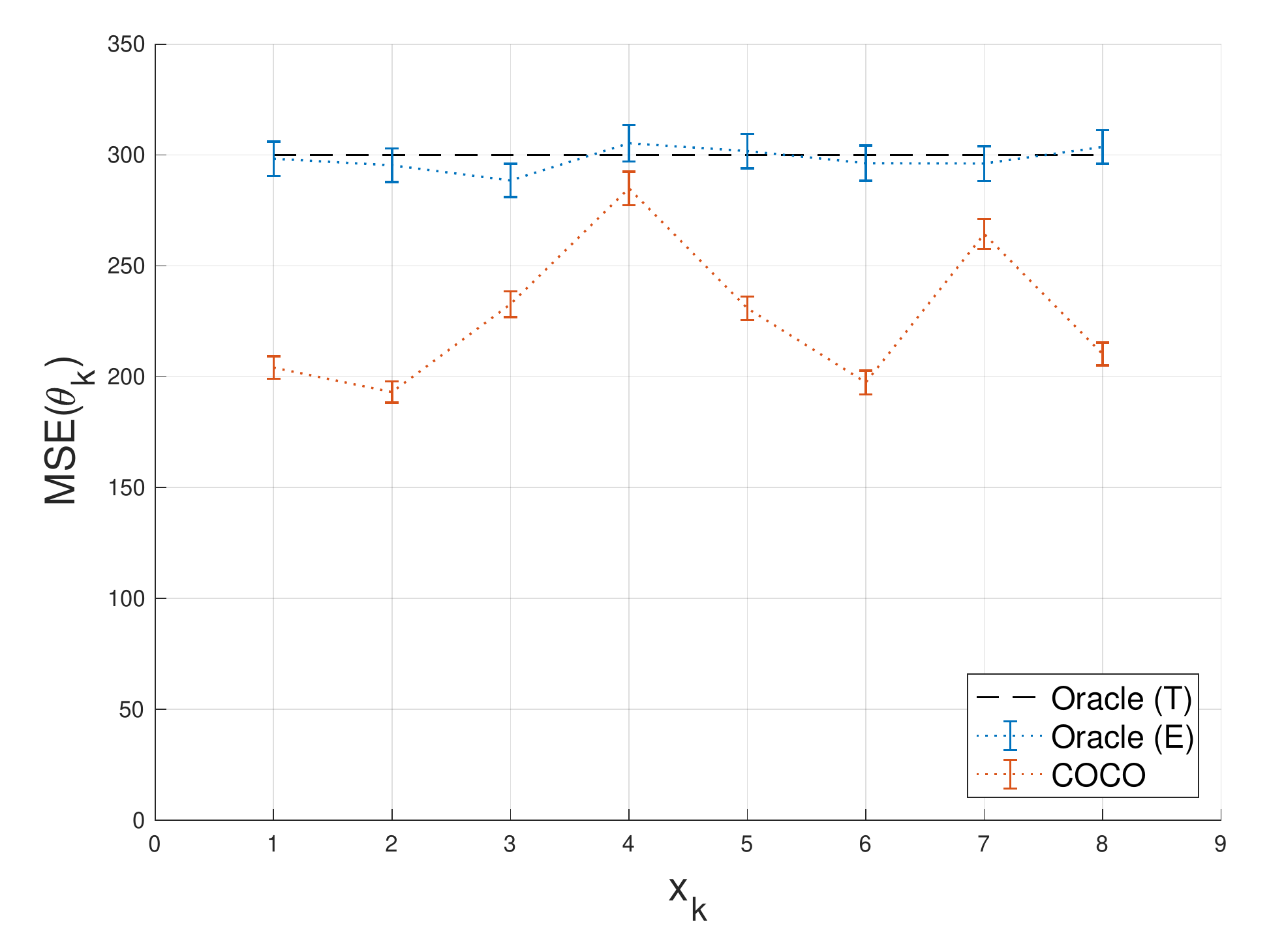}
\includegraphics[width=0.32\textwidth]{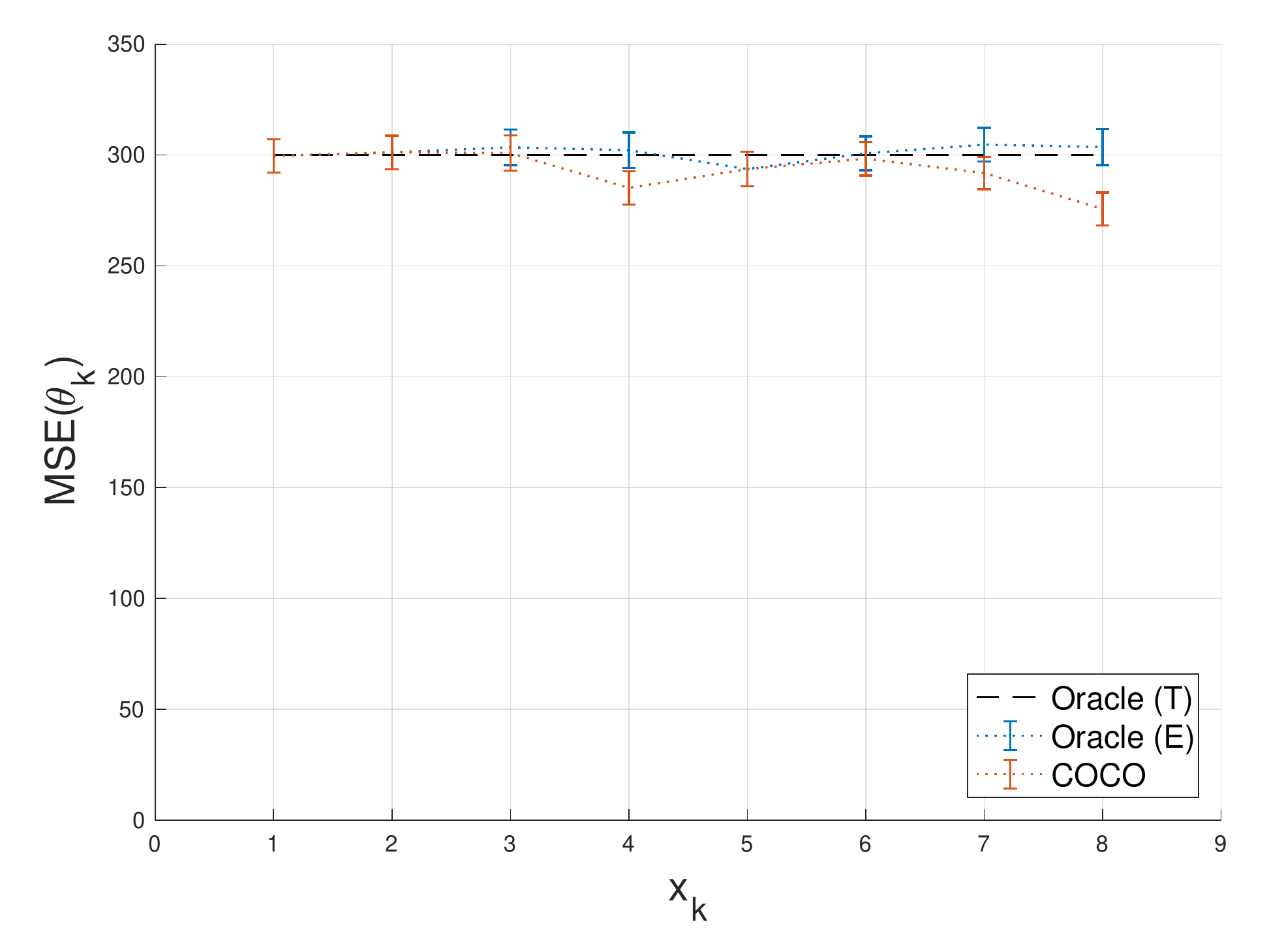}
\caption{Measuring the amount of noise reduction as a function of proximity of the points. $\smash{\widehat{\operatorname{MSE}}}(\hat{\theta}_k)$ (COCO), $\smash{\widehat{\operatorname{MSE}}}(g_k)$ (Oracle (E)), and $\operatorname{MSE}(g_k)$ (Oracle (T)) for 8-point configuration in $\mathbb{R}^3$, each point is sampled from an uniform distribution in a cube centered at the origin with edge length $2l$, \textit{i.e.}, $x_k \in [-l,\; l]^3$. \textit{Left}: $l = 10$; \textit{Center}: $l=100$; \textit{Right}: $l=1000$. We consider a quadratic objective whose eigenvalues of the Hessian are chosen to be linearly spaced between $1$ and ${1}/{3}$ and $\sigma = 10$. Number of Monte-Carlo simulations: $N=1000$.}
\label{fig:ElementwiseMSE}
\end{figure}

We find empirical evidence that the COCO denoiser decreases the elementwise $\operatorname{MSE}$, \textit{i.e.}, that the  $\operatorname{MSE}(\smash{\hat{\theta}_k}) \leq \operatorname{MSE}(g_k)$, for arbitrary $d$ and~$K$ (naturally, $\operatorname{MSE}(\smash{\hat{\theta}_k}) = E[\| \smash{\hat{\theta}_k} - \nabla f(x_k)\|^2]$, $\operatorname{MSE}(g_k) = E[\| g_k - \nabla f(x_k)\|^2]$). This inequality is stronger than the one from \Cref{theorem:Total_MSE_COCO<Total_MSE_Raw}, as the former imposes each term on the left-hand side from the latter to be smaller or equal than the respective term on its right-hand side. This makes explicit the  variance reduction provided by COCO, since $\operatorname{Var}(\smash{\hat{\theta}_k}) \leq \operatorname{MSE}(\smash{\hat{\theta}_k}) \leq\operatorname{MSE}(g_k) =  \operatorname{Var}(g_k)$.
One of the instances generated is represented in \Cref{fig:ElementwiseMSE}. We observe that when points are inside the tighter cube, we obtain the best COCO denoising (lower $\operatorname{MSE}(\smash{\hat{\theta}_k})$). On the other hand, for the looser cube, the COCO denoising capability is almost null, tending to the oracle values. Regarding the intermediate cube, it is shown in \Cref{sec:extension_COCO_elementwise_improvement} that the more isolated points are the ones with worse $\operatorname{MSE}(\smash{\hat{\theta}_k})$.

We also found evidence that closer points (\textit{i.e.}, with tighter COCO constraints) lead to lower $\operatorname{MSE}(\smash{\hat{\theta}_k})$
and that, for sufficiently tight COCO constraints, $\operatorname{MSE}(\smash{\hat{\theta}_k}) = C \sigma^2$, with $C$ being $O({1}/{K})$, while for the oracle, $C$ is obviously $O(1)$ (see \Cref{fig:MSE_totalVsVariance}). This result for $\operatorname{MSE}(\smash{\hat{\theta}_k})$ is the same as for the averaging of normal random variables, enabling a nice interpretation: while direct averaging would require that $K$ gradient observations to be available at each iterate $x_k$, with COCO$_K$, we achieve the same $\operatorname{MSE}(\smash{\hat{\theta}_k})$ without having to be stuck on that point for $K$ iterates. COCO can then be interpreted as an extension to the averaging procedure, allowing to integrate information from different positions.

\section{Stochastic Optimization}\label{section:stochastic_optimization}

In this section, we show how COCO robustly improves the performance of typical first-order methods in convex optimization by providing them with variance-reduced gradient estimates. This analysis is firstly performed in a synthetic dataset that matches the noise model used for designing COCO, i.e., a streaming setting in a convex function with Gaussian noise; then we solve two logistic regression problems with real datasets whose gradient noise model does not satisfy these assumptions.

\paragraph{Synthetic data}

\begin{figure}[tbp]
\centering
\includegraphics[width=0.4\textwidth]{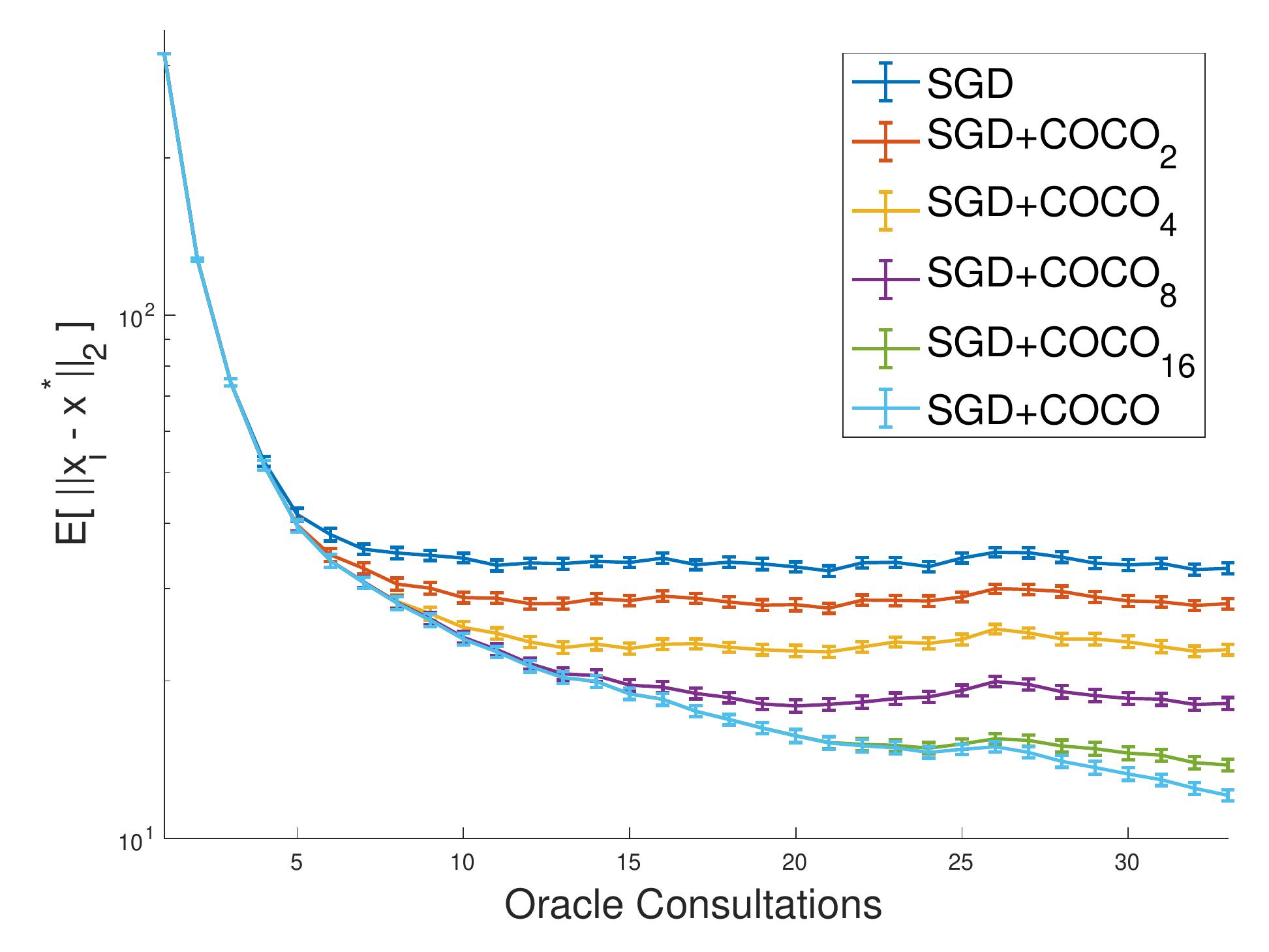}
\includegraphics[width=0.4\textwidth]{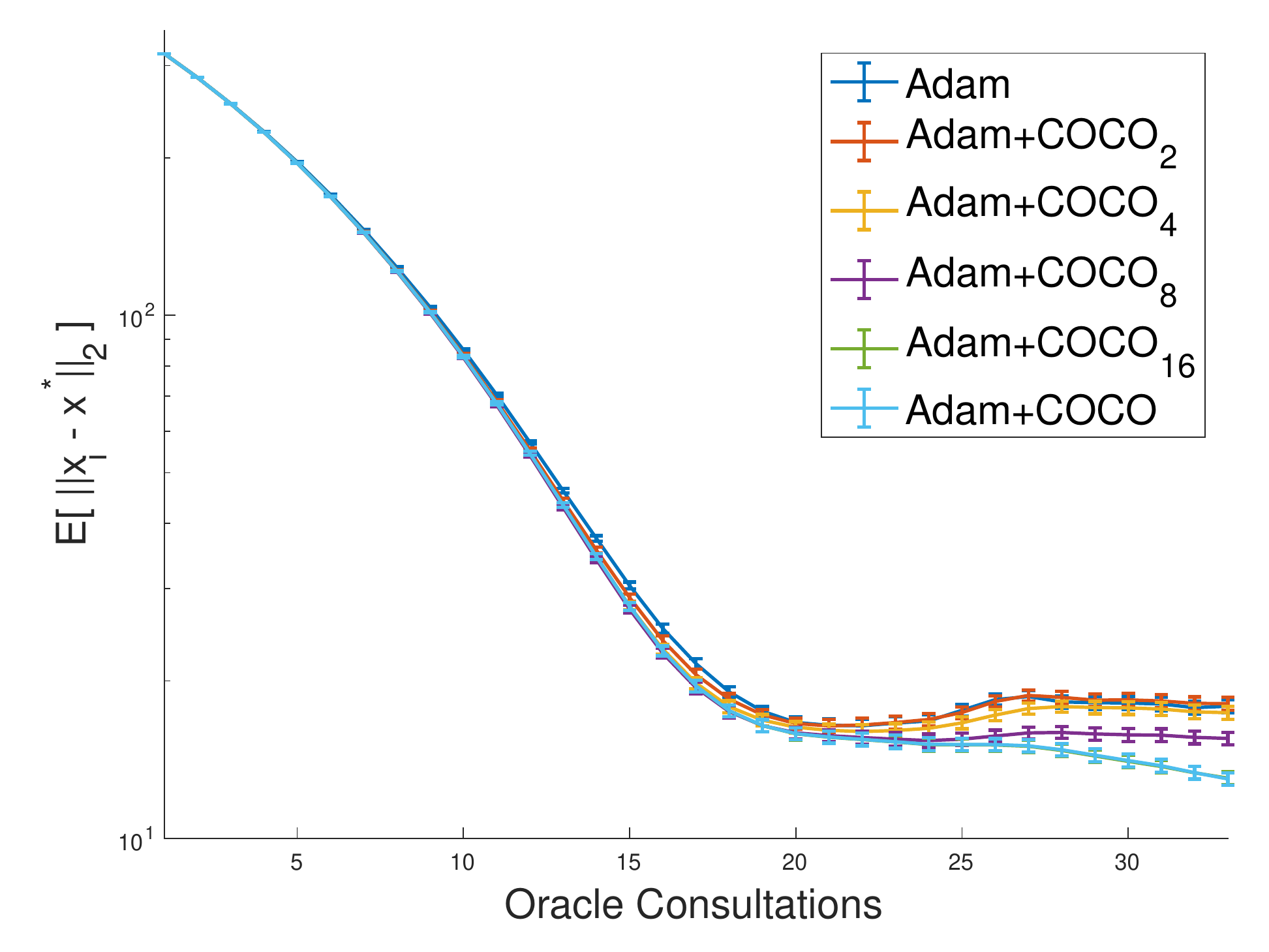}
\caption{COCO denoiser with SGD (\textit{left}) and Adam (\textit{right}) on a synthetic problem satisfying the noise model. $E[\|x_i - x^*\|]$ is averaged over $100$ runs
The width of each marker represents the standard error of the mean. The lines for ``Adam + COCO$_{16}$" and ``Adam + COCO" are superimposed. The notation "COCO" (without subscript) denotes a denoiser that uses all the queried points. We observe monotonic noise reduction with increasing $K$ for both SGD and Adam.
}
\label{fig:synthetic_dataset_SGD&Adam_Performance}
\end{figure}

To assess the usefulness of COCO for stochastic optimization, we first consider a scenario that completely matches \Cref{ass:L_constant} and \Cref{ass:gaussian_noise}. The first-order oracle provides observations whose noise is additive and normally distributed, with $\Sigma = 100\; I$. The objective function is a 10-dimensional ($d=10$) quadratic, $f(x) = {1}/{2}\; x^T A x$, where $A$ is an (anisotropic) Hessian matrix. While this is a simple model, every twice-differentiable convex function can in fact be approximated, at least locally, by a quadratic function.
We use COCO$_K$ as illustrated in \Cref{fig:COCO_schematic}, as a plug-in to both SGD and Adam (representative of the adaptive step size algorithms), obtaining
the results in \Cref{fig:synthetic_dataset_SGD&Adam_Performance}. We also propose a warm-starting procedure for the COCO denoiser iterative solution method (FDPG) for first-order stochastic methods (detailed in \Cref{sec:warm-starting}).

We observe an initial \textit{bias regime}, where all the algorithms converge similarly,
that is successively slowed down and eventually leads to a stagnation, usually called the \textit{variance regime}. We see that COCO leads to improved performance in terms of the variance regime
without compromising the bias regime and that the improvement increases with the number $K$ of gradients simultaneously denoised.

\paragraph{Logistic regression} 

\begin{figure}[tbp]
\centering
\includegraphics[width=0.32\textwidth]{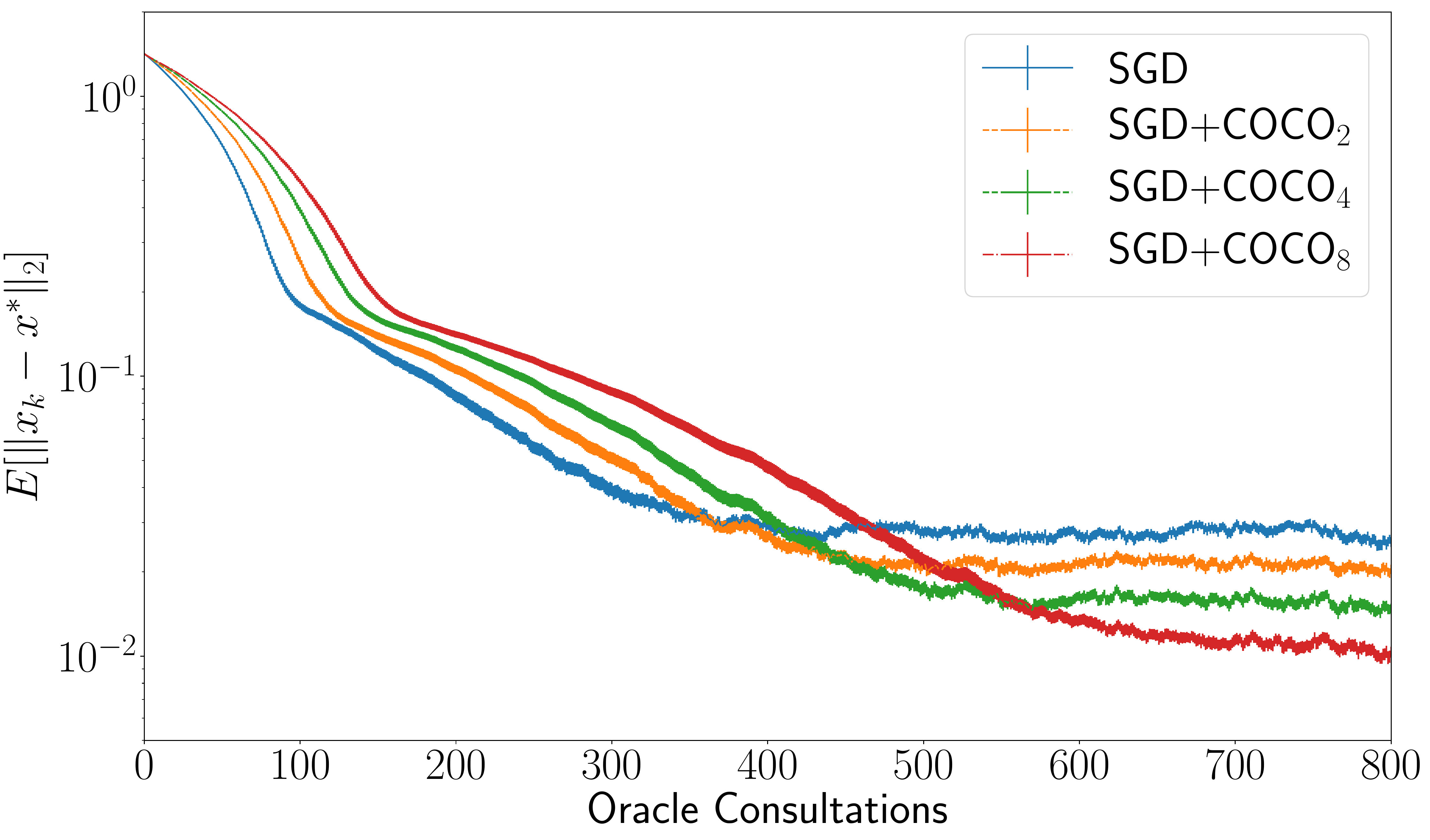}
\includegraphics[width=0.32\textwidth]{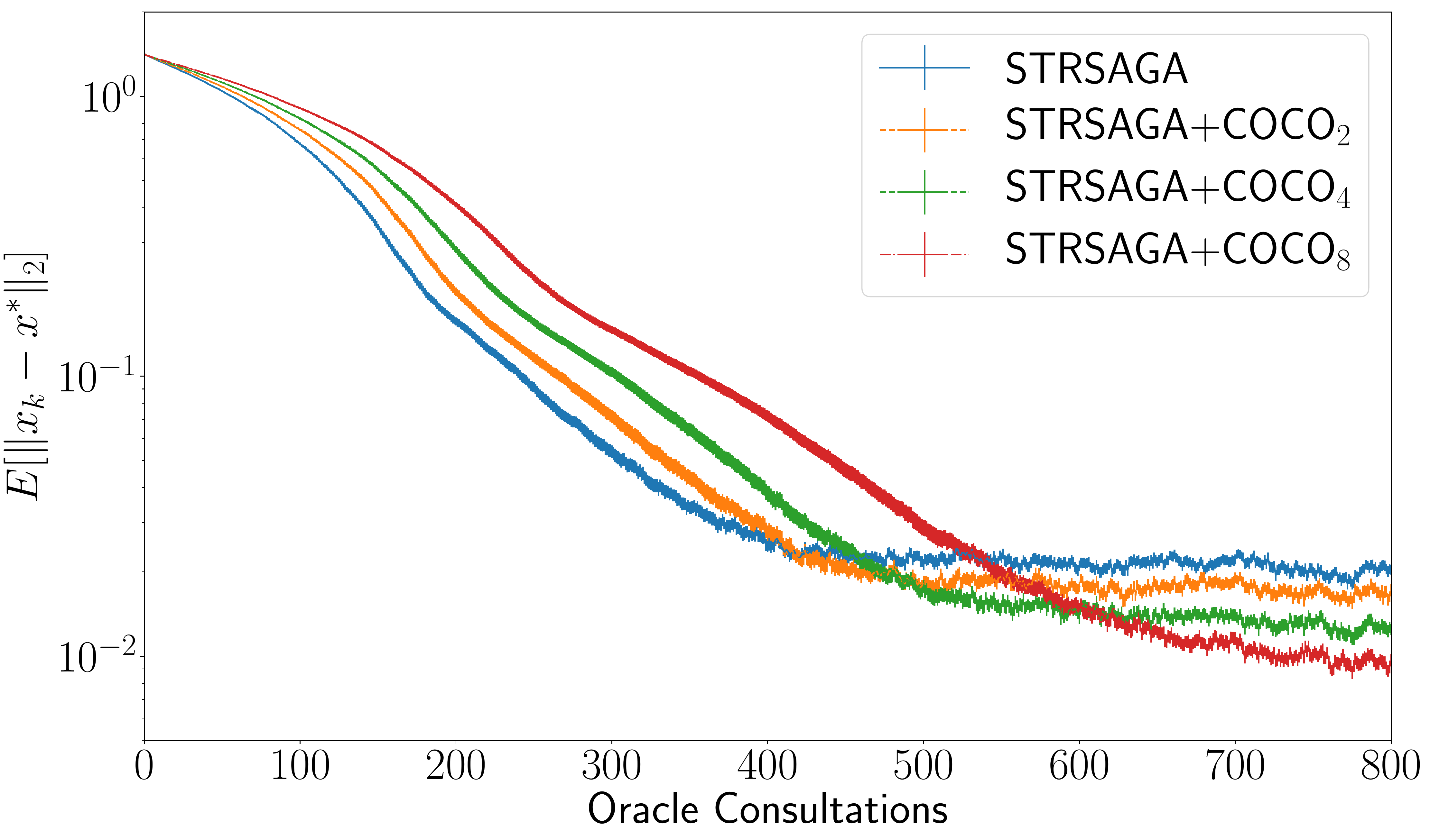}
\includegraphics[width=0.32\textwidth]{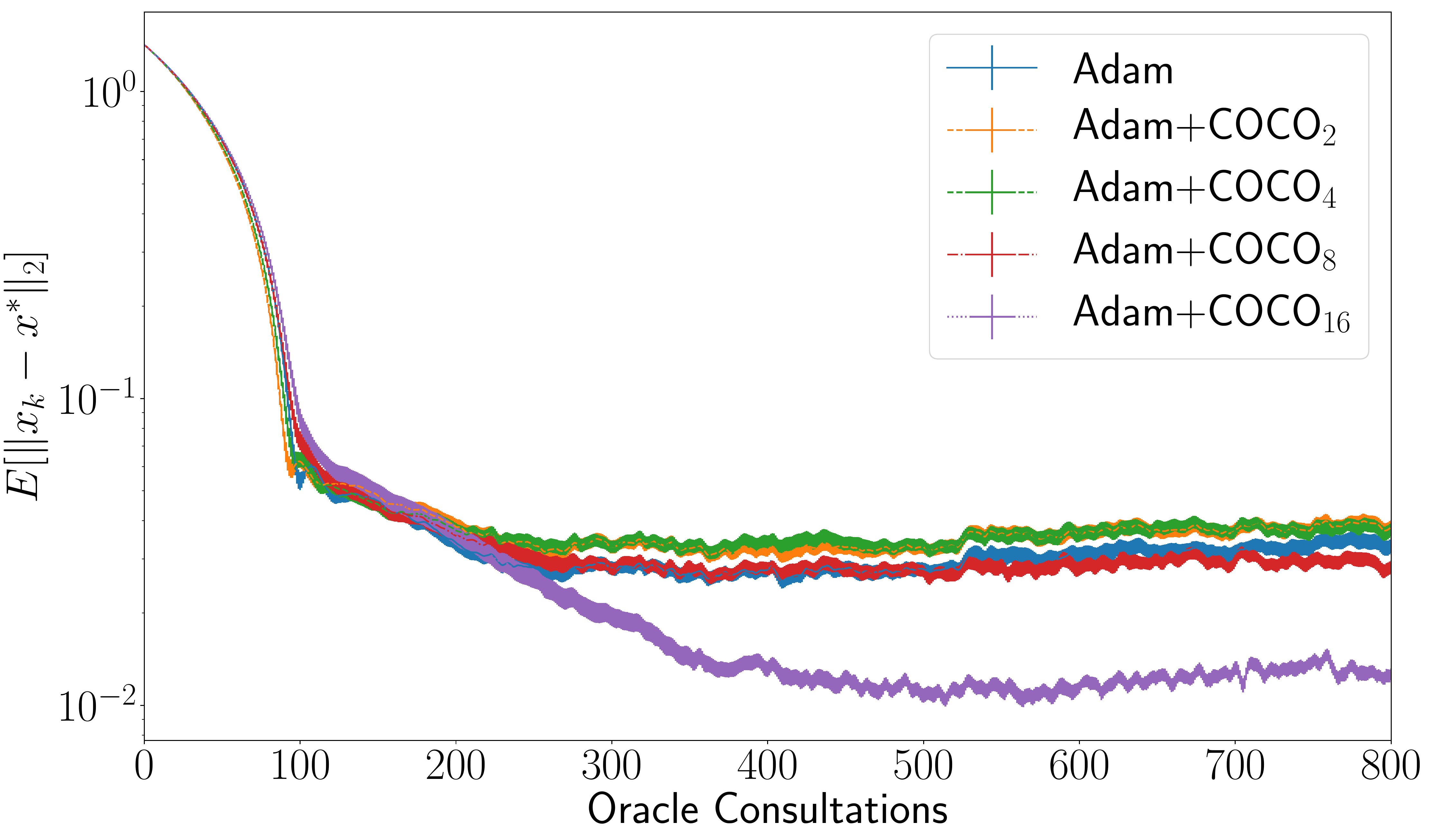}
\includegraphics[width=0.32\textwidth]{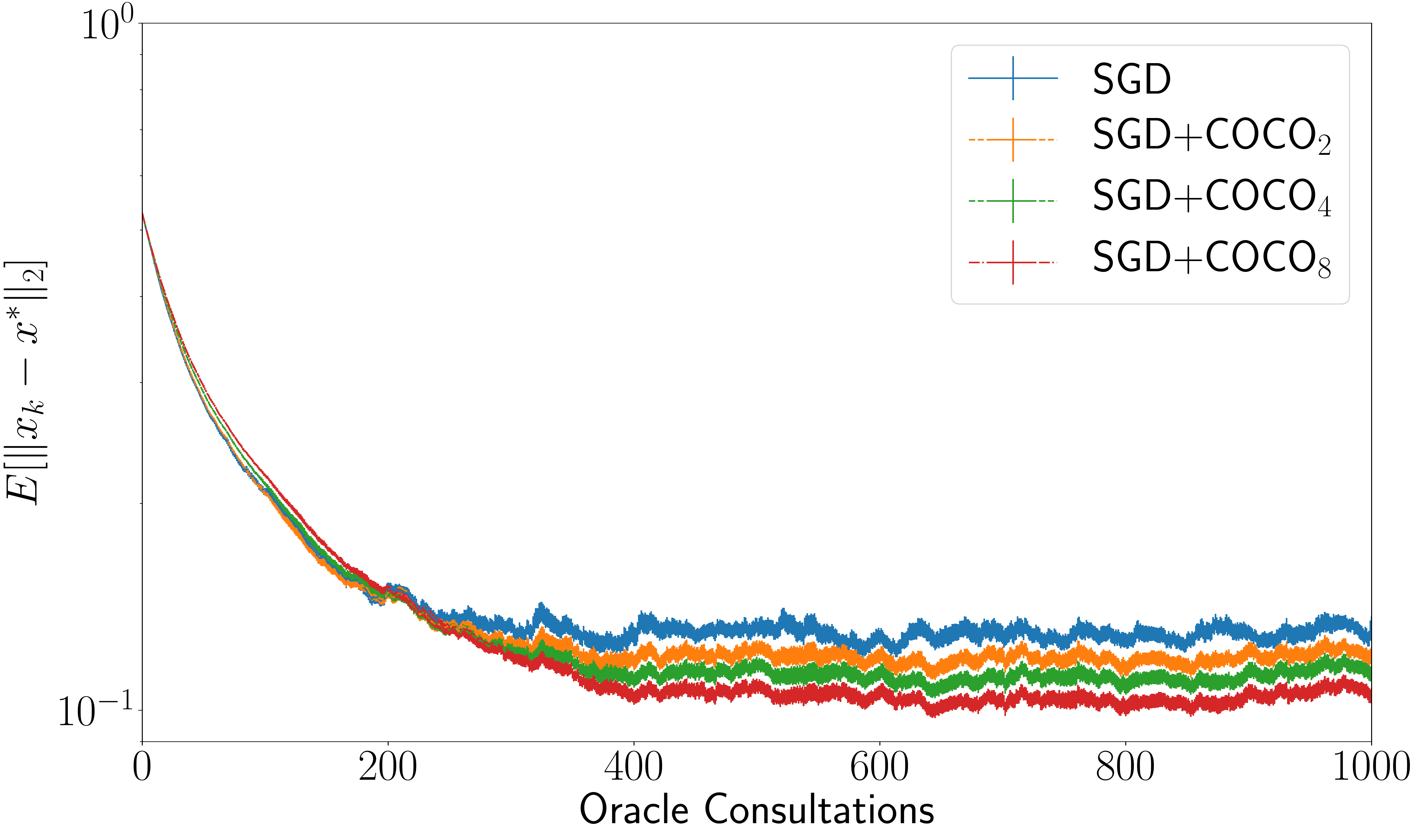}
\includegraphics[width=0.32\textwidth]{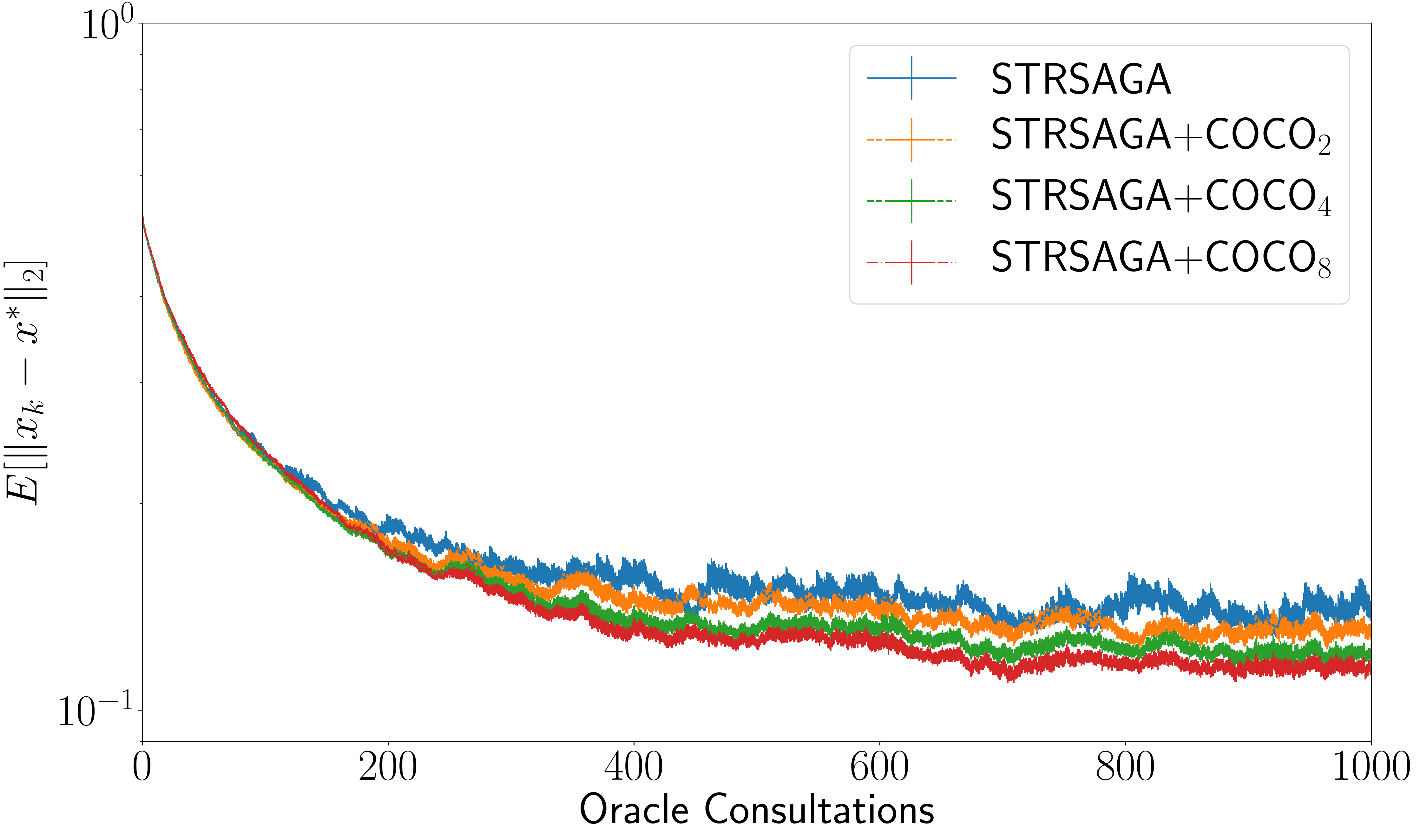}
\includegraphics[width=0.32\textwidth]{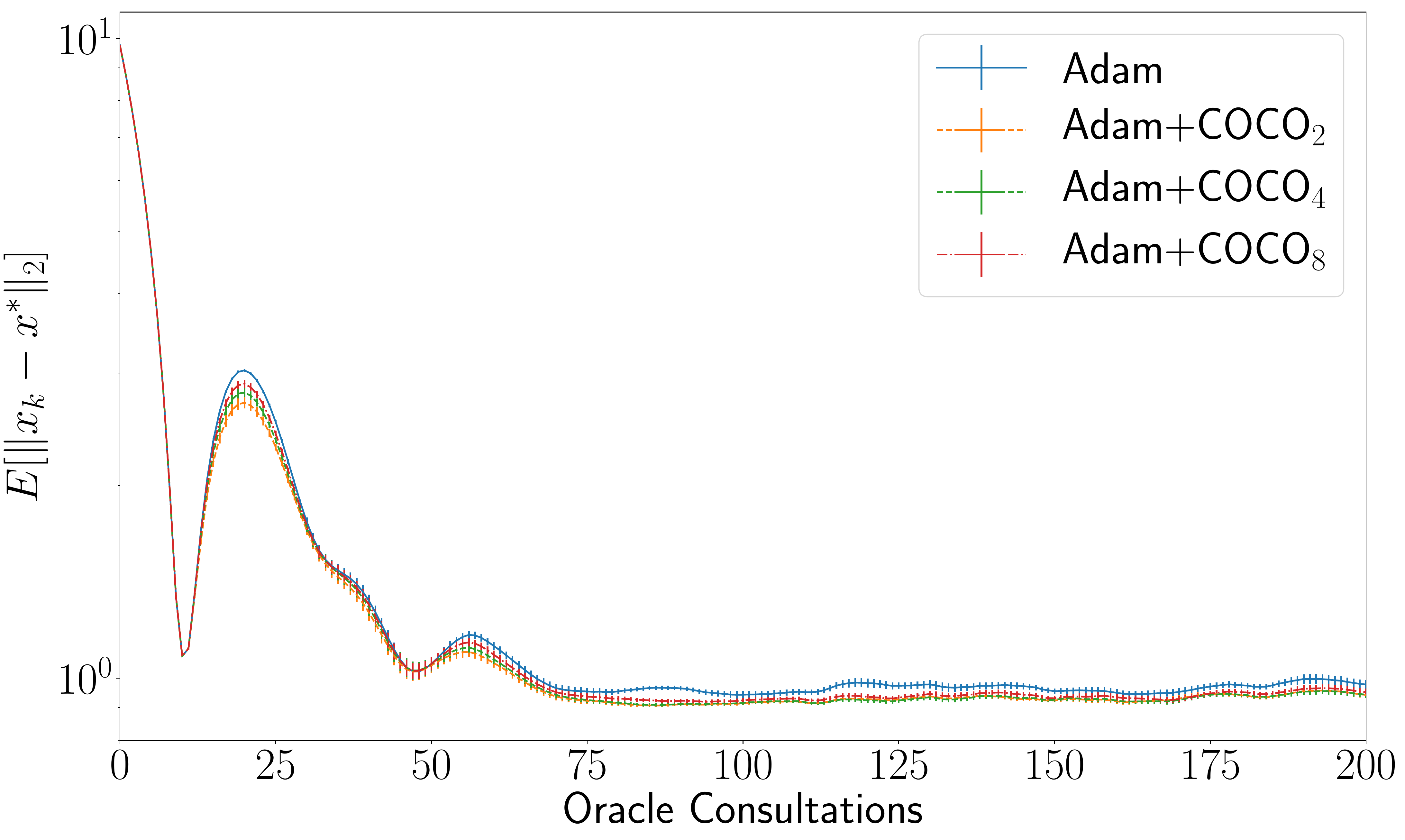}
\caption{COCO denoiser with SGD (\textit{left}), STRSAGA (\textit{center}), and Adam (\textit{right}) for logistic regression. The width of each marker represents the standard error of the mean.
\textit{Top panels}: $100$ runs in the \textit{fourclass} dataset~\citep{chang_libsvm_2011}, starting from the same point. 
\textit{Bottom panels}: $50$ runs in the \textit{mushrooms dataset}~\citep{chang_libsvm_2011}. For SGD and STRSAGA, the results are similar: 
COCO might delay the bias regime but reduces the observed variance, where the amount of variance reduction increases with $K$. For Adam, there are no clear improvements, unless a higher $K$ is used (see top right plot, with $K=16$).}
\label{fig:real_dataset}
\end{figure}

We test the robustness of plugging in COCO in SGD, Adam, and STRSAGA, in real logistic regression problems using the ``fourclass" dataset ($n=862$ data points of dimension $d=2$)
and ``mushrooms" dataset ($n = 8124$, $d = 112$)~\citep{chang_libsvm_2011}. For the ``mushrooms" dataset, we added a Tikhonov regularization term to the objective function, which is formulated according to the typical finite-sum setting. At each gradient evaluation, one of those examples is randomly picked, from which we compute a noisy gradient of the objective. This setup falls out of the assumptions for COCO, since the sampled gradients are not independent and the noise is not additive and normally distributed. 
The results are shown in \Cref{fig:real_dataset}.

For the ``fourclass" dataset, despite the bias delay, consistent variance improvements are observed for SGD and STRSAGA with increasing number $K$ of gradients simultaneously denoised. In contrast, Adam almost does not show bias compromise (due to its adaptive nature), but its variance gains only appear for higher values of $K$. For the ``mushrooms" dataset, consistent variance improvements are observed both for SGD and STRSAGA with increasing $K$ without significant bias delay. Note that although Adam benefits with COCO, its variance improvements do not consistently improve with $K$. We also emphasize that for Adam the number of oracle queries is different from the other two algorithms due to its adaptive nature and, thus, faster convergence towards the variance regime.

\section{Discussion and Future Work}\label{sec:limitations}

Our denoiser formulation is natural and shows the feasibility of using gradient co-coercivity for denoising. The accelerated first-order method for the problem enables us to solve moderately sized instances. Nonetheless, there are some aspects that remain unanswered and would be interesting to consider as future work, which we expect will be taken in part by the community. 

It would be interesting to demonstrate the universality of the evidence provided by our experiments, namely in what respects to the estimator bias and variance (at least for COCO$_2$, for which there is a closed-form solution). Another aspect that deserves further study is the decrease of the elementwise $\operatorname{MSE}$ of COCO$_K$ with~$K$. Regarding the usage of COCO as a plug-in for stochastic optimization, it would be important to study convergence guarantees (which, naturally, also depend on the baseline algorithm) and to quantify the gains in variance reduction. While these have been studied empirically, mathematical analysis for the aspects that contribute to variance reduction remains lacking (e.g., closeness of the queried points). 
Aspects of computational efficiency can also motivate future work. 
Improving the quadratic scaling of the number of constraints with the number of points without worsening too much the denoiser is also important. In fact, our analysis showed that the larger gains in denoising come from closeby points, which could motivate strategies to reduce (maybe to a linear dependence) the number of constraints that could effectively be considered without compromising the results. More exploratory lines of research would address the possibility of denoising gradients using different assumptions on the underlying objective function. For example, strong convexity, which has led to better convergence rates for stochastic optimization algorithms, or the finite sum decomposition that is omnipresent in machine learning applications. In the non-convex setting, it could be interesting to leverage $L$-smoothness instead of gradient co-coercivity. While a formulation based on these properties has not been considered, they have strong parallels with the formulation here provided and should lead to similar algorithms.

\section{Conclusion}\label{sec:conclusion}

We propose a variance reduction plug-in to first-order stochastic optimization algorithms, which leverages gradient co-coercivity (COCO) of the objective function to denoise the observed gradients. The COCO denoiser is obtained from the joint maximum likelihood estimation of the function gradients, for which we derive the closed-form solution when dealing with two observations and introduce a fast iterative method for the general case. We study the estimator properties, emphasizing that the COCO denoiser yields cleaner gradients than the stochastic oracle. Our experiments illustrate that current stochastic first-order methods benefit from using gradients denoised by COCO.

\clearpage
\section*{Acknowledgements}\label{sec:acknowledgements}

The authors would like to thank Prof. Robert Gower, for kindly providing the code used as a starting point for the logistic regression problem considered.


\bibliographystyle{unsrtnat}
\bibliography{COCO-Denoiser-Bibliography}

\newpage
\appendix

\section{Detailed Derivation of the FDPG Method}\label{sec:detailed derivation FDPG}

\subsection{Reformulation of the Problem}

We start by multiplying the objective function \Cref{eq:COCO_formalization} by $1/2$ for the sake of simplicity in the next steps, yielding the following problem:
\begin{equation*}
    \begin{aligned}
        & \underset{\theta_1, \ldots, \theta_K}{\text{minimize}}
    & & \frac{1}{2} \sum_{k=1}^{K} \| g_k - \theta_k \|^2 \\
    & \text{subject to}
    & & \| \theta_m - \theta_l - \frac{L}{2} (x_m - x_l) \| \leq \frac{L}{2} \|x_m - x_l \|, \quad 1 \leq m < l \leq K,
    \end{aligned}
\end{equation*}
This problem remains the same as the one provided in \Cref{eq:COCO_formalization}, where the new form for the constraints is obtained by completing the square in the expression from the original formulation:
\begin{align}
        &\; \frac{1}{L} \|\theta_m - \theta_l\|^2 \leq (\theta_m - \theta_l)^T(x_m-x_l) \nonumber \\
        \Leftrightarrow  &\;\|\theta_m - \theta_l\|^2 - L(\theta_m - \theta_l)^T(x_m-x_l) + \frac{L}{4} \|x_m - x_l\|^2 - \frac{L}{4} \|x_m - x_l\|^2 \leq 0 \label{eq:add_and_subtract}\\
        \Leftrightarrow &\; \|\theta_m - \theta_l - \frac{L}{2}(x_m-x_l)\|^2 \leq \| \frac{L}{2}(x_m-x_l)\|^2 \nonumber \\
        \Leftrightarrow &\; \|\theta_m - \theta_l - \frac{L}{2}(x_m-x_l)\| \leq \| \frac{L}{2}(x_m-x_l)\|,\nonumber 
\end{align}
where in \Cref{eq:add_and_subtract} we add and subtract ${L}\|x_m - x_l\|^2/{4}$ and all the other steps are simple manipulations.
Note that, in this case, $\theta_m - \theta_l \in \mathcal{B}\left({L}(x_m - x_l)/{2},\; {L}\; \| x_m - x_l \|/{2} \right)$\footnote{The notation $\mathcal{B}(c, r)$ denotes the set of points within a ball centered at $c$ and of radius $r$, \textit{i.e.}, $\mathcal{B}(c, r) = \{ x\in\mathbb{R}^n: \|x-c \| \leq r  \}$.}.

Now, performing the change of variables $\alpha_k = \theta_k - g_k$, the problem becomes:
\begin{equation*}
    \begin{aligned}
    & \underset{\alpha_1, \ldots, \alpha_K}{\text{minimize}}
    & &\frac{1}{2} \sum_{k=1}^{K} \| \alpha_k \|^2 \\
    & \text{subject to}
    & &\| \alpha_m - \alpha_l + c_{ml} \| \leq r_{ml}, \quad 1 \leq m < l \leq K,
    \end{aligned}
\end{equation*}
where $c_{ml} = \left(g_m - ({L}/{2})\; x_m\right) - \left(g_l - ({L}/{2})\; x_l\right)$ and $r_{ml} = {L}\|x_m - x_l\|/{2}$.

The indicator function can be defined as
\begin{equation*}
    \mathbf{1}_{E}(x) = 
    \begin{cases} 
        0\quad &\text{if} \; x \in E \\
        \infty\quad & \text{if} \;x \notin E .
    \end{cases}
\end{equation*}

Using this definition, the primal problem can be finally formulated as
\begin{equation*}
    \begin{aligned}
    & \underset{\alpha}{\text{minimize}}
    & & \underbrace{\frac{1}{2} \| \alpha \|^2}_{p(\alpha)}
     + \underbrace{\frac{ }{ }\mathbf{1}_{\mathcal{B}}(A \alpha + c)}_{q(A\alpha)},
    \end{aligned}
\end{equation*}
where $\alpha = [\alpha_{1}, \;\alpha_2,\; \ldots, \;\alpha_K]^T$, $A \alpha = [\alpha_{1} - \alpha_2, \;\alpha_{1} - \alpha_3, \;\ldots, \;\alpha_{1} - \alpha_K, \;\alpha_2 - \alpha_3, \;\ldots, \;\alpha_{K-1} - \alpha_K]^T$, $c = [c_{12},\;c_{13}, \;\ldots, \;c_{1K}, \;c_{23},\;\ldots,\; c_{K-1K} ]^T$ and $\mathcal{B} = \mathcal{B}(0,\;r_{12}) \times  \mathcal{B}(0,\;r_{13}) \times \ldots \times \mathcal{B}(0,\;r_{1K}) \times
\mathcal{B}(0,\;r_{23}) \times \ldots \times
\mathcal{B}(0,\;r_{K-1K})$.


In this formulation, we want to minimize the sum of two convex functions, where the first is differentiable and the second is non-differentiable, but still closed\footnote{A function $f: \mathbb{R}^d \to \mathbb{R}$ is said to be closed if for each $\alpha \in \mathbb{R}$, the sublevel set $\{x \in \text{dom} f | f(x) \leq a \}$ is a closed set.}. This is the setup to which the iterative shrinkage-thresholding algorithms (ISTA) are designed for. In particular, when the non-differentiable function is a simple indicator function, that method can be interpreted as the Projected Gradient Descent. However, in this formulation, that function is composed with a linear map $A$, case in which there is no closed-form for the proximity operator.

Given this, a reformulation using Lagrange duality is used. First, the problem can be rewritten as:
\begin{equation*}
    \begin{aligned}
    & \underset{\alpha, \beta}{\text{minimize}}
    & & p(\alpha) + q(\beta)  \\
    & \text{subject to}
    & & A \alpha = \beta . 
    \end{aligned}
\end{equation*}
It is possible to write the Lagrangian for the reformulated problem:
\begin{align*}
    L(\alpha, \beta, s) 
        &= p(\alpha) + q(\beta) + s^T(A\alpha-\beta)  \\
        &= p(\alpha) + s^T A \alpha + q(\beta) - s^T \beta .
\end{align*}
The Lagrange dual function can be computed:
\begin{align*}
    L(s) &= \inf_{\alpha, \beta} L(\alpha, \beta, s) \\
         &= \inf_{\alpha} \left(p(\alpha) + s^T A \alpha\right) + \inf_{\beta} \left(q(\beta) - s^T \beta\right). 
\end{align*}
Thus,
\begin{align*}
    - L(s) &= \sup_{\alpha} \left((-A^T s)^T  \alpha - p(\alpha)\right) + \sup_{\beta} \left( s^T \beta - q(\beta) \right) . 
\end{align*}
By definition, for a generic function, its (Fenchel) conjugate is defined as $f^*(s) = \underset{x}{\text{sup}}\; (s^Tx - f(x))$. Therefore, it is possible to conclude that:
\begin{align*}
    - L(s) = p^*(-A^Ts) + q^*(s) . 
\end{align*}
It remains to obtain the specific form of $p^*(s)$ and $q^*(s)$. Regarding the former:
\begin{align*}
    p^*(s) 
        &= \sup_{\alpha} \left( s^T  \alpha - \frac{1}{2} \|\alpha\|^2 \right) \\
        & = \frac{1}{2} \|s\|^2,
\end{align*}
where the second equality easily comes from differentiating $s^T  \alpha - {1}/{2}\;  \|\alpha\|^2$ with respect to $\alpha$ and equating to zero. Therefore, the value obtained for $\alpha$ is then replaced on the original expression. 
Regarding $q^*$:
\begin{align*}
        q^*(s) 
            &= \sup_{\beta}\left( s^T  \beta - \mathbf{1}_{\mathcal{B}}(\beta + c)\right)\\
            & = \sup_{\beta}\left\{ s^T  \beta: \beta + c \in \mathcal{B} \right\} \\
            & = \sup_{\beta}\left\{ \sum_{1\leq m < l \leq K} s_{ml}^T \beta_{ml}: \beta_{ml} + c_{ml} \in \mathcal{B}(0, r_{ml}) \right\} \\
            & = \sum_{1\leq m < l \leq K} \sup_{\beta}\left\{ s_{ml}^T \beta_{ml}: \|\beta_{ml} + c_{ml}\| \leq r_{ml} \right\} \\
            & = \sum_{1\leq m < l \leq K} r_{ml} \|s_{ml}\| - s_{ml}^Tc_{ml}. 
\end{align*}
where the last step is obtained via: 
\begin{align*}
    \sup_{b} \left\{ s^T b : \| b - ( - c ) \| \leq r \right\}
        &= \sup_{b} \left\{ s^T ( -c + u ) : \| u \| \leq r \right\}\\
        &= -s^T c + \sup_{b} \left\{ s^T u : \| u \| \leq r \right\} \\
        &= -s^T c + r \| s \|, 
\end{align*}
and the equality $\sup_{b} \left\{ s^T u : \| u \| \leq r \right\} = r \| s \|$ is obtained by (we assume $s$ different from $0$, otherwise, the equality is trivial):

\begingroup
\renewcommand\labelenumi{(\theenumi)}
\begin{enumerate}
\item picking $u = r \frac{s}{\| s \|}$ (note that  $\| u \| \leq r$), we have $s^T u = r \| s \|$. 
\\This shows $\sup \{ s^T u : \| u \| \leq r \} \geq r \| s\| $;
\item From Cauchy-Schwartz inequality:  $s^T u \leq \| s \| \| u \|$. Since $\| u \| \leq r$: $s^T u \leq r \| s \|$. 
\\So, $\sup_{b} \{ s^T u : \| u \| \leq r \} \leq r \| s\| $.
\end{enumerate}
\endgroup

From (1) and (2), we obtain the intended result. Therefore, the minimization problem can be rewritten in the following form:
\begin{equation}\label{eq:COCO_dual_reformulation2}
    \begin{aligned}
    & \underset{s}{\text{minimize}}
    & & \underbrace{\frac{1}{2} \| - A^T s \|^2}_{p^*(-A^Ts)}\;
     + \underbrace{ \frac{ }{ }\sum_{1\leq m < l \leq K} r_{ml} \|s_{ml}\| - s_{ml}^Tc_{ml} }_{q^*(s)} . 
    \end{aligned}
\end{equation}
At this point, the linear mapping $A$ has now been transferred to the differentiable term. This change allows us now to find a closed-form expression for the proximity operator of $q^*(s)$, as the gradient of the first term can still be computed even considering its composition with $A^T$.

\subsection{Proximity Operator Computation}

By definition, the proximity operator of a generic closed, convex function $f$ is:
\begin{equation*}
    \text{prox}_{f}(x) = \argmin_{u} \frac{1}{2} \| u-x \|^2 + f(u). 
\end{equation*}
We are interested in obtaining $\text{prox}_{\mu q^*}(s)$, for any given $\mu>0$. Thus:
\begin{align}
    \text{prox}_{\mu q^*}(s)  
        &= s - \text{prox}_{(\mu q^*)^*}(s) \label{eq:3.10}\\
        &= s - \text{prox}_{(q^{*})^*\cdot\mu}(s) \label{eq:3.11}\\
        &= s - \text{prox}_{q\cdot\mu}(s), \label{eq:3.12}
\end{align}
where in \Cref{eq:3.10} it is applied the well-known Moreau identity $\text{prox}_{f}(x) = x - \text{prox}_{f^*}(x)$; in \Cref{eq:3.11}, we used $(\mu f)^*(x) = f^* \cdot \mu \;(x) = \mu f^*({x}/{\mu})$ and, in \Cref{eq:3.12}, the property $(f^*)^* = f$, which holds for any closed, convex function. Now, note that:
\begin{align}
    q \cdot\mu \;(s) &= \mu \;q\left(\frac{s}{\mu}\right) \nonumber\\
    &= \mu \;\mathbf{1}_{\mathcal{B}}\left(\frac{s}{\mu} + c\right) \nonumber\\
    &= \mathbf{1}_{\mathcal{B}}\left(\frac{s}{\mu} + c\right), \label{eq:3.13}
\end{align}
since, in \Cref{eq:3.13},  $\mu$ can be dropped as $\mathbf{1}_{\mathcal{B}}$ returns either $0$ or $\infty$.
Therefore,
\begin{align}
        \text{prox}_{\mu q^*}(s)  &= s -  \argmin_{u} \left(\frac{1}{2} \| u-s \|^2 + \mathbf{1}_{\mathcal{B}}\left(\frac{u}{\mu} + c\right)\right) \nonumber \\
        &= s - \mu\left(\argmin_{v} \left(\frac{1}{2} \| \mu(v-c)-s \|^2 + \mathbf{1}_{\mathcal{B}}(v) \right)-c\;\right) \label{eq:change_variable_proximity} \\ 
        & = s - \mu\left( \argmin_{v\in\mathcal{B}} \left( \frac{1}{2} \| v-(c+\frac{s}{\mu}) \|^2 \right)-c\;\right) \nonumber \\
        & = s - \mu\;(\;v_{\text{proj}}-c\;)\nonumber,
\end{align}
where the change of variable $v = \frac{u}{\mu} + c$ was used in \Cref{eq:change_variable_proximity} and the orthogonal projection of $c_{ml} + {s_{ml}}/{\mu}$ onto the ball $\mathcal{B}(0,\; r_{ml})$, with $1 \leq m < l \leq K$, is denoted by~$v_{ml}$, whose stacking results in $v_{\text{proj}} = \underset{v\in\mathcal{B}}{\; \mathrm{argmin}}\; \| v-(c+{s}/{\mu}) \|^2$.

\subsection{Fast Dual Proximal Gradient Method}

Recalling \Cref{eq:COCO_dual_reformulation2}, we now have a first term, $p^*(-A^Ts)$, differentiable, for which the gradient has a closed-form and a second term, $q^*(s)$, non-differentiable but for which we can compute also a closed-form and inexpensive proximity operator. We are now in place to apply ISTA, where the iterates are generated by alternating between taking a gradient step of the differentiable function and taking a proximal step.

The gradient for $p^*(-A^Ts)$ can be easily computed: $\nabla_s \; p^*(-A^Ts) =  AA^Ts$. Furthermore, from this expression is straightforward to observe that the first term, $p^*(-A^Ts)$, is necessarily $L$-smooth, with Lipschitz constant $L_{p^*} = \sigma_{\max}(A)^2$. Given this, not only the optimal step size for a gradient update is known ($\gamma = 1/L_{p^*}$), but also, just as happened with first-order algorithms (for differentiable functions) in the deterministic convex setting, it is possible to accelerate the ISTA resorting to a Nesterov acceleration similar scheme, \textit{i.e.}, enabling momentum to contribute in the generated iterates. The accelerated version of ISTA is known as FISTA. Moreover, this perspective of applying FISTA to the dual problem is a well-studied technique, already introduced in this paper as the FDPG method (applied to COCO in Algorithm \ref{algorithm:FDPG}). Through FDPG, it is possible to find an approximation of the optimal solution of the dual problem, $s^*$. However, we are interested in recovering the solution of the primal problem, $\alpha^*$, which, nevertheless, can be easily obtained through $\alpha^* = -A^Ts^*$. Consequently, the gradient estimates are recovered as $\smash{\hat{\theta}}_k
=\alpha_k^* + g_k$.

Strong duality, \textit{i.e.}, $p(\alpha^*) + q(A\alpha^*) = - \left( p^*(-A^Ts^*) + q(s^*)\right)$, holds for this convex optimization problem. For example, a Slater point can be easily obtained by considering $\smash{\hat{\theta}}_k = {L}/{2}\; x_k$, assuming the iterates to be different from each other ($x_i \neq x_j \;\text{if}\; i \neq j$). This is expectable if we presume that these iterates are generated through a stochastic first-order method.

\subsection{Warm-starting}\label{sec:warm-starting}

\begin{figure}[tbp]
\centering
\includegraphics[width=0.45\textwidth]{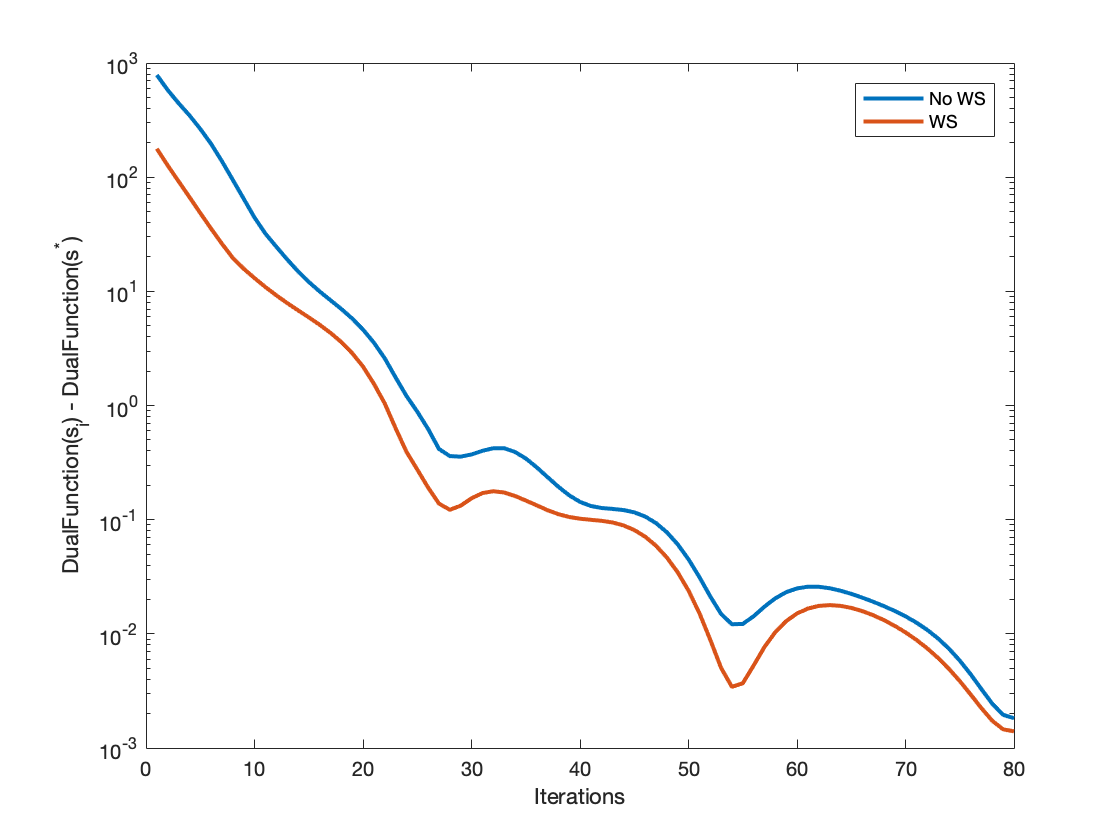}
\caption{Dual objective function obtained for the different iterates of the COCO$_K$ solution method, using the warm-starting procedure (WS) and without using it (No WS). DualFunction$(s)$ is the dual objective function $p^*(-A^Ts) + q^*(s)$, $s_i$ is the vector that results from stacking the different $s_{ml}$ at iteration $i$, and $s^*$ is the corresponding optimal vector.}
\label{fig:warm_starting}
\end{figure}

By coupling a baseline algorithm with COCO$_K$, at iteration $i$, only the oldest gradient ($g_{i-K}$) is forgotten and a new one ($g_i$) is kept in memory. Thus, it is reasonable to consider taking advantage of the COCO$_K$ solution obtained for the previous iterate to obtain a new solution faster. We propose a warm-starting procedure for the COCO$_K$ solution method (FDPG). In particular, we achieve it by a careful initialization of the dual variable, $s$. In fact, $s$ is the vector that results from stacking the different $s_{ml}$, where each $s_{ml}$ addresses the co-coercivity constraint between the COCO estimates for gradient $m$, $\smash{\hat{\theta}_m}$, and for gradient $l$, $\smash{\hat{\theta}_l}$. Since we expect the estimates for old gradients to only have small relative variations among them on the new iterate as they have been ``filtered" at least once, we initialize these $s_{ml}$ to the values obtained for the correspondent dual variables in the previous COCO$_K$ solution. For the multiple $s_{ml}$ concerning the new gradient, we do not have any information yet, thereby being initialized to a default value. Our implementation of this warm-starting procedure allows the iterative method to start with a much better guess of $s^*$, thereby achieving satisfactory approximate solutions faster (see \Cref{fig:warm_starting}).

\section{Proofs of Theorems} \label{section:appendix_proofs}

For all the proofs of the theorems below, the starting point is the COCO denoiser formulation for a generic $\Sigma$:
\begin{equation}\label{COCO_formulation_generic_Sigma}
    \begin{aligned}
    & \underset{\theta_1, \ldots, \theta_K}{\text{minimize}}
    & & \frac{1}{2} \sum_{k=1}^{K}  (g_k - \theta_k)^T  \Sigma^{-1}  (g_k - \theta_k) \\
    & \text{subject to}
    & & \frac{1}{L} \| \theta_m - \theta_l \|^2 \leq \langle 
    \theta_m - \theta_l, x_m - x_l \rangle, \; 1 \leq m < l \leq K.
    \end{aligned}
\end{equation}

\subsection{Proof of Theorem 3.1} \label{section:proof_theorem_closedformK=2}

    

\begin{proof}

For $K=2$, \Cref{COCO_formulation_generic_Sigma} becomes:
\begin{equation*}
    \begin{aligned}
    & \underset{\theta_1, \theta_2}{\text{minimize}}
    & & (g_1 - \theta_1)^T \Sigma^{-1}(g_1 - \theta_1) + (g_2 - \theta_2)^T \Sigma^{-1}(g_2 - \theta_2) \\
    & \text{subject to}
    & & \| \theta_1 - \theta_2 \|^2 -  L \langle 
    \theta_1 - \theta_2, \; x_1 - x_2 \rangle \leq 0 . 
    \end{aligned}
\end{equation*}

In order to solve this problem, the Karush-Kuhn-Tucker (KKT) conditions will now be used. It can be observed that there are no equality constraints. We have:
\begin{equation*}
    \begin{aligned}
    & f (\theta_1, \theta_2) = (g_1 - \theta_1)^T \Sigma^{-1}(g_1 - \theta_1) + (g_2 - \theta_2)^T \Sigma^{-1}(g_2 - \theta_2) 
    & &  \\
    & f_1 (\theta_1, \theta_2) = \| \theta_1 - \theta_2 \|^2 -  L \langle 
    \theta_1 - \theta_2, \; x_1 - x_2 \rangle . 
    & & 
    \end{aligned}
\end{equation*}

Since both functions are differentiable and convex, we can use $\partial(f(x)) = \{ \nabla f(x)\}$\footnote{$\partial(\cdot)$ denotes the subdifferential operator. For a continuous function $f: \mathbb{R}^d \to \mathbb{R}$, $c \in \mathbb{R}^d$ is a subgradient of $f$ at $x \in \mathbb{R}^d$ if and only if $f(y) - f(x) \geq c^T (y - x)$, with $y \in \mathbb{R}^d$. The set of all the subgradients of $f$ at $x$ is called the subdifferential of $f$ at $x$, $\partial(f(x))$.}. This can be applied for simplification of the stationarity condition, through the linearity of the gradient operator. Therefore, the KKT conditions yield the following system of equations:
\begin{align*}
    \begin{cases}
        2\Sigma^{-1} (\hat{\theta}_1 - g_1) + \mu_1 \left[2(\hat{\theta}_1 - \hat{\theta}_2)-L(x_1-x_2)\right] = 0 & \text{[ \textit{i.} Stationarity in order to } \hat{\theta}_1\text{ ]} \\
        2\Sigma^{-1}(\hat{\theta}_2 - g_2) - \mu_1 \left[2(\hat{\theta}_1 - \hat{\theta}_2)-L(x_1-x_2)\right] = 0 & \text{[ \textit{ii.} Stationarity in order to } \hat{\theta}_2\text{ ]} \\
        \mu_1 \;\left(\| \hat{\theta}_1 - \hat{\theta}_2 \|^2 - L \langle \hat{\theta}_1 -\hat{\theta}_2,\;x_1 - x_2\rangle\right) = 0 & \text{[ \textit{iii.} Complementary Slackness ]}\\
        \| \hat{\theta}_1 - \hat{\theta}_2 \|^2 - L \langle \hat{\theta}_1 -\hat{\theta}_2,\;x_1 - x_2\rangle \leq 0 & \text{[ \textit{iv.} Primal Feasibility ]}\\
        \mu_1 \geq 0 & \text{[ \textit{v.} Dual Feasibility ]} . 
    \end{cases}
\end{align*}

From \textit{iii.}, two cases must be considered:
\begin{itemize}
    \item $\mathbf{\mu_1 = 0}:$ In this case, from complementary slackness (\textit{iii.}), $\| \smash{\hat{\theta}}_1 - \smash{\hat{\theta}}_2 \|^2 \leq L \langle \smash{\hat{\theta}}_1 -\smash{\hat{\theta}}_2,\;x_1 - x_2\rangle$
\end{itemize}
In that case, from \textit{i.} and \textit{ii.}, it is easy to conclude that $\smash{\hat{\theta}}_1 = g_1$ and $\smash{\hat{\theta}}_2 = g_2$. Therefore, we note that this happen when $\| g_1 - g_2 \|^2 \leq L \langle g_1 - g_2,\;x_1 - x_2\rangle$.

\begin{itemize}
    \item $\mathbf{\mu_1 > 0}:$ In that case, from complementary slackness (\textit{iii.}), $\| \smash{\hat{\theta}}_1 - \smash{\hat{\theta}}_2 \|^2 = L \langle \smash{\hat{\theta}}_1 -\smash{\hat{\theta}}_2,\;x_1 - x_2\rangle$.
\end{itemize}
By summing \textit{i.} and \textit{ii.}:
\begin{equation*}
    \smash{\hat{\theta}}_1 + \smash{\hat{\theta}}_2 = g_1 + g_2 . 
\end{equation*}

This equality is particularly interesting and further developed in the proof of \Cref{theo:sum_thetas_equal_sum_gs}. By replacing it in~\textit{i.} and~\textit{ii.}, we obtain:
\begin{equation}\label{eq:theta_1_and_2_function_of_mu}
    \begin{aligned}
    &\hat{\theta}_1 = (\Sigma^{-1} + 2 \mu_1 I)^{-1} [(\Sigma^{-1} + \mu_1 I) g_1 + \mu_1 g_2 + \mu_1 \frac{L}{2}(x_1-x_2)] \\
    &\hat{\theta}_2 = (\Sigma^{-1} + 2 \mu_1 I)^{-1} [ \mu_1 g_1 + (\Sigma^{-1} + \mu_1 I) g_2 - \mu_1 \frac{L}{2}(x_1-x_2)] . \\
    \end{aligned}
\end{equation}
Then, by replacing those results in $\| \smash{\hat{\theta}}_1 - \smash{\hat{\theta}}_2 \|^2 = L \langle \smash{\hat{\theta}}_1 -\smash{\hat{\theta}}_2,\;x_1 - x_2\rangle$, it yields the following expression:
    \begin{equation}
        I \mu_1^2 + \Sigma^{-1} \mu_1 - (\Sigma^{-1})^2 C = 0,
    \end{equation}
with $C = ({\| g_1-g_2 \|^2 - L \langle g_1 -g_2,\;x_1 - x_2\rangle})/({L^2 \| x_1 - x_2 \|^2})$. Note that $C \geq 0$, since, otherwise, we would have $\| g_1-g_2 \|^2 < L \langle g_1 -g_2,\;x_1 - x_2\rangle$ and we would be in the case of $\mu_1 = 0$. By considering that $\Sigma = \sigma^2  I$, the equation above yields for each diagonal entry:
\begin{equation}\label{eq:quadratic_expression}
    \mu_1^2 + \frac{1}{\sigma^2} \mu_1 - \left(\frac{1}{\sigma^2}\right)^2 C = 0 . 
\end{equation}

The non-diagonal entries are not informative, as they are all zero. The only solution of \Cref{eq:quadratic_expression} that respects dual feasibility (\textit{v.}) is:
\begin{equation*}
    \mu_1 = \frac{1}{\sigma^2} \left( \frac{-1 + \sqrt{1+4C}}{2} \right) . 
\end{equation*}
Replacing this value of $\mu_1$ in \Cref{eq:theta_1_and_2_function_of_mu}, we obtain the intended result.
\end{proof}

\subsection{Proof of Theorem 4.1}\label{sec:proof_sum_thetas_equal_sum_gs}


\begin{proof}
From the COCO denoiser formalization for generic $\Sigma$ (\Cref{COCO_formulation_generic_Sigma}) and $K$ points considered, the KKT conditions yield $K$ stationarity
equations. 
Its $i$-th equation is of the form:
\begin{equation*}
    2 \Sigma^{-1}(\hat{\theta}_i - g_i) + \sum_{j=1, j \neq i}^{K} \mu_{ij} [\; 2(\hat{\theta}_i - \hat{\theta}_j) - L(x_i - x_j)\;] = 0,
\end{equation*}

Summing the $K$ equations, all the constraint terms cancel out pairwisely, yielding:
\begin{equation*}
    \sum_{i=1}^{K} 2 \Sigma^{-1}(\hat{\theta}_i - g_i) = 0  \Leftrightarrow 2 \Sigma^{-1} \sum_{i=1}^{K} (\hat{\theta}_i - g_i) = 0 \\
    \Leftrightarrow  \sum_{i=1}^{K} \hat{\theta}_i = \sum_{i=1}^{K} g_i . 
\end{equation*}
\end{proof}

\subsection{Proof of Theorem 4.2} \label{sec:proof_Total_MSE_COCO<Total_MSE_Raw}


\begin{proof}
The Orthogonal Projection operator on a set $S$ is defined as 
\begin{equation*}
    \begin{aligned}
    & P_S(x): &&\mathbb{R}^d &&\to \mathbb{R}^d \\
    &&& x &&\mapsto \underset{y \in S}{\mathrm{argmin}}\; \|x-y\| . 
    \end{aligned}
\end{equation*}
In the case in which $S \subset \mathbb{R}^d$ is closed and convex, the following property holds:
\begin{equation*}
    \|P_S(a) - P_S(b)\| \leq \|a-b\| . 
\end{equation*}


Let also $S$ be the feasible set of the problem in \Cref{COCO_formulation_generic_Sigma}. Note that, in that case, $S$ is a convex and closed set as it results from the intersection of ellipsoids, which are convex and closed sets themselves. Moreover, when $\Sigma = \sigma^2 I$, \Cref{COCO_formulation_generic_Sigma} yields: 
\begin{equation*}
    \hat{\theta} = \underset{\theta \in S}{\mathrm{argmin}}\; {1}/{\sigma^2}\; \| \theta - g \|^2 = \underset{\theta \in S}{\mathrm{argmin}}\; \| \theta - g \| = P_S(g)
\end{equation*}
Noting that $\nabla f = P_S(\nabla f)$ since $\nabla f \in S$, \textit{i.e.}, the true gradients of an $L$-smooth and convex function are necessarily co-coercive\footnote{Note that this statement is only true for $L \geq L_{\text{real}}$, where $L_{\text{real}}$ denotes the minimal Lipschitz constant of $\nabla f$.}, it follows:
\begin{equation}\label{eq:MSE_theta_leq_MSE_g}
    \begin{aligned}
        & &&\| \hat{\theta} - \nabla f\|= \| P_S(g) - P_S(\nabla f) \| \leq \| g - \nabla f \| . 
    \end{aligned}
\end{equation}
Squaring both sides of the inequality in \Cref{eq:MSE_theta_leq_MSE_g} and applying the Expectation operator, the result intended is obtained.
\end{proof}

\section{Estimator Properties}\label{sec:app_estimator_properties}

\subsection{Theoretical Analysis Extension for 1D}\label{sec:app_coco_tightness}

In order to find a reasonable answer to the problem addressing the COCO constraints tightness (raised in \Cref{sec:estimator_properties_theoretical}), the following setup is proposed: for the sake of simplicity, our focus remains on the one-dimensional situation ($d=1$) where we have access to two different points, $x_1$ and $x_2$. Without loss of generality, let us assume $x_1 > x_2$. The true gradients on those points are $\nabla f(x_1)$ and $\nabla f(x_2)$, whose noisy versions (provided by the oracle) are $g_1$ and $g_2$. Therefore, $g_1 \indep g_2$\footnote{The notation $\indep$ denotes independence between random variables.} and $\Sigma = \sigma^2$, which is as general as possible for the one-dimensional case. We obtain the following result for the probability of $g_1$ and $g_2$ being co-coercive, $p_{\text{inactive}}$:
\begin{equation}\label{eq:p_inactive_expression}
    p_{\text{inactive}} = \Phi \left( \frac{L\Delta_x - \Delta_{\nabla f}}{\sqrt{2}\sigma} \right) - \Phi \left( \frac{- \Delta_{\nabla f}}{\sqrt{2}\sigma} \right),
\end{equation}
where $\Delta_x = x_1 - x_2$ and $\Delta_{\nabla f} = \nabla f(x_1) - \nabla f(x_2)$.

\begin{proof}
     Note that $g_i \sim \mathcal{N}(\nabla f(x_i),\sigma^2)$. Moreover, the co-coercivity constraint between $g_1$ and $g_2$ is inactive when:
\begin{align*}
    \|g_1 - g_2\|^2 < L \; \langle g_1 - g_2, \; x_1 -x_2 \rangle 
    & \Leftrightarrow (g_1 - g_2)^2 - L(g_1-g_2)(x_1-x_2) < 0 \\
    & \Leftrightarrow (g_1 - g_2)(g_1 - g_2 - L(x_1-x_2)) < 0 \\
    & \Leftrightarrow 0 < g_1 - g_2 < L(x_1 - x_2) .
\end{align*}

Therefore, noticing that  $g_1 - g_2  \sim  \mathcal{N}(\nabla f(x_1)-\nabla f(x_2),  2 \sigma^2)$ and defining $\Delta_x = x_1 - x_2$ and $\Delta_{\nabla f} = \nabla f(x_1) - \nabla f(x_2)$:
\begin{align*}
    P(\|g_1 - g_2\|^2 < L \; \langle g_1 - g_2, \; x_1 -x_2 \rangle)
    & = P(0 < g_1 - g_2 < L \Delta x)\\
    & = \Phi \left( \frac{L\Delta_x - \Delta_{\nabla f}}{\sqrt{2}\sigma} \right) - \Phi \left( \frac{- \Delta_{\nabla f}}{\sqrt{2}\sigma} \right) \\
    & = p_{\text{inactive}} .
\end{align*}
\end{proof}

In fact, taking into account that $\Phi(-\infty) = 0$, $\Phi(0) = 0.5$ and $\Phi(+\infty) = 1$, some high level observations about the behaviour of $p_{\text{inactive}}$ as a function of $\Delta_x$ can be readily determined:
\begin{equation*}
    \begin{cases}
    \text{if $L$ is overestimated:} &p_{\text{inactive}} \overset{\Delta_x \to 0}{\to} 0;\; p_{\text{inactive}} \overset{\Delta_x \to +\infty}{\to} 1 \\
    \text{if $L$ is perfectly estimated:} &p_{\text{inactive}} \overset{\Delta_x \to 0}{\to} 0;\; p_{\text{inactive}} \overset{\Delta_x \to +\infty}{\to} 0.5 \\
    \text{if $L$ is underestimated:} &p_{\text{inactive}} \overset{\Delta_x \to 0}{\to} 0;\; p_{\text{inactive}} \overset{\Delta_x \to +\infty}{\to} 0  .
    \end{cases}
\end{equation*} 

\begin{figure}[tbp]
\centering
\includegraphics[width=0.5\textwidth]{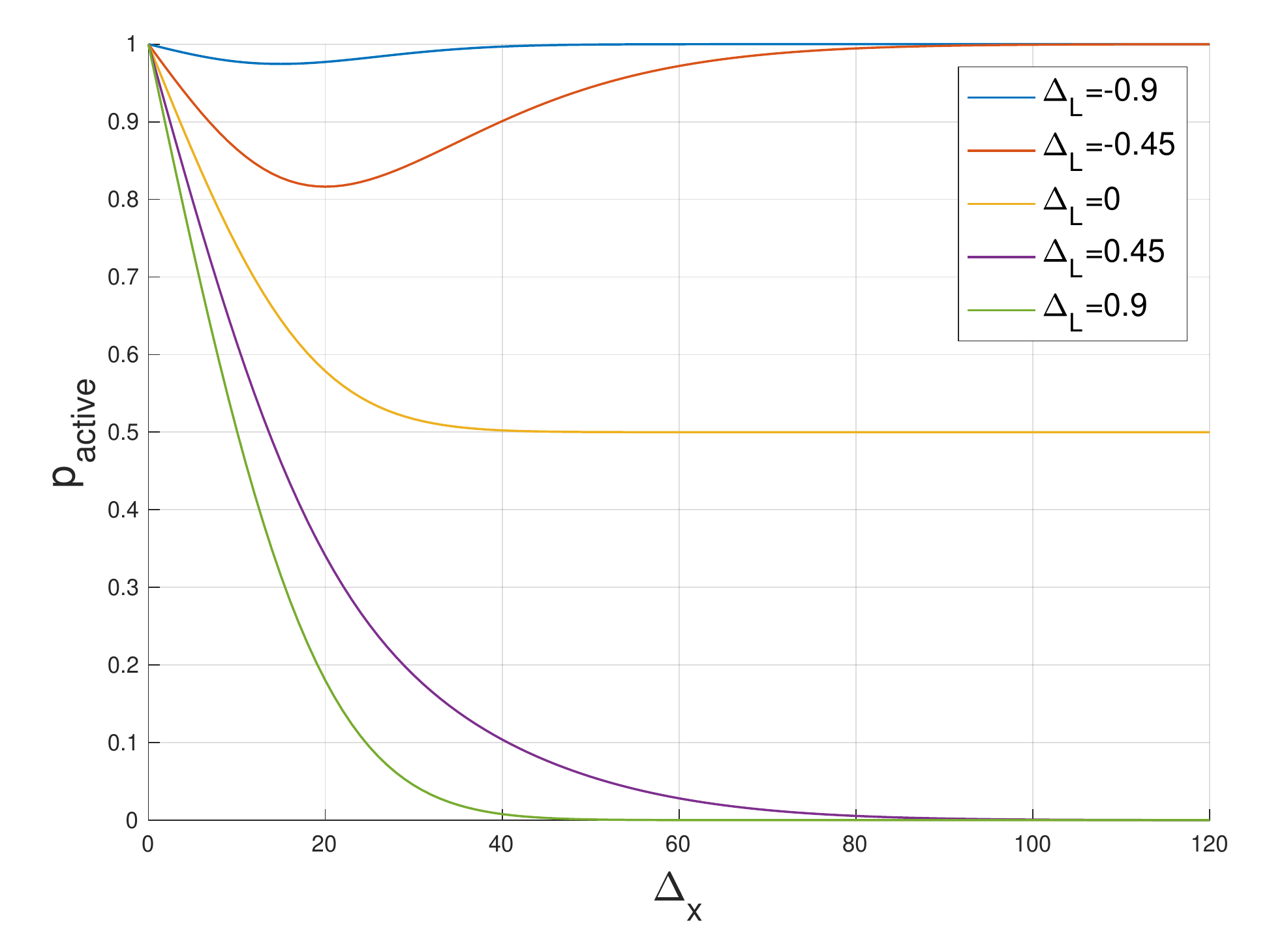}
\caption{Representation of the probability of two sampled noisy gradients, $g_1$ at $x_1$ and $g_2$ at $x_2$, not being co-coercive as a function of the distance between $x_1$ and $x_2$ ($p_{\text{active}} (\Delta_x)$) for different $\Delta_L = L - L_{\text{real}}$. We considered $f(x) = x^2/2$ (thus, $L_{\text{real}} = 1$) and $\sigma = 10$.}
\label{fig:constraints_tightness}
\end{figure}

To further illustrate this behavior, $p_{\text{active}}$ ($= 1 - p_{\text{inactive}}$) is represented as a function of $\Delta_x$ for a quadratic objective function in \Cref{fig:constraints_tightness}. The motivation of using a quadratic comes not only, as we are in the one-dimensional case, from the second derivative of this function being, actually, $L_{\text{real}}$ in all the domain (see footnote 7 for the meaning of $L_{\text{real}}$), but also from the fact that every twice-differentiable convex function can be approximated, at least locally, by a quadratic function. Moreover, this analysis is made considering that the Lipschitz constant used as an input for the denoiser, $L$, may not correspond to $L_{\text{real}}$. 

These results can be intuitively explained. These explanations are only provided for $p_{\text{active}}(\Delta x)$, as those are the ones represented in \Cref{fig:constraints_tightness} and can be easily transferred to $p_{\text{inactive}}(\Delta_x)$ due to their complementarity:

\begin{itemize}
    \item Independently of $\Delta_L$, $p_{\text{active}}(0) = 1$ as from the co-coercivity inequality (\Cref{eq:co-coercivity}), the right-hand side is zero, and therefore the constraint will always be violated with exception of the  case in which $g_1 = g_2$, a set of points which, nevertheless, has zero Lebesgue measure. Moreover, as $\Delta_x$ increases, the right-hand side of \Cref{eq:co-coercivity} increases, allowing the increase of the Lebesgue measure of the set of points which do not violate co-coercivity. Given this, obviously $p_{\text{active}}$ decreases for all $\Delta_L$. After this point, different $\Delta_L$ lead to different behaviours.
    \item When the curvature is overestimated ($\Delta_L > 0$), the true gradients, $\nabla f(x_1)$ and $\nabla f(x_2)$ are always co-coercive, and if only those were considered, there wouldn't be active constraints. Therefore, the decrease in the tightness of the constraint (increase of the right-hand side of \Cref{eq:co-coercivity}) is directly related to the increasing relevance of the true gradients with the distance, explaining why $p_{\text{active}}$ tends to zero. Note that, naturally, the higher the $\Delta_L$, the faster $p_{\text{active}}$ tends to zero.
    \item When the curvature is precisely the one estimated, $\Delta_L = 0$, the true gradients always lead to a case of equality in \Cref{eq:co-coercivity}). Due to noise, when the perturbation leads the noisy gradients to be further away than supposed, the constraint activates. By the same token, the constraint is inactive when the perturbation decreases the difference between gradients. As the noise follows a Gaussian distribution, it increases or decreases the difference between gradients with the same probability, explaining why $p_{\text{active}}$ tends to 0.5 for larger distances.
    \item When the curvature is underestimated, $\Delta_L < 0$, the variation of the true gradients is always incoherent with the $L_{estimated}$ (true gradients change faster than allowed). Consequently, the constraint tends to be active ($p_{\text{active}}$ tends to 1) as the distance increases. Note that the lower the $\Delta_L$, the more incoherent the true gradients are with the co-coercivity constraint and, therefore, the faster the $p_{\text{active}}$ tends to 1.
\end{itemize}

\begin{figure}[tbp]
\centering
\includegraphics[width=0.5\textwidth]{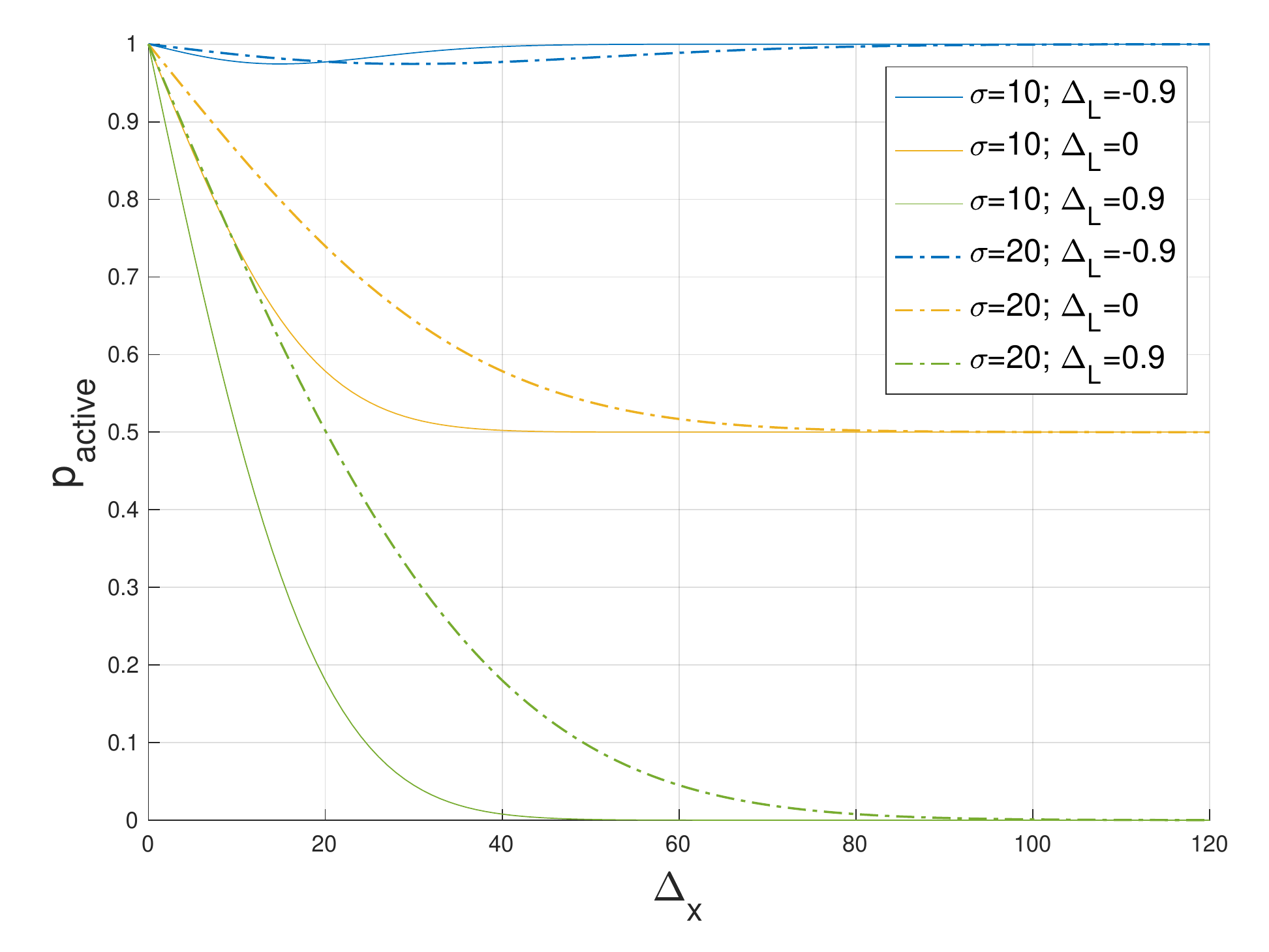}
\caption{Comparison of $p_{\text{active}}$ for different levels of noise. The solid lines are retrieved from \Cref{fig:constraints_tightness} ($\sigma = 10$), while the dashed lines are obtained by doubling noise magnitude ($\sigma = 20$). The real value of the parameter $L_{\text{real}}$ is the same: $L_{\text{real}} = 1$.}
\label{fig:theoretical_prob_constraints_sigma}
\end{figure}

A careful observation of \Cref{eq:p_inactive_expression} also exposes the dependence of $p_{\text{active}}$ (or $p_{\text{inactive}}$) on the level of noise in the problem, expressed through $\sigma$. Therefore, it is expected that higher noise leads to slower convergence of those probabilities to the aforementioned values as it attenuates the influence from the true gradients. This intuition is confirmed in \Cref{fig:theoretical_prob_constraints_sigma}.

\subsection{Extension of \texorpdfstring{\Cref{theorem:Total_MSE_COCO<Total_MSE_Raw}}{Theorem}}\label{sec:MSEvsDistance}

\begin{figure}[tbp]
\centering
\includegraphics[width=\textwidth]{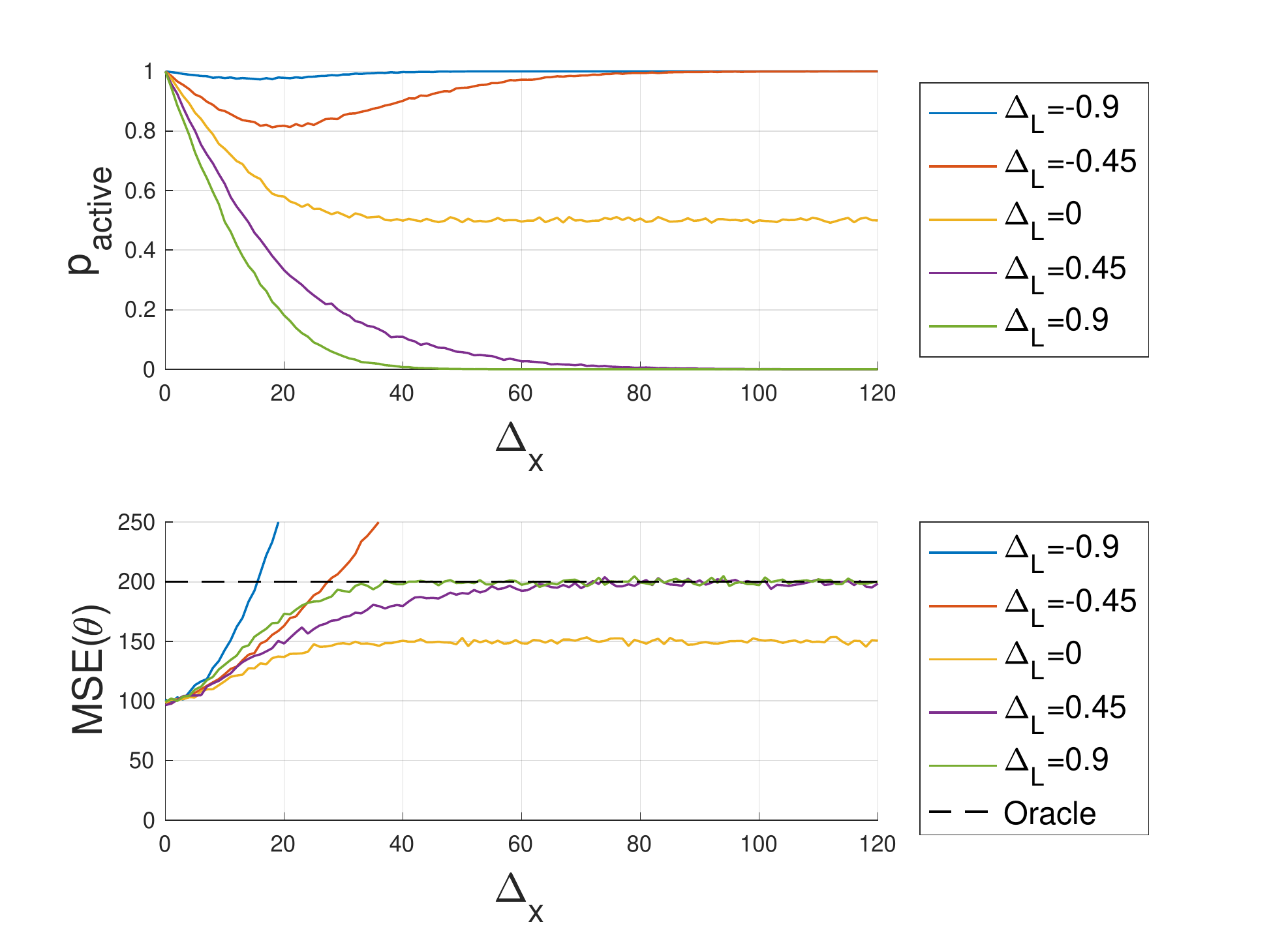}
\caption{
\textit{Top}: Experimental plot for $p_{\text{active}}$ as a function of $\Delta_x$ for different values of $\Delta_L$ (empirically recovers \Cref{fig:constraints_tightness}).
\textit{Bottom}: Computed $\operatorname{MSE}(\smash{\hat{\theta}})$ (number of Monte-Carlo simulations: $N=10000$). We have $\operatorname{MSE}(g) = 200$, represented as a dashed line. Both plots are obtained for $f(x) = x^2/2$, with one point fixed at $x_1 = 0$ and a variable point at $x_2 = \Delta_x$. The oracle provides gradient estimates with additive Gaussian noise with $\Sigma = \sigma^2 = 100$.}
\label{fig:MSEvsDistance}
\end{figure}

As a result of \Cref{theorem:Total_MSE_COCO<Total_MSE_Raw}, we analyze to what extent the COCO estimator outperforms the oracle. In fact, it is possible to obtain a closed-form result for the $\operatorname{MSE}(g)$ for a general number of points considered, $K$, a general dimension $d$ and $\Sigma = \sigma^2 I$: $\operatorname{MSE}(g)= Kd\sigma^2$. Regarding $\operatorname{MSE}(\smash{\hat{\theta}})$, even though without a closed-form solution, we were able to observe its direct dependence on the COCO constraints tightness, as represented in \Cref{fig:MSEvsDistance}.

\begin{itemize}
    \item For $\Delta_x = 0$, all the curves have $p_{\text{active}} = 1$ and $\operatorname{MSE}(\smash{\hat{\theta}}) = 100 = {\sigma^2}/{2}$. This recovers a well known result for the average of $K$ random variables with Gaussian distributions: their $\operatorname{MSE}$\footnote{The $\operatorname{MSE}$ corresponds to the variance of an unbiased estimator, which is the case of the average of random variables following normal distributions.} is ${\sigma^2}/{K}$. In fact, when $\Delta_x = 0$, COCO$_K$ denoiser outputs the average of the observed gradients (recall closed-form solution for COCO$_2$ - \Cref{theorem:closed_form_COCO_2}). Furthermore, COCO denoiser can therefore be considered an extension for the variance reduction through averaging method, but which tolerates samples from different points. This can be viewed as one of the main advantages of COCO;
    \item When the $L$ is underestimated ($\Delta_L<0$), the $\operatorname{MSE}(\smash{\hat{\theta}})$ is not guaranteed to be lower than $\operatorname{MSE}(g)$. Nevertheless, there still is a range of $\Delta_x$ where $\operatorname{MSE}(\smash{\hat{\theta}}) \leq \operatorname{MSE}(g)$. The more underestimated $L$ is, the smaller this region becomes. This observation not only recalls that the result from \Cref{theorem:Total_MSE_COCO<Total_MSE_Raw} only holds for $\Delta_L \geq 0$, but also reinforces the importance of ensuring that the $L$ considered for COCO is an upper bound for $L_{\text{real}}$;
    \item When the $L$ is perfectly estimated ($\Delta_L$ = 0), just as the $p_{\text{active}}$ tends to an intermediate value, so it happens with $\operatorname{MSE}(\smash{\hat{\theta}})$. This is the ideal situation, as $\operatorname{MSE}(\smash{\hat{\theta}})$ is minimal for every $\Delta_L$. Moreover, note that when the $p_{\text{active}}$ curve stabilizes, the $\operatorname{MSE}(\smash{\hat{\theta}})$ also stabilizes, reinforcing the expected relation between those curves;
    \item When the $L$ is overestimated ($\Delta_L> 0$), just as $p_{\text{active}}$ tends to 0, the $\operatorname{MSE}(\smash{\hat{\theta}})$ also tends to the $\operatorname{MSE}(g)$ reference curve. Moreover, it is possible to see that when  $p_{\text{active}}$ stabilizes around 0, so it happens to $\operatorname{MSE}(\smash{\hat{\theta}})$ around the oracle's curve. This is easily explained, again, by the fact that when the constraints are loose, the COCO denoiser outputs the oracle results without any ``filtering";
    \item Regarding the noise variance, $\sigma^2$, it should be stated that, as previously seen in \Cref{fig:theoretical_prob_constraints_sigma}, its increase would shift the stabilization of the curves from the $\Delta_L >0$ cases towards higher $\Delta_x$.
\end{itemize}

\subsection{Extension to COCO Elementwise \texorpdfstring{$\boldsymbol{\operatorname{MSE}}$}{MSE} Improvement} \label{sec:extension_COCO_elementwise_improvement}

\begin{figure}[tbp]
\centering
\includegraphics[width=0.32\textwidth]{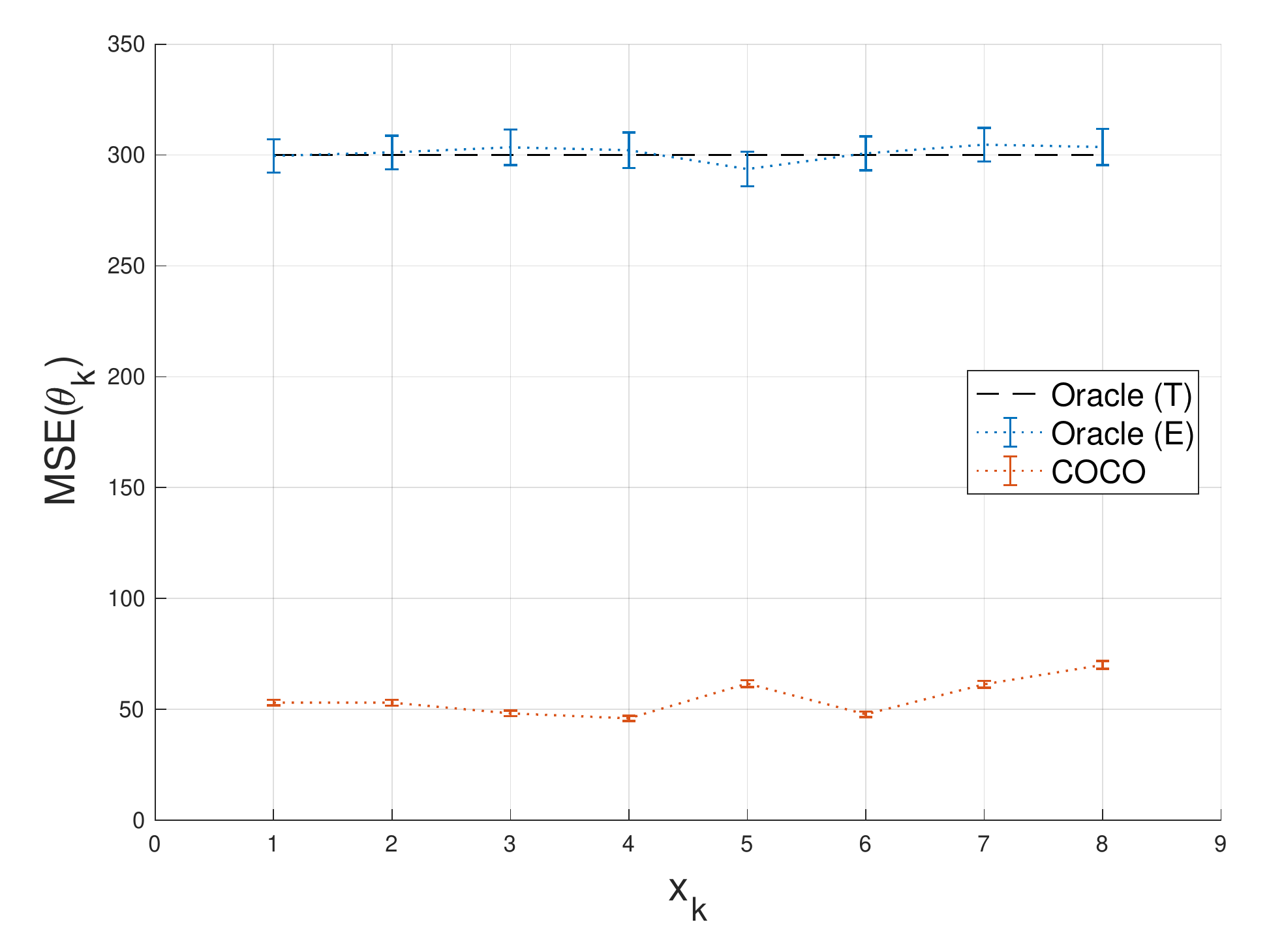}
\includegraphics[width=0.32\textwidth]{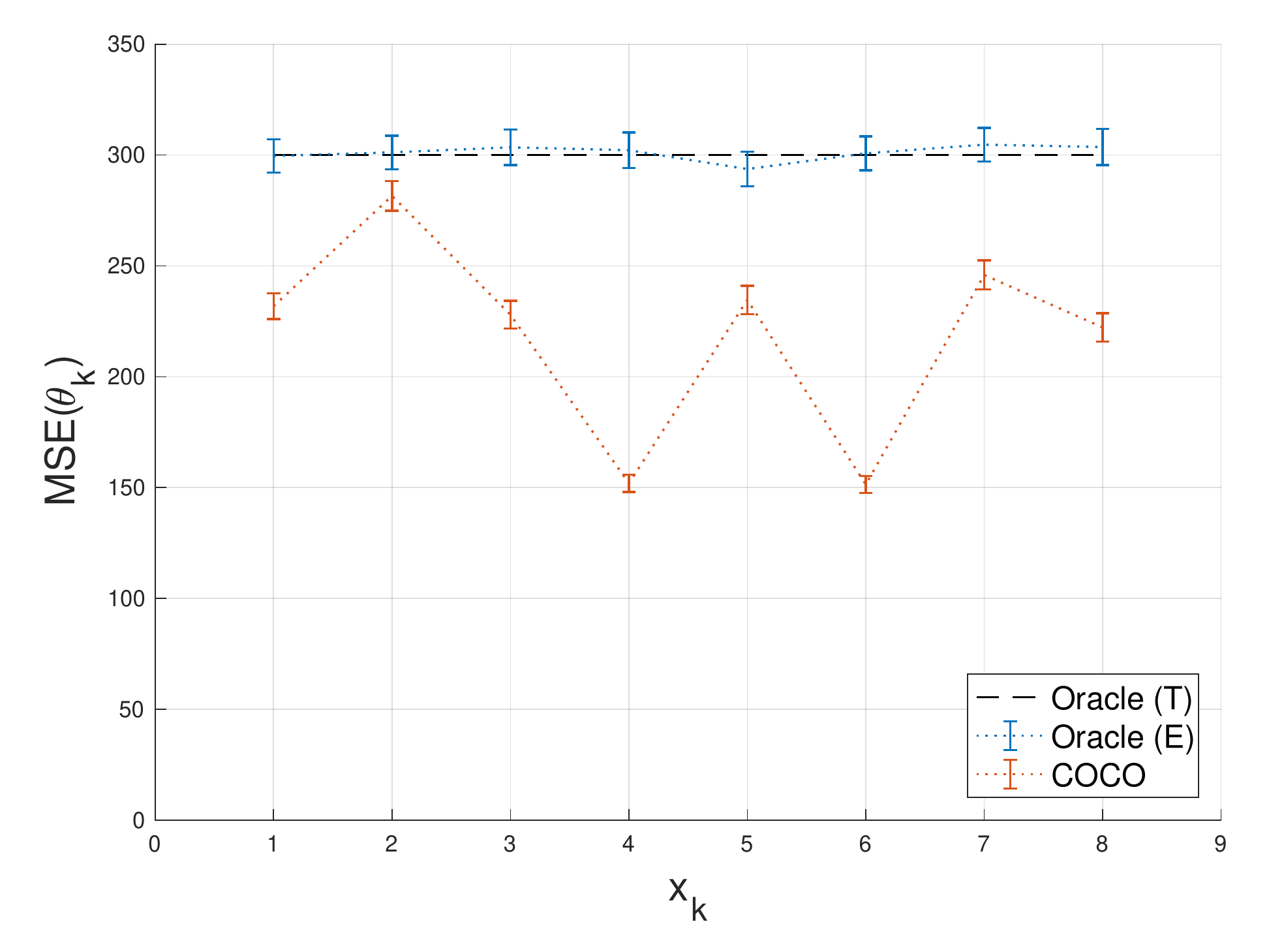}
\includegraphics[width=0.32\textwidth]{Images/MSE_eachpoint_results3_1.pdf}\\
\includegraphics[width=0.32\textwidth]{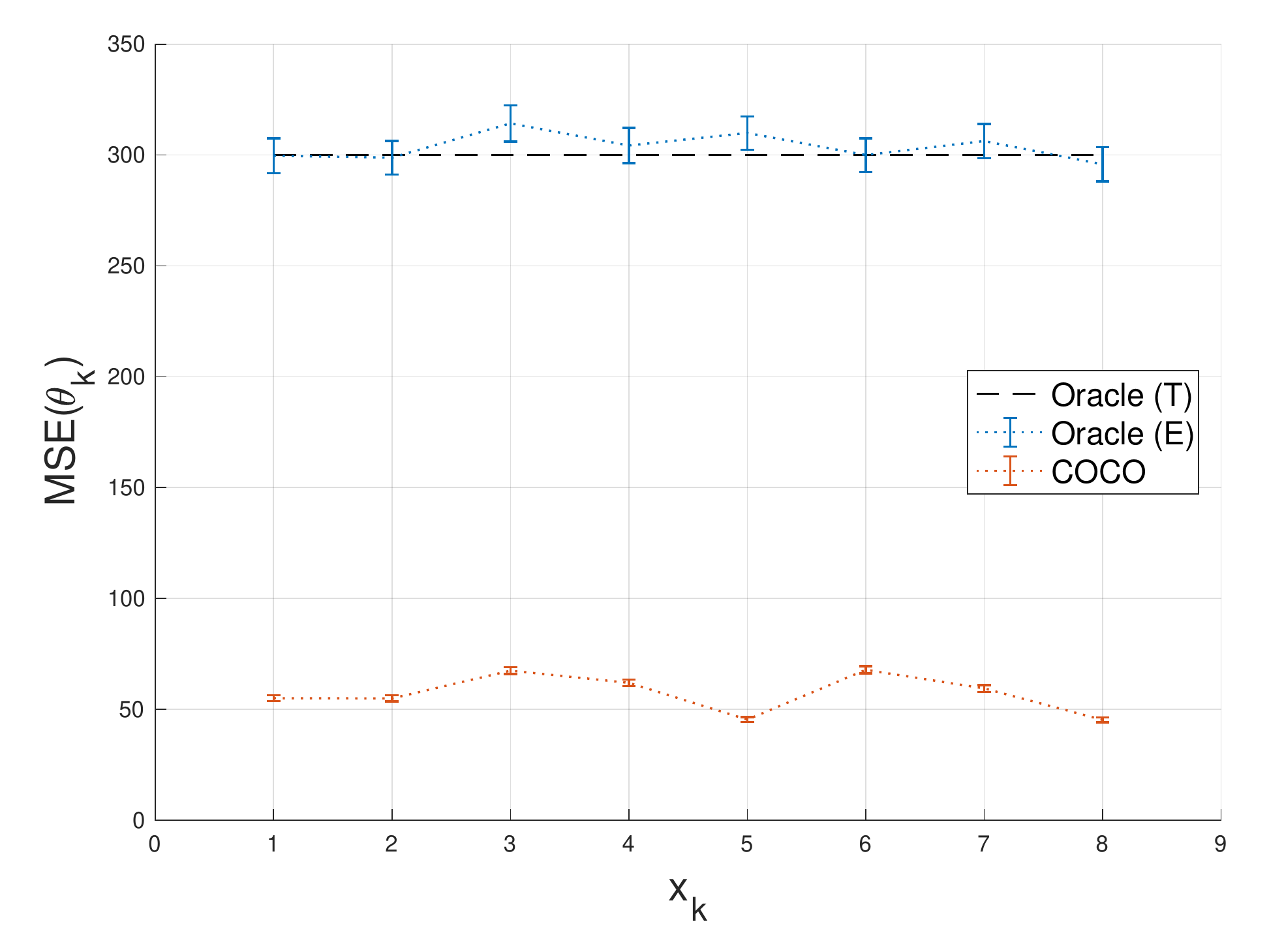}
\includegraphics[width=0.32\textwidth]{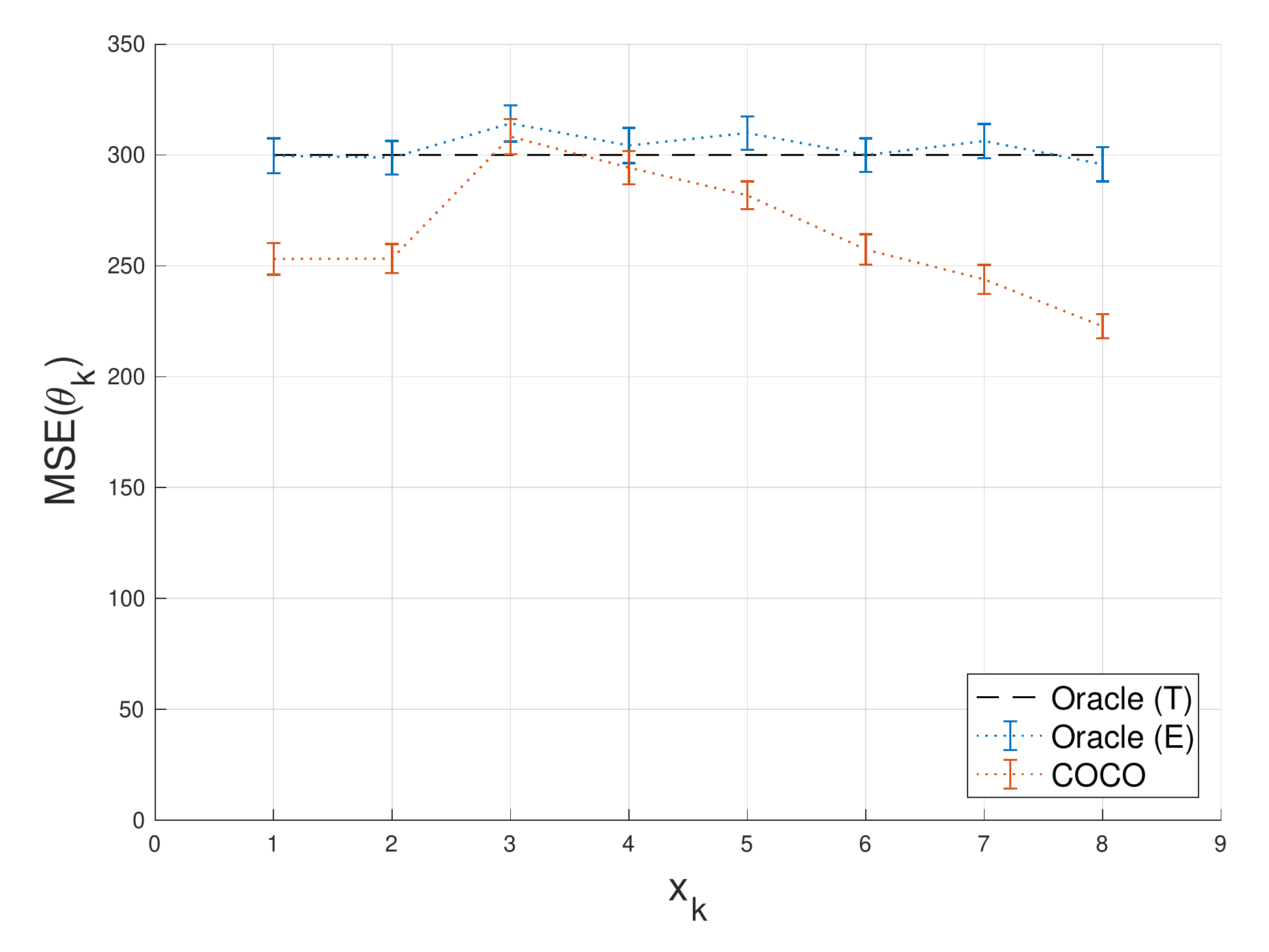}
\includegraphics[width=0.32\textwidth]{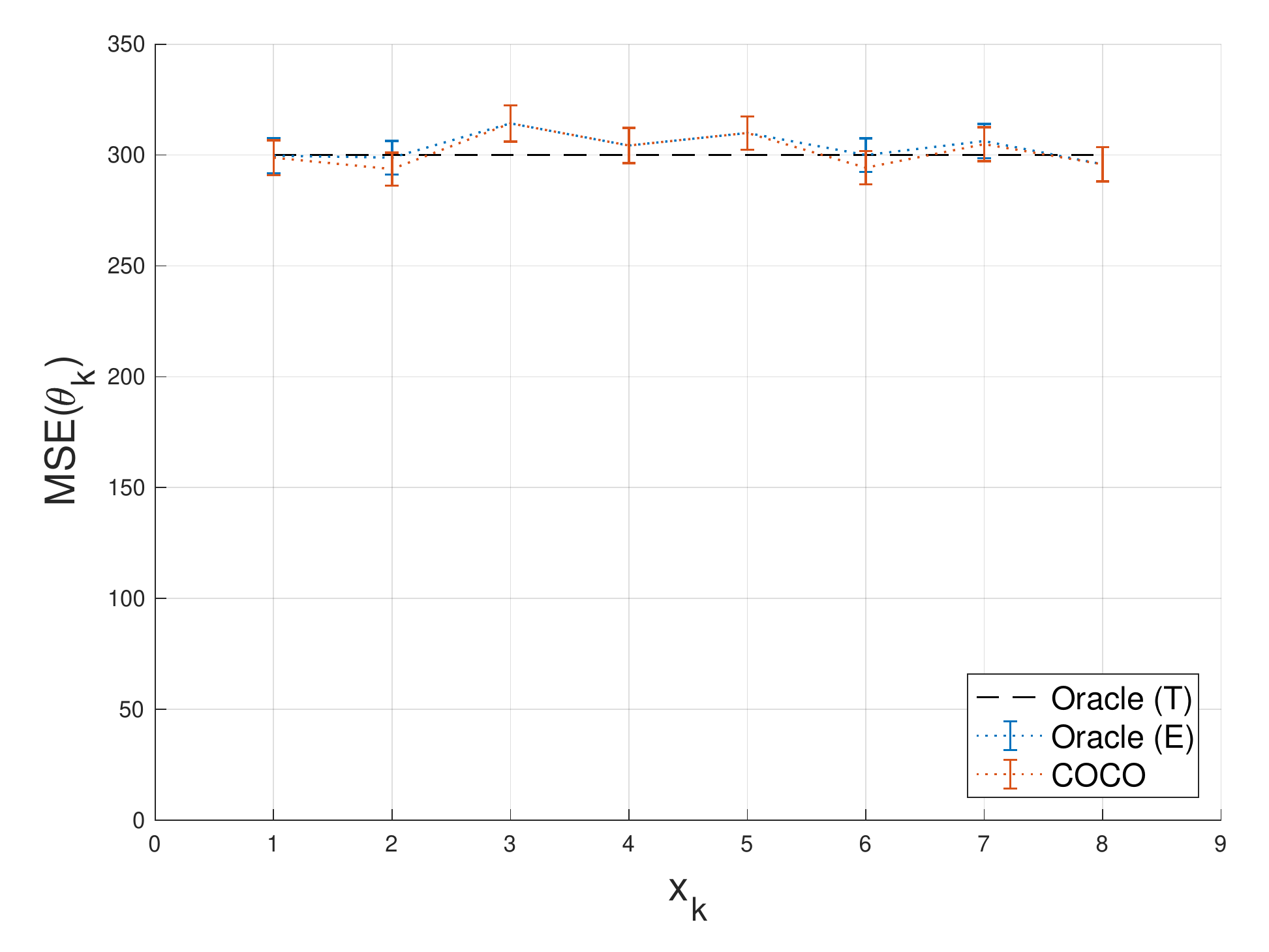}\\
\includegraphics[width=0.32\textwidth]{Images/MSE_eachpoint_results1_3.pdf}
\includegraphics[width=0.32\textwidth]{Images/MSE_eachpoint_results2_3.pdf}
\includegraphics[width=0.32\textwidth]{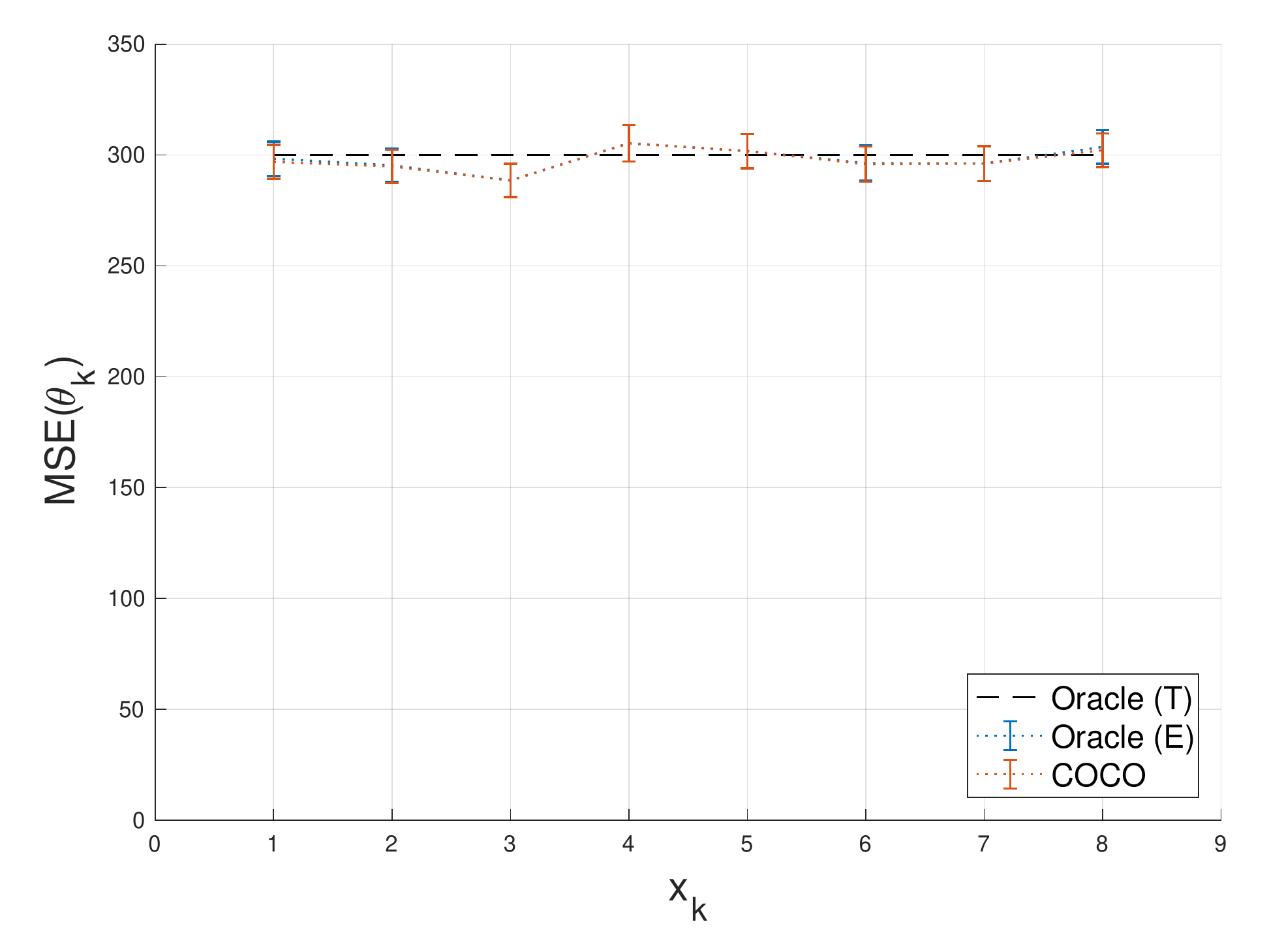}\\
\includegraphics[width=0.32\textwidth]{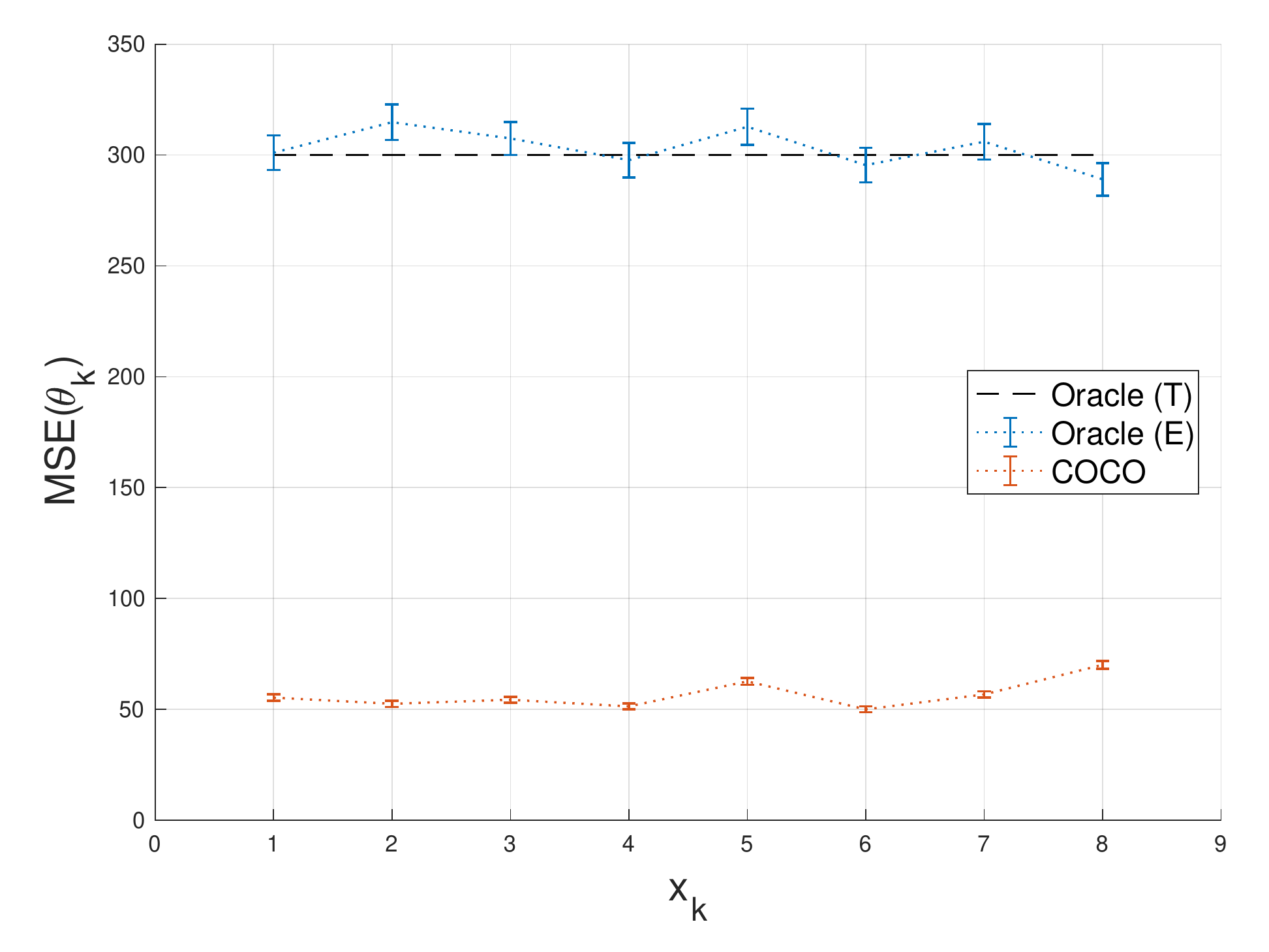}
\includegraphics[width=0.32\textwidth]{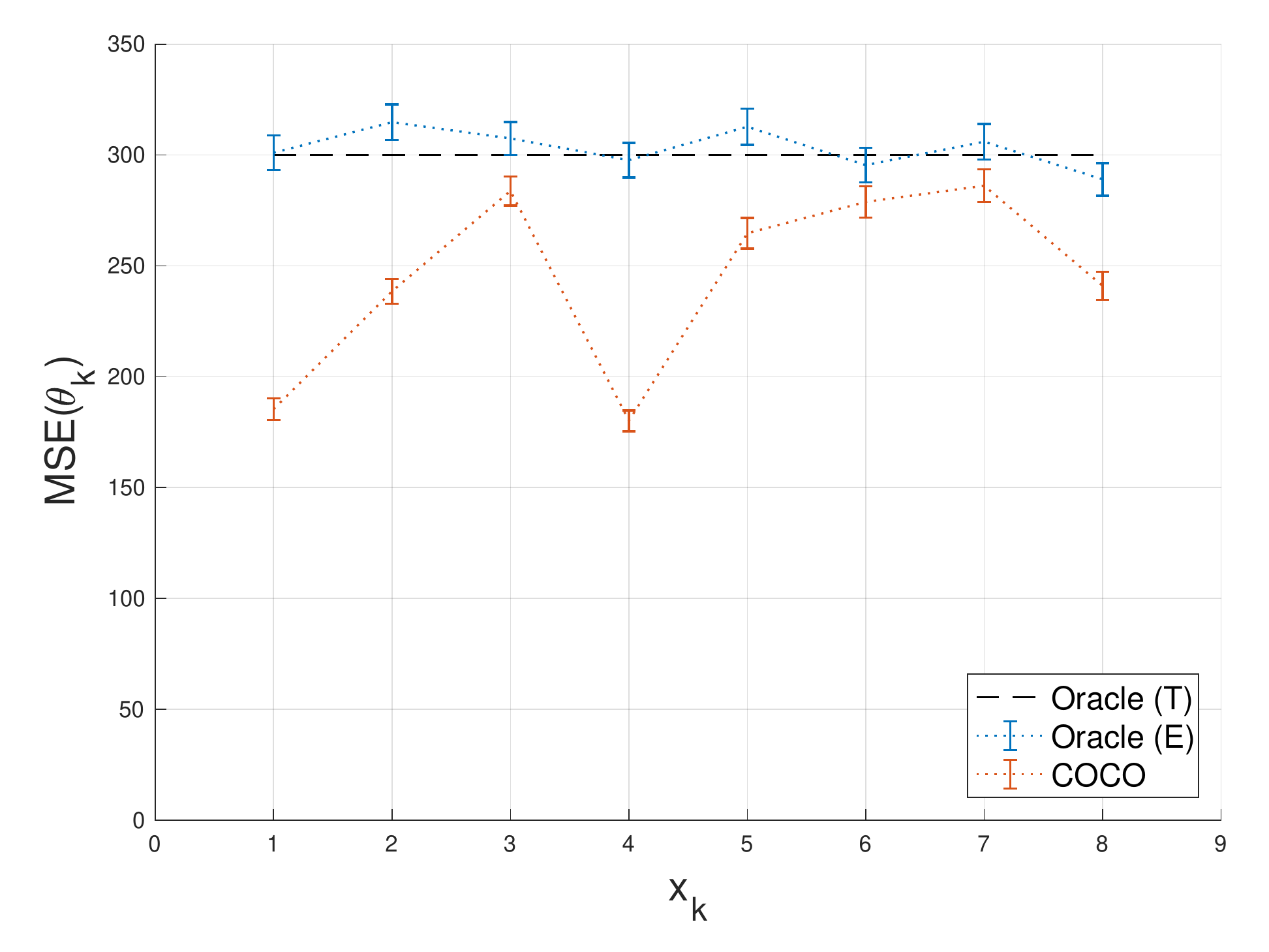}
\includegraphics[width=0.32\textwidth]{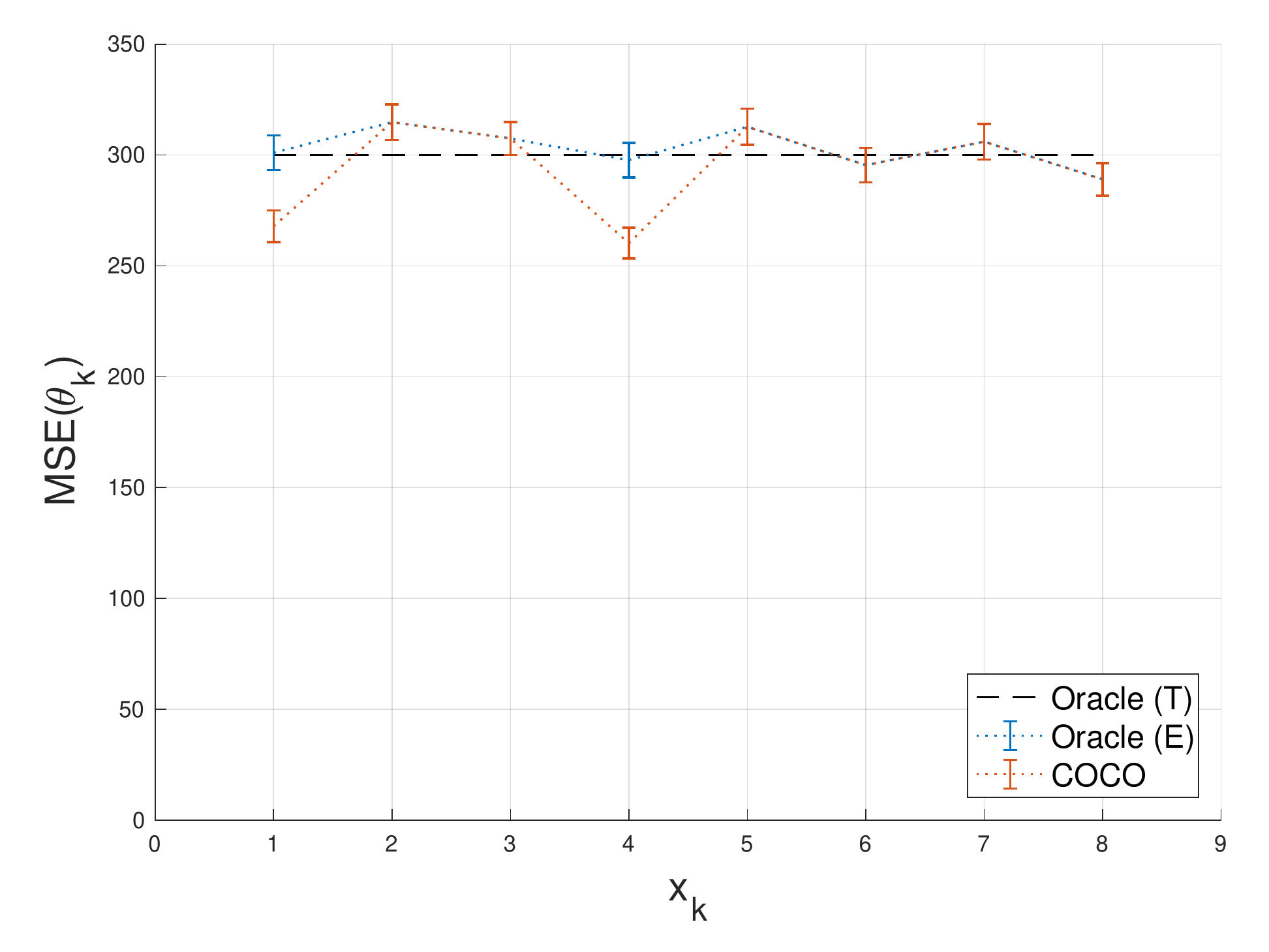}\\
\includegraphics[width=0.32\textwidth]{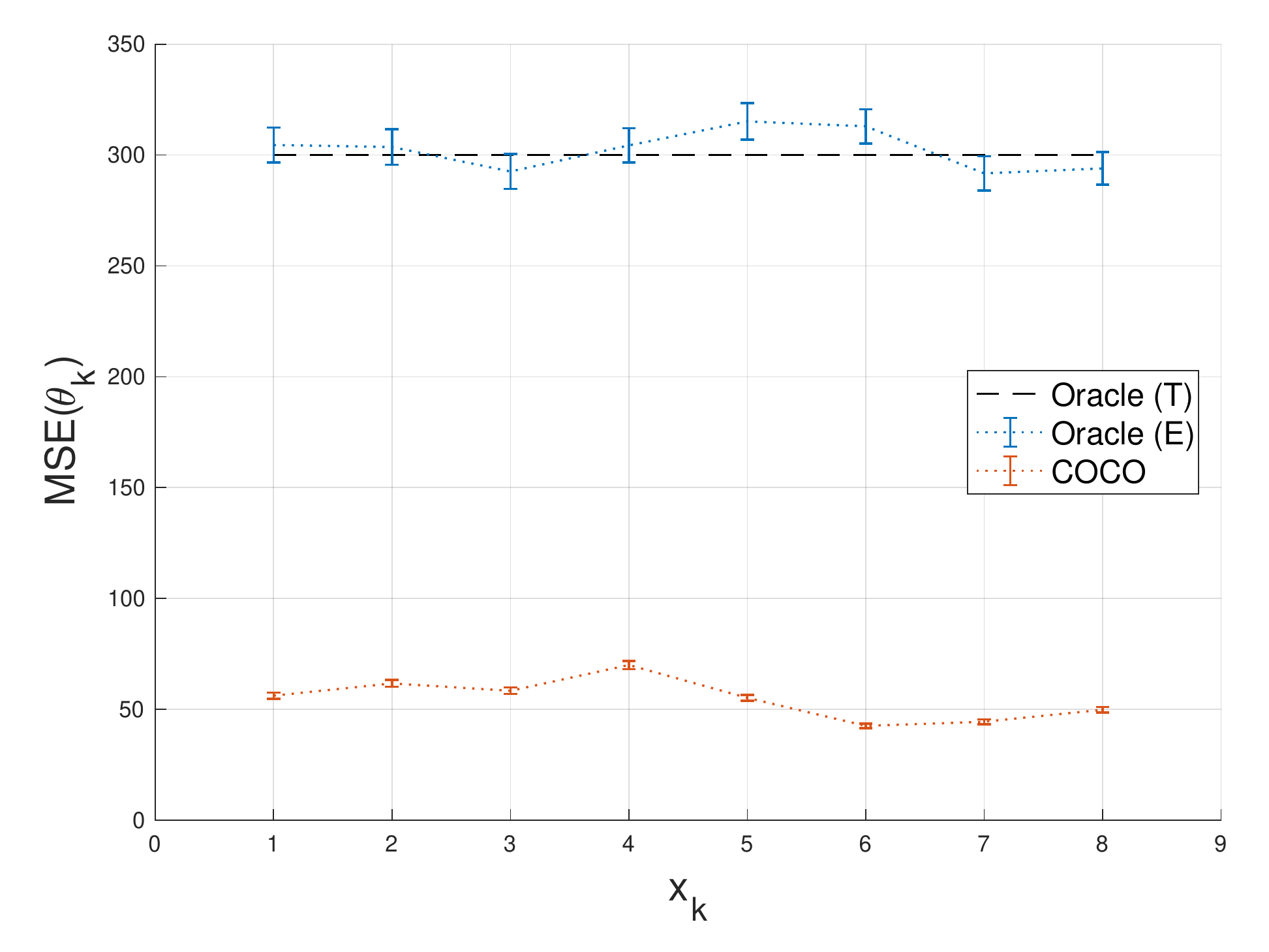}
\includegraphics[width=0.32\textwidth]{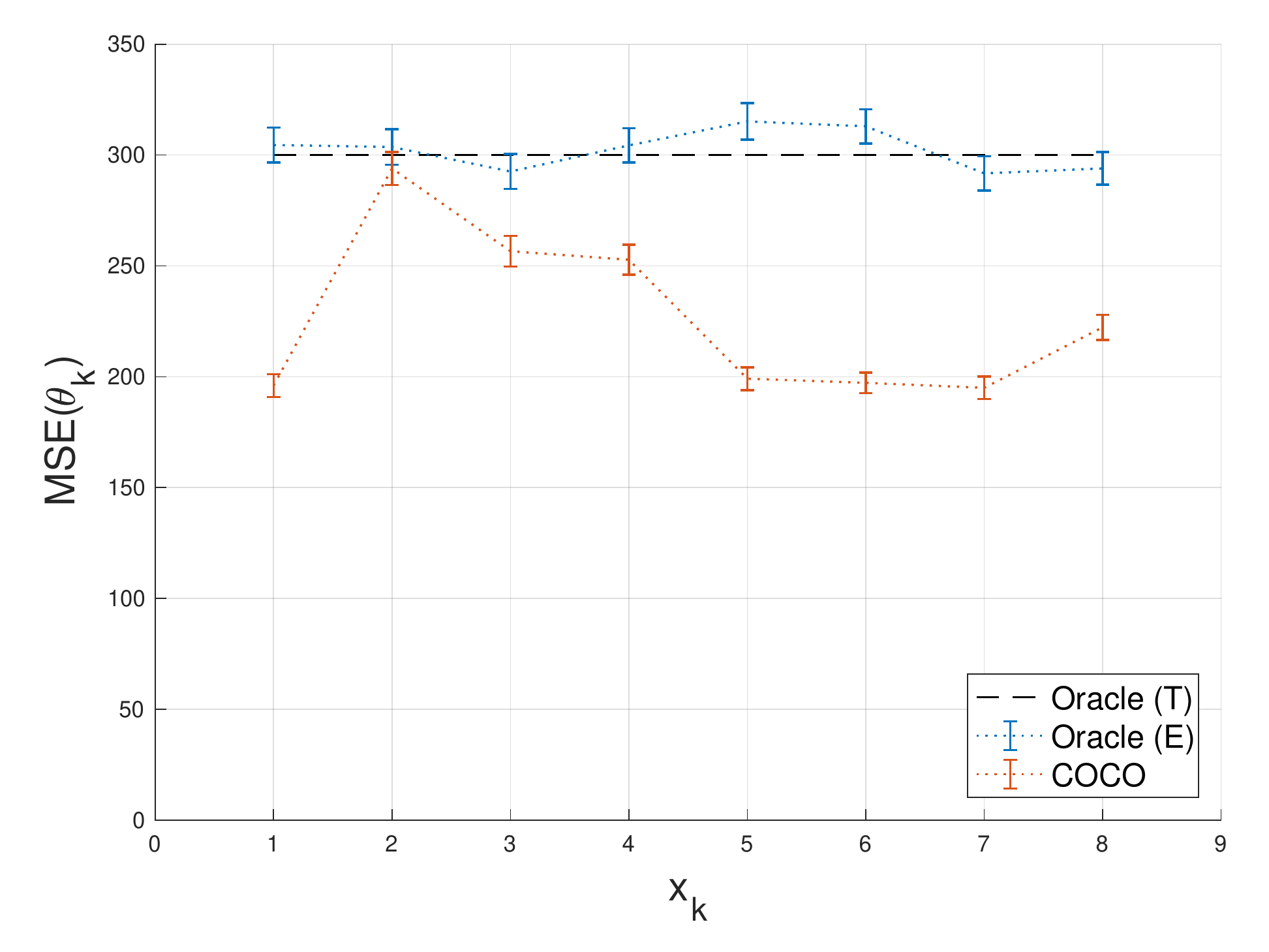}
\includegraphics[width=0.32\textwidth]{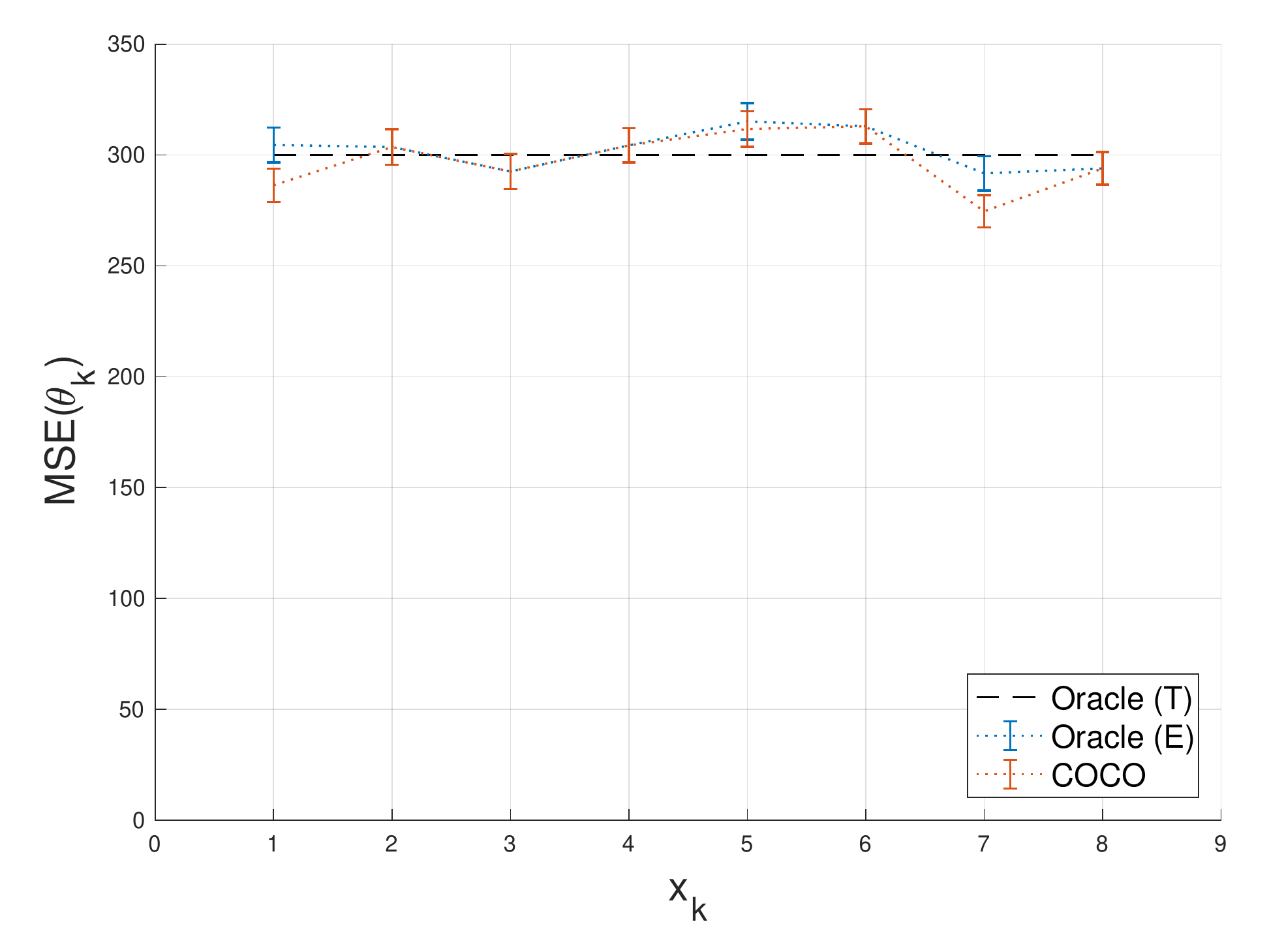}\\
\includegraphics[width=0.32\textwidth]{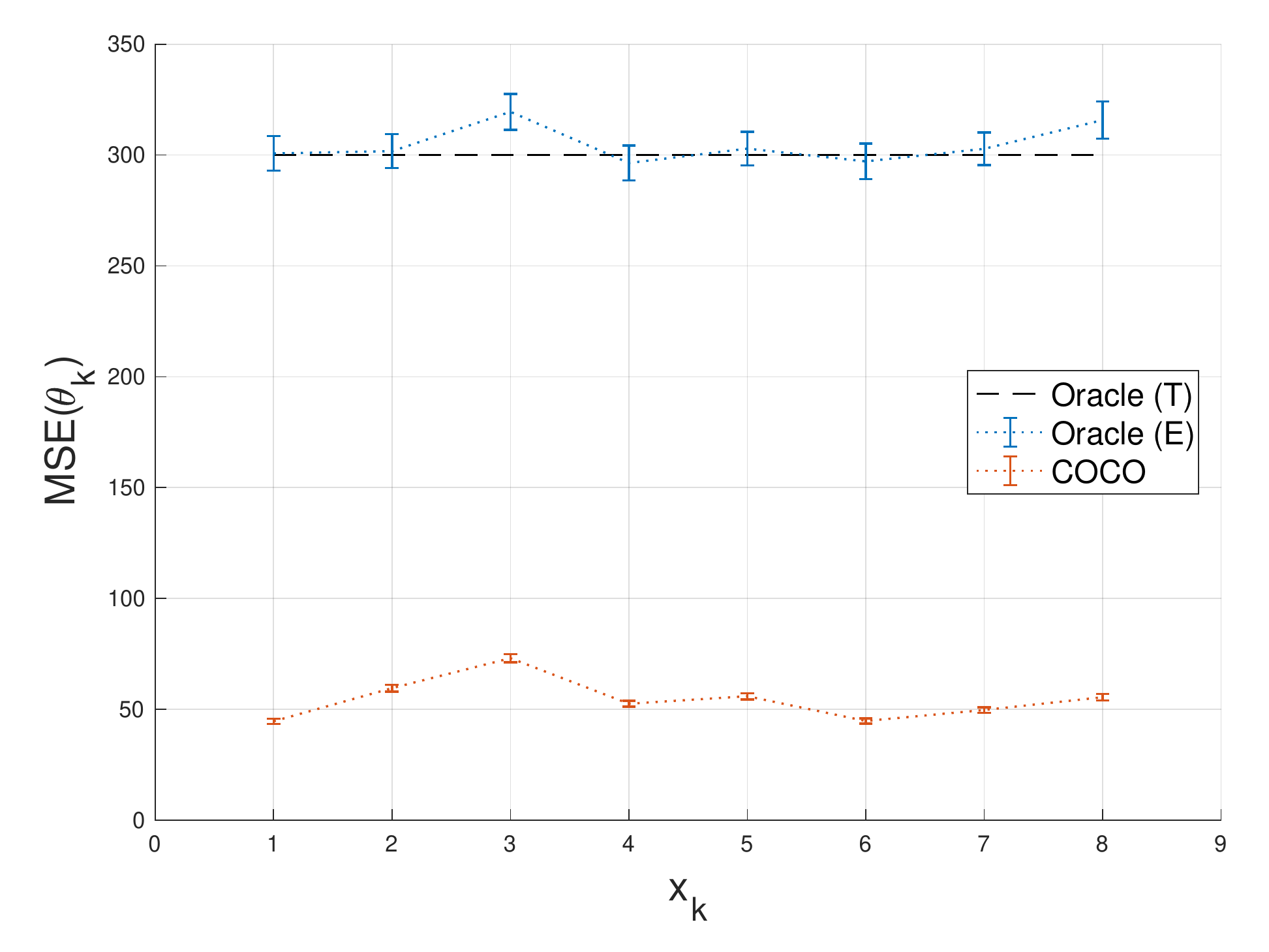}
\includegraphics[width=0.32\textwidth]{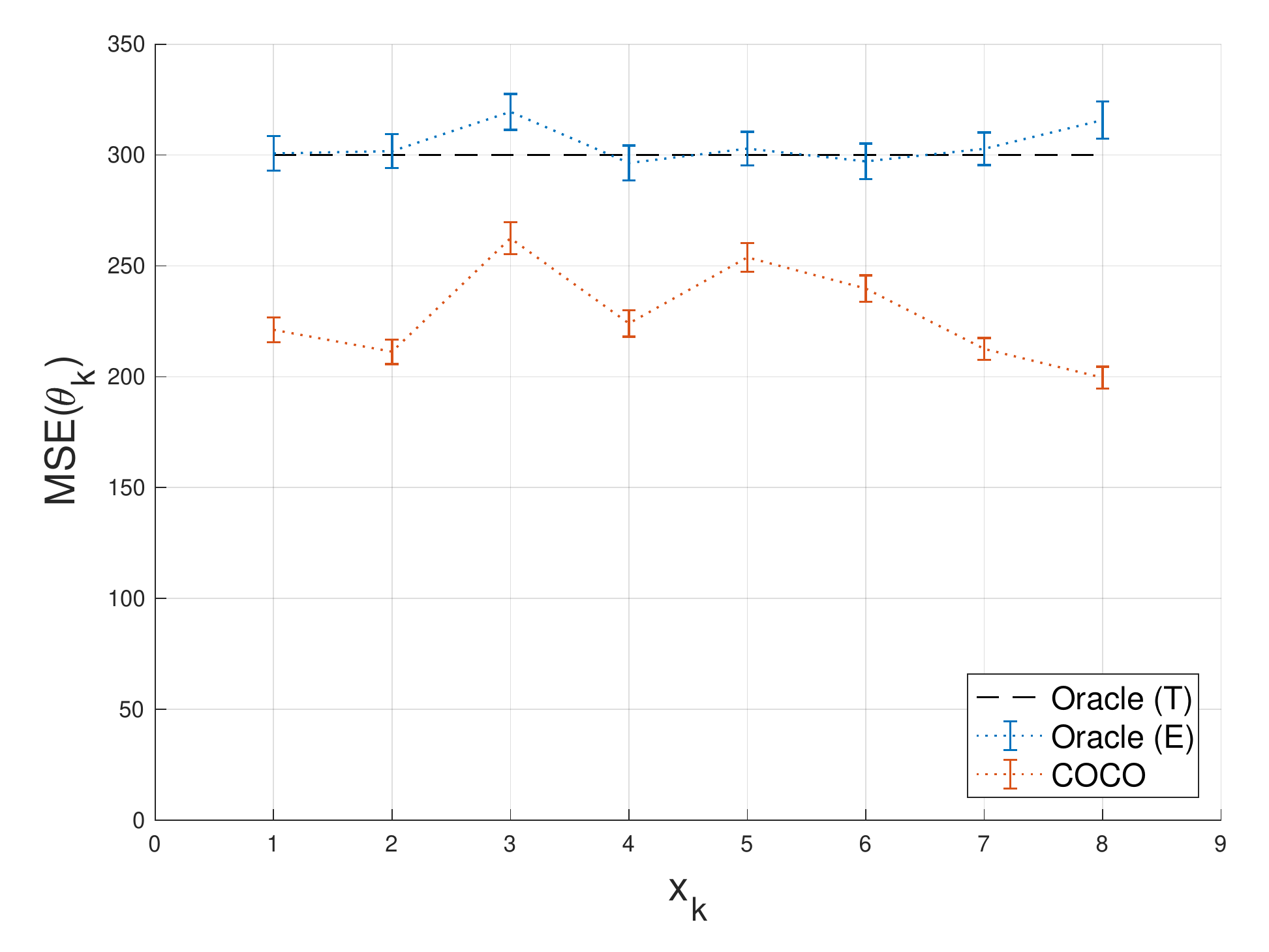}
\includegraphics[width=0.32\textwidth]{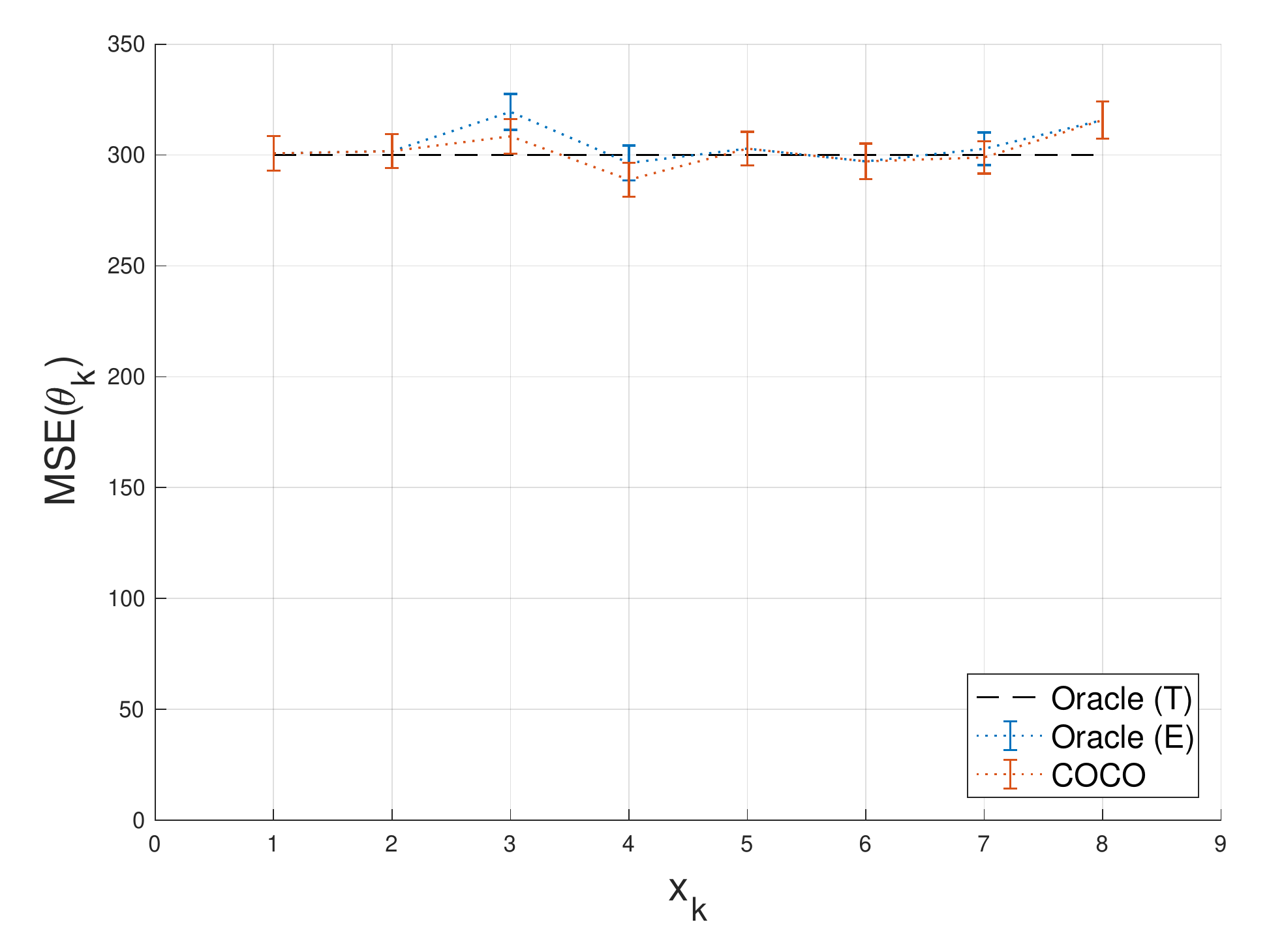}\\
\caption{More instances of the same setup of \Cref{fig:ElementwiseMSE}, with \textit{Left}: $l = 10$; \textit{Center}: $l=100$; \textit{Right}: $l=1000$..}
\label{fig:ElementwiseMSE_SM}
\end{figure}

In this section, we provide additional empirical evidence obtained as far as the elementwise $\operatorname{MSE}$ is concerned, by providing more instances of the results shown in the main body (in \Cref{fig:ElementwiseMSE}). Those results are depicted in \Cref{fig:ElementwiseMSE_SM}. In particular, we observe that $\operatorname{MSE}(\smash{\hat{\theta}}_k) \leq \operatorname{MSE}(g_k)$ for every point in every tested setting. 

\begin{figure}[tbp]
\centering
\includegraphics[width=0.7\textwidth]{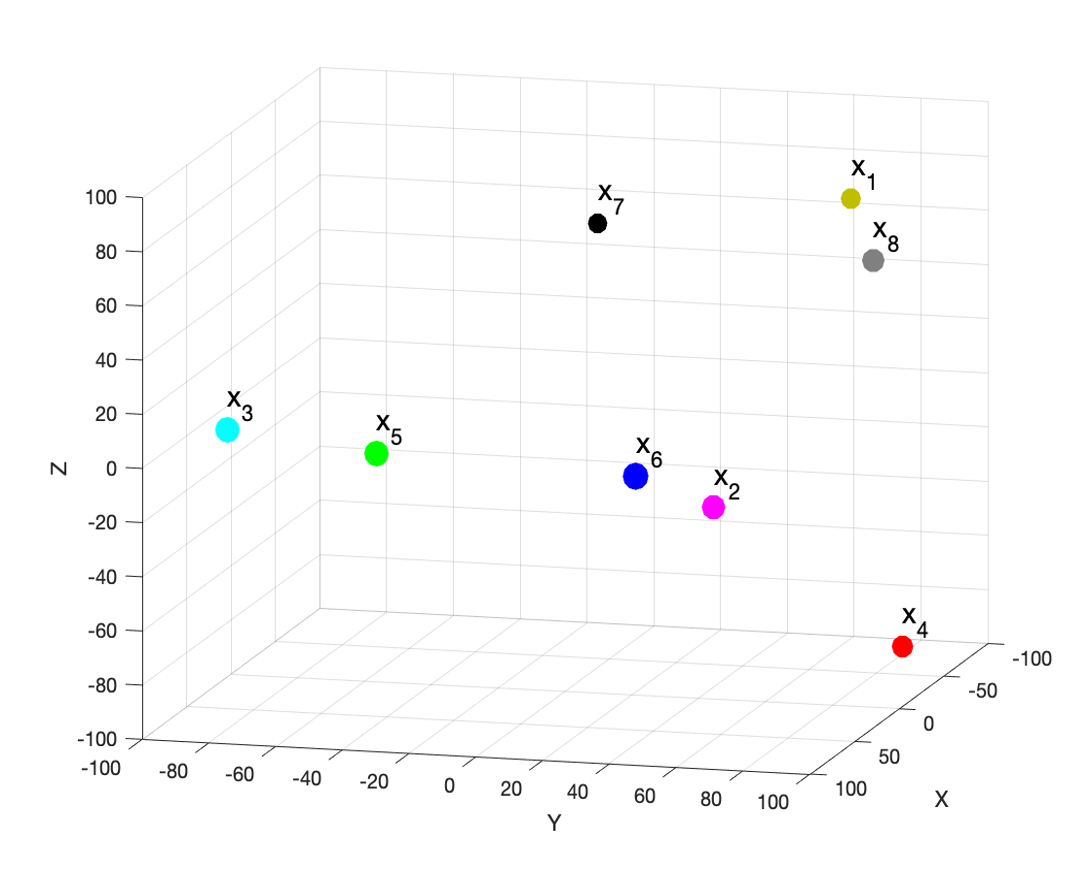}
\caption{Spatial configuration that yields the results in the plot on the center of \Cref{fig:ElementwiseMSE} (to provide some depth insight, marker size is proportional to the point $x$-coordinate). From \Cref{fig:ElementwiseMSE}, it is possible to observe that $x_1, x_2, x_6$ and $x_8$ are the points with the best $\operatorname{MSE}(\smash{\hat{\theta}}_k)$, followed by $x_3$ and $x_5$. Finally, the worst $\operatorname{MSE}(\smash{\hat{\theta}}_k)$ is obtained for $x_4$ and $x_7$. Here we see that this denoising performance can be assigned to the closeness to other points.}
\label{fig:MSE_singular_RndConfig_disposition}
\end{figure}

We also emphasize that $\operatorname{MSE}(\smash{\hat{\theta}})$ does not distribute evenly among the different points, as the $\operatorname{MSE}(\smash{\hat{\theta}}_k)$ varies from point to point. In particular, points which have other points closer have lower $\operatorname{MSE}(\smash{\hat{\theta}}_k)$, whereas more isolated points show higher $\operatorname{MSE}(\smash{\hat{\theta}}_k)$. This can be easily assessed by comparing the relative positions of the points represented in \Cref{fig:MSE_singular_RndConfig_disposition} with the $\operatorname{MSE}(\smash{\hat{\theta}}_k)$ obtained for each of them (center plot from \Cref{fig:ElementwiseMSE}).

\subsection{Bias}

\begin{figure}[tbp]
\centering
\includegraphics[width=1\textwidth]{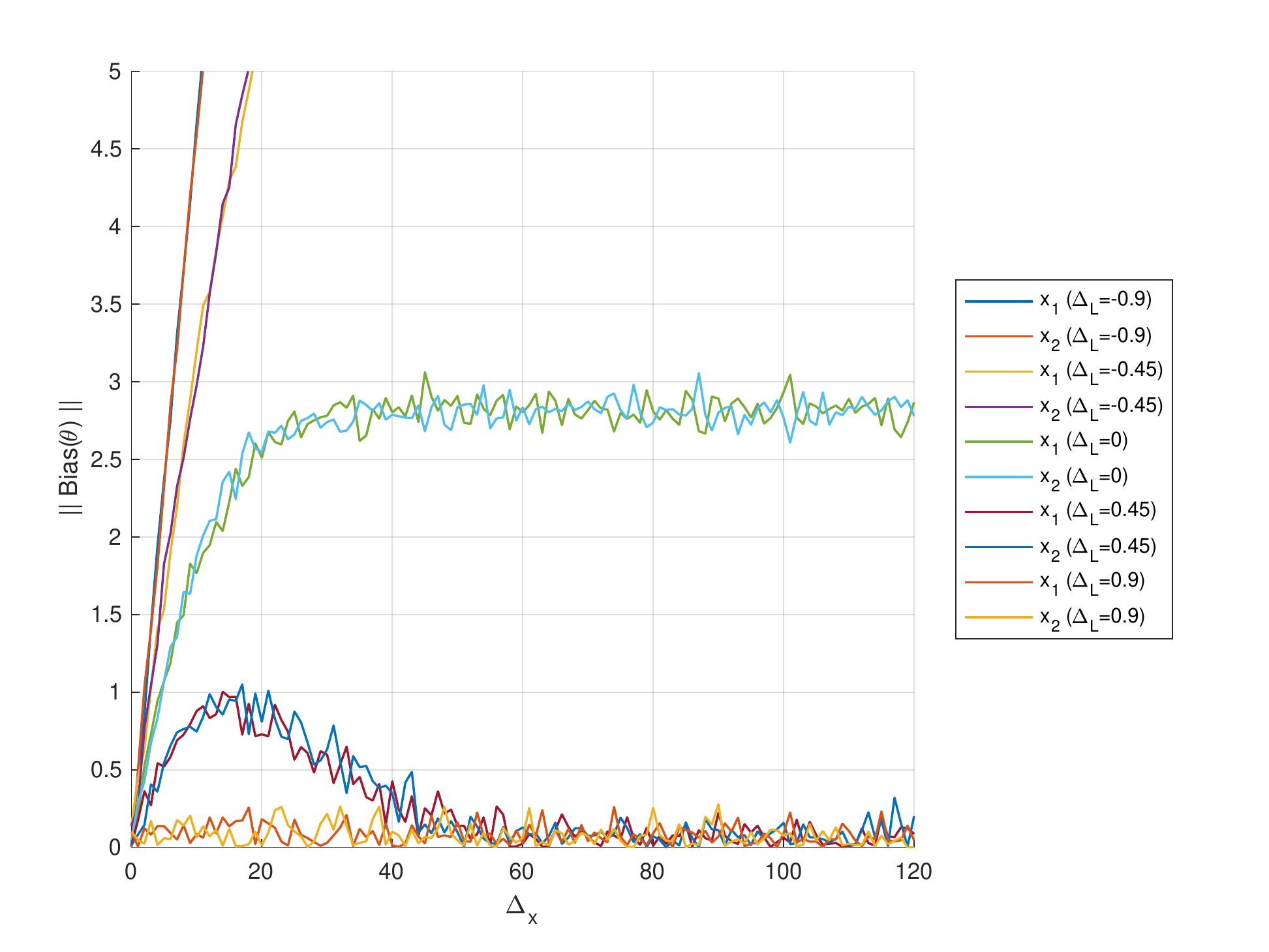}
\caption{
Estimated bias of the COCO gradient estimates as a function of the distance between the points considered, for the setup of \Cref{fig:MSEvsDistance}.}
\label{fig:Bias}
\end{figure}

The bias of the gradient estimators here analysed can be formulated as: for each point $x_k$, $\operatorname{Bias}(\smash{\hat{\theta}}_k)= E[\smash{\hat{\theta}}_k - \nabla f(x_k)] = E[\smash{\hat{\theta}}_k] - \nabla f(x_k)$ and $\operatorname{Bias}(g_k)= E[g_k - \nabla f(x_k)] = E[g_k] - \nabla f(x_k)$. By definition, the oracle whose noise follows the additive and normally distributed model is an unbiased estimator of the gradient since $E[g_k] = \nabla f(x_k)$. We are interested in also characterizing  the behavior of $\smash{\hat{\theta}}_k$ in this respect. In order to test the bias of the COCO denoiser estimator, we estimated $\| \operatorname{Bias}(\smash{\hat{\theta}}_k) \|$ (via Monte-Carlo simulations) in the same setup of \Cref{fig:MSEvsDistance}. These results are presented in \Cref{fig:Bias}:
\begin{itemize}
    \item In all cases, for $\Delta_x = 0$, the COCO estimator is unbiased since it consists of the averaging estimator;
    \item For $\Delta_L<0$, the bias of this estimator seems to grow linearly with $\Delta_x$. The smaller the $\Delta_L$, the higher the slope of that linear relation;
    \item For $\Delta_L = 0$, the estimator is biased as well. That bias grows until a stabilization, which happens at the $\Delta_x$ that it happened with $p_{\text{active}}$;
    \item For $\Delta_L > 0$, the estimator is  also biased. As the COCO estimator outputs become more similar to the ones of the oracle, its bias decreases. Moreover, the higher the $\Delta_L$, the lower the bias (as the constraints are less restrictive).
\end{itemize}
These observations suggest that if the constraint between two gradient estimates is active, then it imposes bias on the COCO estimator. On the other hand, when the constraint is inactive, the COCO estimator outputs the oracle estimates, which are known to be unbiased. 

\section{Comparison to Other Variance Reduction Approaches in Stochastic Optimization}\label{sec:appendix_comparison_other_VR}

\begin{figure}[tbp]
\centering
\includegraphics[width=.9\textwidth]{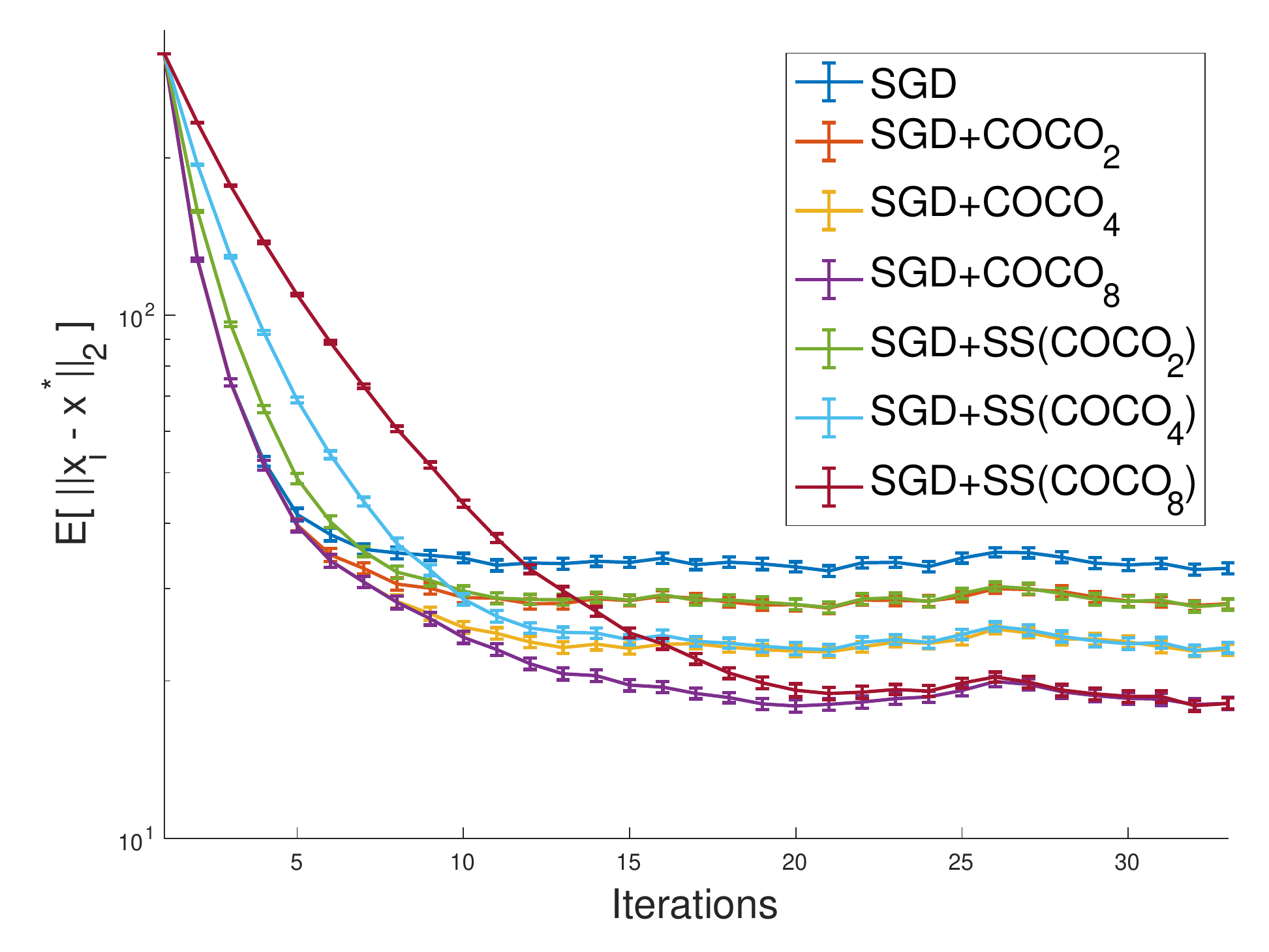}
\caption{
On the same setup considered in \Cref{fig:synthetic_dataset_SGD&Adam_Performance}, we compare the performance obtained by using COCO$_K$ coupled to SGD (SGD+COCO$_K$, $K=2,4,8$) to the performance of SGD when using a step size that allows to achieve the same variance regime (SGD+SS(COCO$_K$), $K=2,4,8$). When comparing methods that attain the same variance regime, the advantage of using COCO rather than a smaller step size is clear in the bias regime.}
\label{fig:COCOvstunedSStovariance}
\end{figure}

We now show that the estimation improvement provided by COCO is non-trivial, namely whether SGD with a smaller step size strictly dominates COCO in both bias and variance. We settle a step size for SGD that allows us to achieve the same variance regime as SGD+COCO. Those step sizes are tuned empirically. In those conditions, we are interested in assessing the differences between methods in terms of bias regime. The results obtained are shown in \Cref{fig:COCOvstunedSStovariance}. We observe that SGD with a smaller step size is clearly outperformed by SGD+COCO, as the latter exhibits a much faster convergence towards its variance regime.

\begin{figure}[tbp]
\centering
\includegraphics[width=.9\textwidth]{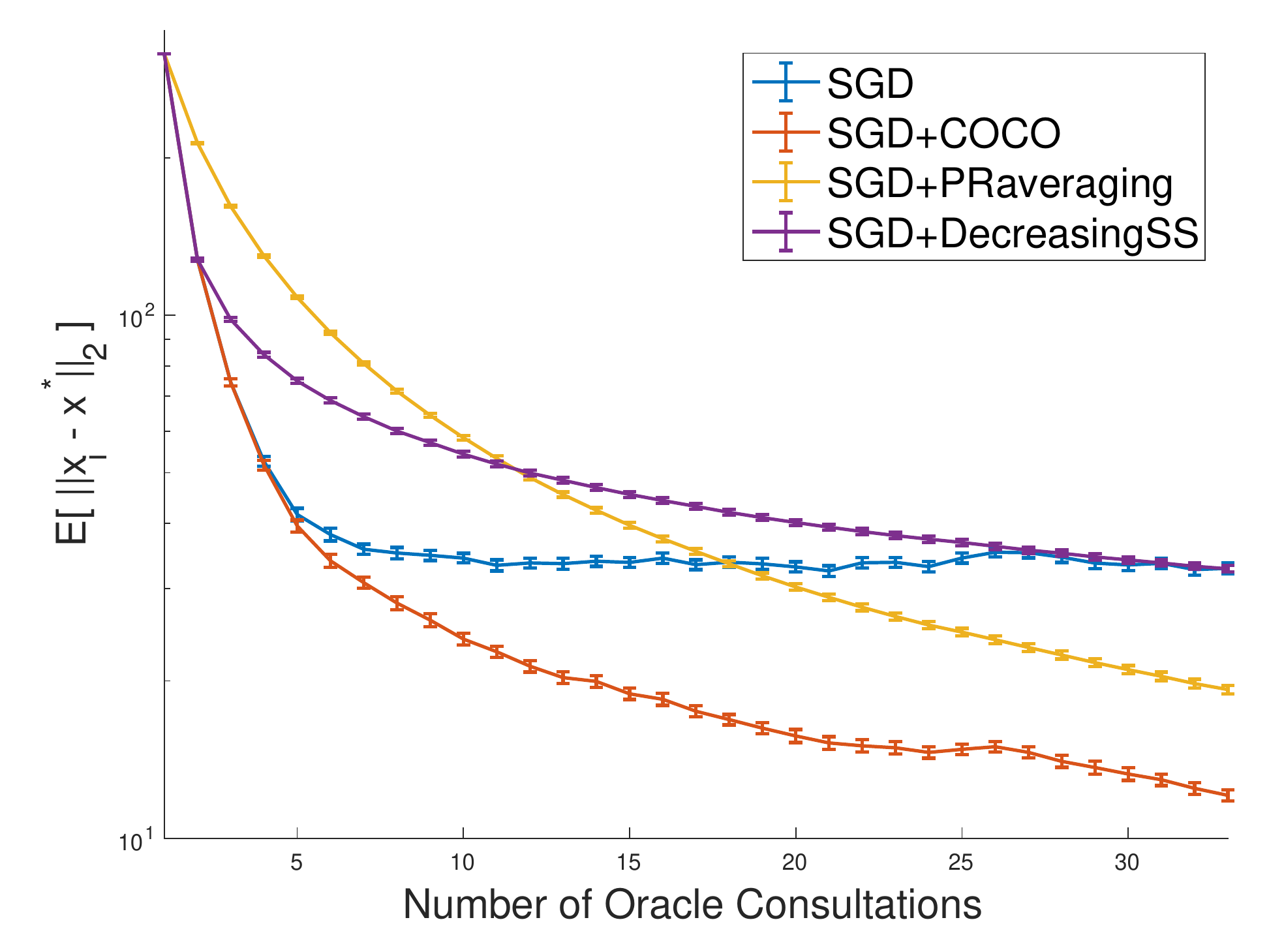}
\caption{
On the same setup considered in \Cref{fig:synthetic_dataset_SGD&Adam_Performance}, we compare the improvement brought by using COCO coupled to SGD (SGD+COCO) with the $O(1/k)$ decreasing step sizes approach (SGD+DecreasingSS) and Polyak-Ruppert averaging of the iterates (SGD+PRaveraging). While SGD with fixed step size stagnates in a variance regime, SGD+DecreasingSS overcomes this limitation, albeit with a bias regime delay. SGD+PRaveraging is even slower at the beginning but quickly outperforms the previous two methods. Nevertheless, using COCO without truncated history yields indisputably the best performance at any iteration.}
\label{fig:COCOvsOtherVR}
\end{figure}

We are also interested in comparing the performance obtained by coupling SGD to COCO with the other variance reductions introduced in \Cref{sec:related_work}: decreasing step sizes with $O(1/k)$ and Polyak-Ruppert averaging of the iterates. We represent those results in \Cref{fig:COCOvsOtherVR}. It is possible to observe that while SGD has the fastest bias regime until stagnation in the variance regime, all the others continue to converge towards the objective minimum. Moreover, the best performance is obtained by COCO coupling, with a bias regime as fast as the SGD and a variance regime convergence rate that seems to be similar to the obtained via Polyak-Ruppert averaging. We note, nevertheless, that to COCO achieve this variance regime convergence, it has to consider all the queried points until that iteration. Naturally, this imposes a computational burden that is not tractable for much further iterations, contrarily to the other two methods, which continue to iterate equally fast.

\end{document}